\documentclass[10pt]{article} % For LaTeX2e
\usepackage[preprint]{tmlr}
\usepackage{amsmath,amsfonts}
\usepackage{algorithmic}
\usepackage{algorithm}
\usepackage{array}
\usepackage[caption=false,font=normalsize,labelfont=sf,textfont=sf]{subfig}
\usepackage{textcomp}
\usepackage{stfloats}
\usepackage{verbatim}
\usepackage{graphicx}
\usepackage{amsmath,amsfonts,bm}

\def\eqref#1{equation~\ref{#1}}
\def\1{\bm{1}}

\DeclareMathAlphabet{\mathsfit}{\encodingdefault}{\sfdefault}{m}{sl}
\SetMathAlphabet{\mathsfit}{bold}{\encodingdefault}{\sfdefault}{bx}{n}

\usepackage{amsmath}                            % (1)
\usepackage{amssymb}                            % (1)
\usepackage{amsthm}                             % (2)
\usepackage{graphicx}                           % (3)
\usepackage{url}

\usepackage{tikz}
\usepackage{comment}
\usepackage{color}
\usepackage{mathabx}
\usepackage{mdframed}
\usepackage{custom_symbols}

\usepackage{float}
\usepackage[accsupp]{axessibility}  % Improves PDF readability for those with disabilities.

\usepackage[utf8]{inputenc} % allow utf-8 input
\usepackage[T1]{fontenc}    % use 8-bit T1 fonts
\usepackage{booktabs}       % professional-quality tables
\usepackage{amsfonts}       % blackboard math symbols
\usepackage{nicefrac}       % compact symbols for 1/2, etc.
\usepackage{microtype}      % microtypography
\usepackage{xcolor}         % colors
\usepackage{epsfig}
\usepackage{graphicx}
\usepackage{multirow}
\usepackage{colortbl}
\usepackage{hhline}
\usepackage{paralist}
\usepackage{ulem}
\usepackage{makecell}

\usepackage{caption}
\usepackage{xspace}
\usepackage{rotating}

\usepackage{pifont}% http://ctan.org/pkg/pifont

\definecolor{forestgreen}{rgb}{0.13, 0.55, 0.13}\usepackage[pagebackref,breaklinks,colorlinks,citecolor = forestgreen]{hyperref}

\usepackage[capitalize]{cleveref}
\crefname{section}{Sec.}{Secs.}
\Crefname{section}{Sec.}{Secs.}
\Crefname{table}{Tab.}{Tabs.}
\crefname{table}{Tab.}{Tabs.}
\Crefname{figure}{Fig.}{Figs.}
\crefname{figure}{Fig.}{Figs.}

\def\month{MM}  % Insert correct month for camera-ready version
\def\year{YYYY} % Insert correct year for camera-ready version
\def\openreview{\url{https://openreview.net/forum?id=XXXX}} % Insert correct link to OpenReview for camera-ready version

\newcommand{\celloccupancy}{\ensuremath{\boxed{\bullet}}}

\newcommand{\cellvisittime}{\ensuremath{\boxed{t}}}
\newcommand{\cellexplored}{\ensuremath{\boxed{\text{x}}}}
\newcommand{\cellsemantics}{\ensuremath{\boxed{\smalltriangleup}}}

\newcommand{\revision}[1]{#1}
\newcommand{\citecolor}[2]{\hypersetup{citecolor=forestgreen}\cite{#2}\hypersetup{citecolor=forestgreen}}
\newcommand{\citepcolor}[2]{\hypersetup{citecolor=forestgreen}\citep{#2}\hypersetup{citecolor=forestgreen}}
\newcommand{\revisedcitep}[1]{\citepcolor{blue}{#1}}
\newcommand{\revisedcite}[1]{\citecolor{blue}{#1}}

\DeclareMathOperator{\mean}{mean}
\DeclareMathOperator{\GRU}{GRU}

\let\emph\relax
\DeclareTextFontCommand{\emph}{\itshape}

\def\month{MM}  % Insert correct month for camera-ready version
\def\year{YYYY} % Insert correct year for camera-ready version
\def\openreview{\url{https://openreview.net/forum?id=XXXX}} % Insert correct link to OpenReview for camera-ready version

% ...
% ...
% ...
\title{Semantic Mapping in Indoor Embodied AI -- A Survey on Advances, Challenges, and Future Directions}

\author{\name Sonia Raychaudhuri \email sraychau@sfu.ca \\
       \name Angel X. Chang \email angelx@sfu.ca
       }

\begin{document}

\maketitle

\begin{abstract}

Intelligent embodied agents (e.g. robots) need to perform complex semantic tasks in unfamiliar environments. Among many skills that the agents need to possess, building and maintaining a semantic map of the environment is most crucial in long-horizon tasks. A semantic map captures information about the environment in a structured way, allowing the agent to reference it for advanced reasoning throughout the task.
% for the agent to refer to anytime during the task for complex reasoning. 
% The embodied AI community has studied various semantic map building approaches in conjunction with different types of downstream tasks.
While existing surveys in embodied AI focus on general advancements or specific tasks like navigation and manipulation, this paper provides a comprehensive review of semantic map-building approaches in embodied AI, specifically for indoor navigation. 
% We unify the approaches by examining them across two major axes - how maps are structured and how they store various types of information. 
We categorize these approaches based on their structural representation (spatial \textit{grids}, \textit{topological} graphs, dense \textit{\revision{geometric}} or \textit{hybrid} maps) and the type of information they encode (\textit{implicit} features or \textit{explicit} environmental data).
% Irrespective of how they are structured, these maps may store a learned feature encoding of the observed images (\textit{implicit}) or predefined information about the environment such as occupancy, etc. (\textit{explicit}). 
We also explore the strengths and limitations of the map building techniques, highlight current challenges, and propose future research directions. We identify that the field is moving towards developing open-vocabulary,
queryable, task-agnostic map representations, while high memory demands and computational
inefficiency still remaining to be open challenges. This survey aims to guide current and future researchers in advancing semantic mapping techniques for embodied AI systems.

% This survey examines various approaches towards building semantic map representations in Embodied AI tasks.
% An embodied agent interacts with its environment in order to perform various tasks.
% A crucial capability that this agent needs to learn while performing the tasks is that of creating and maintaining a map of its environment as it moves around. 
% Such a map typically encodes the semantics of the agent's visual observations and can be structured either as a spatial metric map or a graph-based topological map. Moreover, the map may encode various types of semantic information that may be helpful for the downstream tasks.
% Additionally, recent advances in foundation models have led to the development of open-vocabulary semantic map. This survey summarizes these approaches and discusses advantages and limitations of each approach.

\end{abstract}

% \begin{IEEEkeywords}
% Embodied AI, Semantic mapping, Spatial map, Topological map, Open-vocabulary map
% \end{IEEEkeywords}

\section{Introduction}
% \IEEEPARstart{O}{ver}
\begin{figure*}[ht]
\begin{center}
% \framebox[0.9\linewidth]{
% \fbox{\rule[-.5cm]{0cm}{4cm} \rule[-.5cm]{4cm}{0cm}
\includegraphics[width=\linewidth]{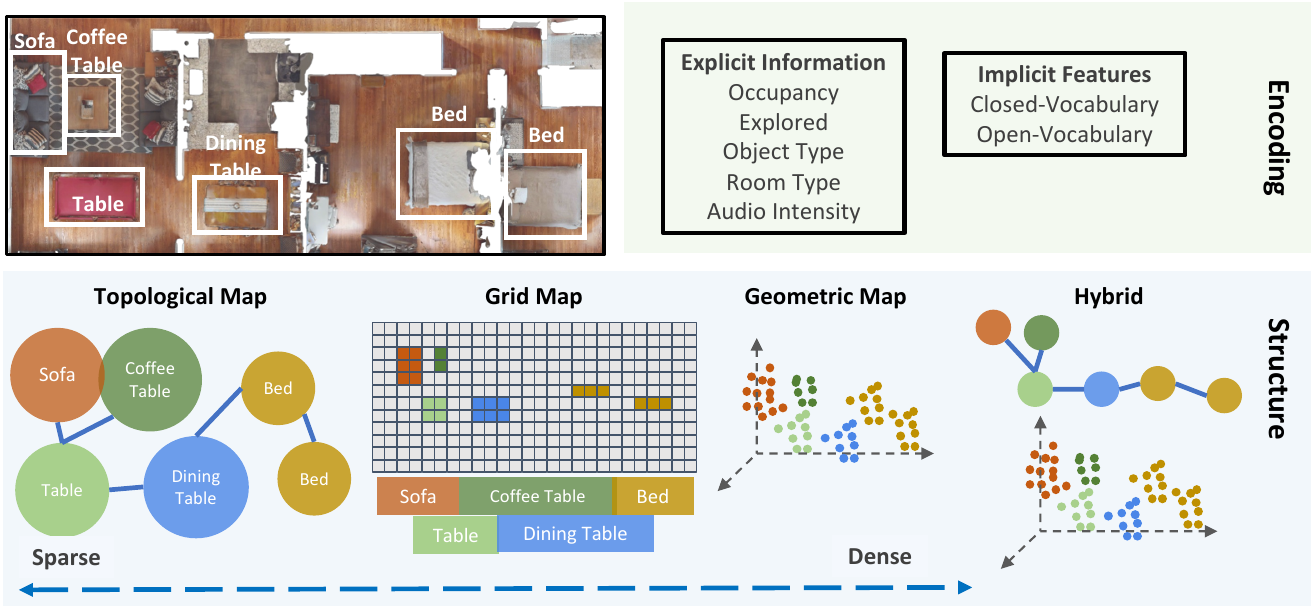}
% }
\end{center}
\vspace{-10pt}
\caption{
% \textbf{Structure and encoding.} 
\revision{\textbf{Semantic maps.}
The survey categorizes semantic map building methods in embodied agents based on their structure and the encoding it stores.
\emph{Structure}: a map of a physical environment can be structured as a topological map (with nodes and edges), a spatial grid, a \revision{dense geometric map} or a hybrid map combining two or more of the others. 
\emph{Encoding}: the structured maps can store either explicit (occupancy, object type, etc.) or implicit information (learned neural features) corresponding to the observation made at that location. 
% The figure shows examples of types of explicit information that can be stored at each cell or node.  Implicit features are typically extracted and aggregated using vision encoders such as ResNet for close-set vocabulary, or large-vision language models such as CLIP~\cite{radford2021learning} for open-vocabulary semantics.
}
}
\label{fig:mapping_overview}
\end{figure*}

\revision{
Intelligent agents, whether physical robots or virtual embodied systems, must operate in complex, unstructured environments where effective behavior demands more than just low-level sensing and actuation. To act meaningfully, agents must form structured internal representations that link perception to reasoning and decision-making. Semantic maps serve this role by encoding not only spatial geometry but also high-level semantics of the environment (object categories, affordances, etc.). Semantic mapping is thus foundational in both robotics and embodied AI, particularly in open-world settings like autonomous driving~\citep{bao2023review,li2024lanesegnet}, search-and-rescue operations~\citep{gautham20233d}, automated cleaning robots~\citep{singh2023vision} among others.
While traditional mapping techniques prioritize geometric accuracy for localization and obstacle avoidance, recent progress in deep learning, computer vision and multi-modal perception has shifted focus toward semantically rich maps. Despite extensive literature, semantic mapping remains an open and rapidly evolving research area.
}

\revision{
Existing surveys review semantic mapping within the context of its usage in downstream applications, thus largely centered on task advancement~\citep{pfeifer2004embodied,kostavelis2015semantic,cadena2017past,xia2020survey,duan2022survey,deitke2022retrospectives,zhu2021deep,zhang2022survey,wu2024embodied,lin2024embodied,garg2020semantics,achour2022collaborative,zheng2024survey}. \cite{song2025semantic} has recently reviewed semantic mapping in indoor robotics applications, but focused on the type of semantics acquired and the techniques of such extraction. 
In contrast, our survey offers a comprehensive review of semantic map-building methods, organized around the underlying map representations themselves, independent of specific downstream tasks. This perspective is both timely and necessary, given the recent advances in foundation models and the growing demand for general-purpose, open-vocabulary, task-agnostic representations. }

\revision{
To provide a principled understanding of semantic map-building methods, this survey categorizes the literature along two fundamental axes (see \Cref{fig:mapping_overview}): map \emph{structure} (e.g., topological graphs, spatial grids, dense geometric, and hybrid representations) and semantic \emph{encoding} (explicit annotations vs. learned implicit features). This taxonomy reflects core design choices that influence map scalability, interpretability, generalization, multi-modal fusion and querying. By organizing approaches in this manner, we aim to unify diverse strands of research (see \Cref{tab-sem-map-grouped} for a summary of the reviewed papers), highlight the trade-offs between different representations and present key challenges and future opportunities in semantic mapping.
This survey centers on semantic mapping approaches in the context of indoor mobile robots, a domain that offers a well-defined, practically relevant and technically rich environment for research. We focus on semantic mapping within embodied AI, that enables controlled environments capable of isolating high-level reasoning abilities of semantic maps from the complexities of noisy sensors and real-world observations.
At the same time the survey establishes connections to SLAM-based approaches to underscore their methodological overlap and to bridge foundational techniques in robotics with emerging paradigms in embodied AI.
}

\revision{
This survey is organized as follows. We begin the survey in \Cref{sec:prelims} with background context to ground our discussion on semantic mapping. \Cref{sec:semantic_map} introduces semantic maps and lays out the foundational components. We then explore different map structures in \Cref{sec:structure}, followed by encoding strategies in \Cref{sec:representations}. 
\Cref{sec:evaluation} reviews existing evaluation methodologies. In \Cref{sec:challenges}, we examine the key challenges in building semantic maps, leading into \Cref{sec:future}, which outlines promising directions for future research.
We conclude in \Cref{sec:conclusion} by synthesizing the main insights and takeaways from the survey.
}

\begin{table*}[t]
\caption{
% \textbf{Semantic maps in indoor embodied AI.} 
\revision{\textbf{Summary table.}}
We characterize works that use maps for embodied AI \revision{as well as robotics} by the type of structure they use (\textbf{Grid}, \textbf{Topological}, \revision{\textbf{Dense geometric maps}}) and how information is encoded in the map (\textbf{Exp}licit vs \textbf{Imp}licit). 
% Prior works structure semantic maps as: \textit{spatial grid} (Grid), \textit{topological} (Topo) and \textit{point-cloud} (PC). 
\emph{Explicit} encodings are pre-selected information such as occupancy \celloccupancy, explored-area \cellexplored, object category \cellsemantics, visitation time \cellvisittime\xspace and others.  
\emph{Implicit} encodings are learned representations such as visual (V) or visual-and-language (VL) features.  
The use of VL features (typically from large pretrained models) enable building \emph{open vocabulary} maps.  
Works that aggregate implicit features onto a grid map, but finally decode into explicit encodings are marked as `Implicit to Explicit' in this table.
% There are also work that aggregate implicit features into a grid tensor map, but decode from the grid tensor into explicit encodings which are used for planning.  We consider these maps to be explicit maps, with an intermediate implicit map and indicate them as being `Implicit to Explicit' in this table.   
}
\label{tab-sem-map-grouped}
\centering
% remove gaps in vertical lines while keeping a bit of gap between rows
\renewcommand{\arraystretch}{1.5}%
\aboverulesep = 0pt
\belowrulesep = 0pt
   \resizebox{\linewidth}{!}
    {
    \begin{tabular}{|c|c|c|c|}
    \toprule
    & \multicolumn{3}{c|}{\bf Structure} \\
    \midrule
    {\bf Encoding} & {\bf Grid} & {\bf Topological} & {\revision{\bf Dense geometric map}} \\
    \midrule
    \multirow{3}{*}{Explicit (no semantics)} & 
      \makecell{
      \revision{Occupancy Grids}~\citep{30720} \celloccupancy \\
         ANS~\citep{chaplot2020learning} \celloccupancy\cellexplored \\
         \revision{UPEN}~\citep{georgakis2022uncertainty} \celloccupancy \\
      } & \makecell{
         \revision{Topo SLAM}~\citep{choset2001topological}\\ 
         \revision{Topomap}~\citep{blochliger2018topomap} \\
         \revision{Fast}~\citep{chen2022fast}\\
      }
      &\multirow{2}{*}{\makecell{
         \revision{6D-SLAM}~\citep{nuchter20076d}\\ 
         \revision{Robust 3D mapping}~\citep{may2009robust}\\
         % \revision{ElasticFusion}~\citep{whelan2015elasticfusion}\\
         \revision{Droid-SLAM}~\citep{teed2021droid} \\
         \revision{Voldor+ SLAM}~\citep{min2021voldor+} \\
      }} \\
      \cmidrule{2-3}
      & \multicolumn{2}{c|}{
        \makecell{
Integrating~\citep{thrun1998integrating} \celloccupancy  \\
Combining~\citep{tomatis2001combining} \celloccupancy \\  
        }
      } & \\
      \midrule
    Explicit (semantics) & 
      \makecell{
         \revision{Curious George}~\citep{meger2008curious} \celloccupancy\cellexplored\cellsemantics \\
        \revision{Efficient}~\citep{hu2013efficient} \cellsemantics \\
         SemExp~\citep{chaplot2020object}
         \celloccupancy\cellexplored\cellsemantics \\
         \revision{L2M}~\citep{georgakis2021learning} \celloccupancy\cellsemantics \\
         \revision{SEAL}~\citep{chaplot2021seal}\celloccupancy\cellsemantics\\
        \revision{CM$^2$}~\citep{georgakis2022cross}\\
         MOPA~\citep{raychaudhuri2023mopa}\cellsemantics \\
         GOAT-Bench~\citep{khanna2024goat}
         \celloccupancy\cellexplored\cellsemantics \\ MapNav~\citep{zhang2025mapnav}
         \celloccupancy\cellexplored\cellsemantics \\
        \revision{Instruction-guided}~\citep{wang2025instruction}\cellsemantics \\
      } & 
      \makecell{
      \revision{SLAM++}~\citep{salas2013slam++} \cellsemantics \\
      \revision{Imitation}~\citep{duvallet2013imitation}\\
      \revision{Compact}~\citep{patki2019inferring} \\ \revision{Multimodal}~\citep{arkin2020multimodal}\\ \revision{LIFGIF}~\citep{raychaudhuri2025zeroshotobjectcentricinstructionfollowing} \cellsemantics \\
      \revision{ASHiTA}~\citep{chang2025ashitaautomaticscenegroundedhierarchical}\\
      }
      & 
      \multirow{2}{*}{
      \makecell{
        \revision{Towards semantic maps}~\citep{nuchter2008towards} \cellsemantics \\
        \revision{3D object-class map}~\citep{stuckler2012semantic} \cellsemantics \\
        \revision{Parsing}~\citep{triebel2012parsing} \cellsemantics \\
        \revision{Street Scenes}~\citep{floros2012joint} \cellsemantics \\
        \revision{Dense 3D}~\citep{hermans2014dense} \cellsemantics \\
        \revision{SemanticFusion}~\citep{mccormac2017semanticfusion} \cellsemantics \\
      }}
    \\
    \cmidrule{2-3}
    & \multicolumn{2}{c|}{
        \makecell{
        \revision{Following directions}~\citep{matuszek2010following}\\ \revision{Conceptual}~\citep{zender2008conceptual} \\
        \revision{SemanticGraph}~\citep{hemachandra2014learning}\\
        \revision{Learning semantic maps}~\citep{walter2014framework} \\
        \revision{Learning models}~\citep{hemachandra2015learning} \\
        \revision{Inferring maps}~\citep{duvallet2016inferring} \\
BEVBert~\citep{an2023bevbert} \celloccupancy\cellsemantics  \\
\revision{LaneSegNet}~\citep{li2024lanesegnet} \\
        }
      } & \\
      \cmidrule{2-4}
      & & \multicolumn{2}{c|}{
        \makecell{
        \revision{Contextually guided}~\citep{anand2013contextually} \cellsemantics  \\
        \revision{3DSG}~\citep{armeni20193d} \\
        }
      } \\
      \cmidrule{2-4}
      & \multicolumn{3}{c|}{
        \makecell{
3D-DSG~\citep{Rosinol20rss-dynamicSceneGraphs} \celloccupancy\cellsemantics \\
\revision{Hydra}~\citep{hughes2022hydra} \celloccupancy\cellsemantics \\
        \revision{S-graphs}~\citep{shen2019situational} \celloccupancy\cellsemantics \\
        \revision{S-graphs+}~\citep{bavle2023s} \celloccupancy\cellsemantics \\
        \revision{CURB-SG}~\citep{greve2024collaborative} \celloccupancy\cellsemantics \\
        }
      } \\
      \midrule
    Implicit to Explicit &
      \makecell{
SemanticMapNet~\citep{cartillier2020semantic} 
\cellsemantics
      } &\multicolumn{2}{c|}{
        \makecell{
        \revision{SceneGraphFusion}~\citep{wu2021scenegraphfusion}  \\
        }
      } \\
    \midrule
    Implicit (V) & 
      \makecell{
        CMP~\citep{gupta2017cognitive} \\ 
        MapNet~\citep{henriques2018mapnet} \\
        MultiON~\citep{wani2020multion}\\
        \revision{RNR-Map}~\citep{kwon2023renderable}\\
      }
      & 
      \makecell{
        SPTM~\citep{savinov2018semi} \\ 
        NTS~\citep{chaplot2020neural} \\
        CMTP~\citep{chen2021topological} \\
        VGM~\citep{kwon2021visual} \cellvisittime 
      } 
      &
      \makecell{
      \revision{KPConv~\citep{thomas2019kpconv}} \\
      \revision{PointResNet}~\citep{desai2022pointresnet} \\
      \revision{Neural Fusion}~\citep{mazur2023feature} \\
      }
      \\ 
    \midrule
    \multirow{3}{*}{Implicit (VL)} & 
      \makecell{
        CoW~\citep{gadre2023cows} \\ VLMap~\citep{huang2023visual} \\
        \revision{Le-RNR-Map}~\citep{taioli2023language}\\
        NLMap~\citep{chen2023open} \\ VLFM~\citep{yokoyama2023vlfm} \\ 
        \revision{VoxPoser}~\citep{huang2023voxposer}\\
        InstructNav~\citep{longinstructnav} \\
        \revision{OVL-MAP}~\citep{wen2025ovl}\\
      } & 
      \makecell{
      \revision{LM-Nav}~\citep{shah2022lmnav}\\
      RoboHop~\citep{garg2024robohop}\\
      \revision{Clio}~\citep{maggio2024clio}\\
      }
      & 
      \makecell{
        OpenScene~\citep{Peng2023OpenScene} \\
        CLIPFields~\citep{shafiullah2023clip} \\
        \revision{ConceptFusion}~\citep{conceptfusion}\\
        CLIP2Scene~\citep{chen2023clip2scene} \\
        3D aware ObjNav~\citep{zhang20233d}\\
      } \\
      \cmidrule{3-4}
      & & \multicolumn{2}{c|}{
        \makecell{
            ConceptGraphs~\citep{conceptgraphs}\\
            \revision{HOV-SG}~\citep{Werby-RSS-24} \\
        }
      } \\
      \cmidrule{2-4}
      & \multicolumn{3}{c|}{
        \makecell{
StructNav~\citep{chen2023structnav}  
        }
      } \\
    \bottomrule
    \end{tabular}
    }
\end{table*}

% In addition to studying how to build intelligent agents, research in robotics has to consider various low-level aspects (low-level path planning and control, hardware sensors, robot hardware, etc.).  In contrast, embodied AI research can focus more on high-level task planning by abstracting out the low-level details. This has led embodied AI researchers to explore map building techniques as part of the high-level task planning and address questions such as `is mapping even necessary'~\citep{partsey2022mapping}, `what should be the structure of the map', `what type of information to store inside the map', and `which type of information is useful for what tasks'. There has also been a recent shift in focus to build general-purpose AI solutions by leveraging foundation models \citep{radford2021learning,oquab2023dinov2,openai2023gpt4} which has allowed the community to explore building general-purpose, open-vocabulary semantic maps independent of the downstream tasks.  These open-vocabulary maps can be later queried using natural language \citep{conceptgraphs,Peng2023OpenScene,chen2023open} or images.
% which are trained on a vast amount of data and transferred to any number of downstream tasks. 
\begin{figure*}[ht]
\begin{center}
\includegraphics[width=\linewidth]{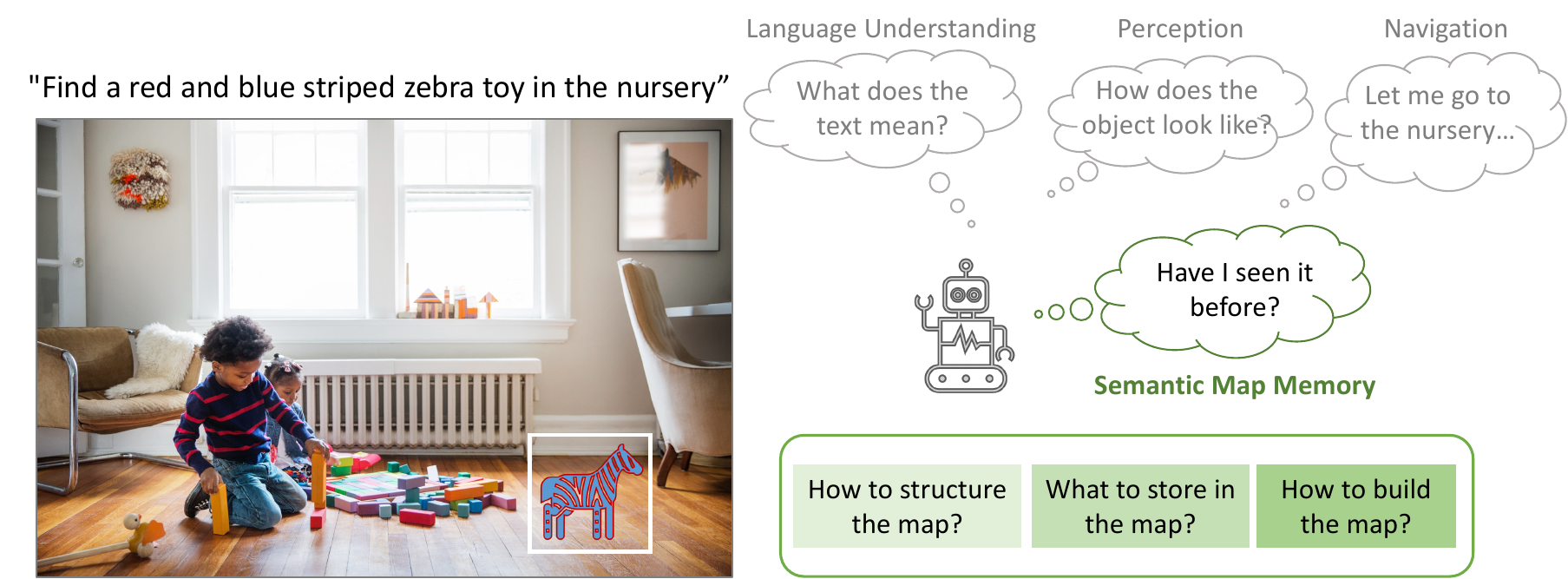}
\end{center}
\vspace{-10pt}
\caption{
% \textbf{Semantic mapping.} 
\textbf{Motivation.} 
To perform a complex task in an indoor environment, the robotic agent must possess multiple skills of language understanding, visual perception, navigation, etc. Among these the most crucial is building and maintaining a semantic map of the environment so that it can come back to it while performing the task.
}
\label{fig:what_is_mapping}
\end{figure*}

\section{Background Reading}
\label{sec:prelims}

\revision{
In this section, we provide background readings that help to contextualize the survey on semantic mapping.
We first introduce key tasks (\Cref{sec:embodied_tasks}) in both robotics (\Cref{sec:robotics_tasks}) and  embodied AI (\Cref{sec:embodied_ai_tasks}), since semantic mapping is often studied in conjunction with such tasks. In \Cref{sec:slam}, we discuss the basics (\Cref{sec:slam_basics}) and techniques of SLAM (\Cref{sec:slam_techniques}) as well as the historic evolution of Semantic SLAM (\Cref{sec:semantic_slam}), which has an overlap with the map building techniques that we later discuss throughout the survey. Next we briefly discuss two popular system design choices in \Cref{sec:system_design}, end-to-end learning (\Cref{sec:end-to-end}) and modular pipeline (\Cref{sec:modular_methods}), since it helps the readers understand how semantic map construction varies between end-to-end and modular approaches.
}

% In this section we provide a brief overview of different types of embodied AI tasks (\Cref{sec:embodied_tasks}) and various approaches towards solving them, including end-to-end (\Cref{sec:end-to-end}) and modular (\Cref{sec:modular_methods}) approaches. 
% As a full survey on Embodied AI tasks is beyond the scope of this survey, please see \citep{deitke2022retrospectives} for a more detailed survey on the tasks and their current state of research.
% In \Cref{sec:slam}, we  discuss classical SLAM based techniques. This discussion aims to provide a background to the readers about the general advance in the embodied AI research and its connection to traditional SLAM-based techniques before we dive in to discuss semantic mapping. 

% \input{chapters/classical_approaches}
% \IEEEpubidadjcol

\subsection{\revision{Embodied tasks}}
\label{sec:embodied_tasks}
\revision{
Embodied tasks involve an intelligent agent, either a physical robot or a simulated body, perceiving and interacting with the environment through its embodiment (sensors, actuators etc.). These tasks require the agent to not only understand the world (via vision, language, etc.) but also take meaningful actions within it (such as navigation or object manipulation). In this section, we provide a brief overview of the embodied tasks explored in both robotics and embodied AI, highlighting their evolution as a growing research direction. This serves to contextualize our focus on semantic mapping, which is often studied in conjunction with such embodied tasks.
}

\subsubsection{\revision{Robotics tasks}}
\label{sec:robotics_tasks}
\revision{
The era of modern robotics began with Unimate~\citep{detesan2024unimate}, the first industrial manipulator arm and Shakey~\citep{center1984shakey}, the first mobile robot.
Since then the field has evolved to have increased autonomy and complex reasoning skills.
This has been enabled by research in various areas spanning basic navigation and obstacle avoidance to complex real-world capabilities involving perception, mapping, and manipulation. Early research focus on collision avoidance~\revisedcitep{Fox1997TheDW}, Monte Carlo Localization~\revisedcitep{Thrun2001RobustMC} and SLAM frameworks enabling robots to localize while mapping unknown environments~\revisedcitep{thrun2002probabilistic,thrun2006graph,taheri2021slam}. As sensors have improved over the years, semantic mapping gained traction, integrating object recognition and scene understanding into spatial maps~\revisedcitep{salas2013slam++}. Concurrently, robotic manipulation has become a central focus, with tasks such as pick-and-place, insertion, and rearrangement. Classical approaches like force-control-based manipulation~\revisedcitep{10.1115/1.3139652,Yoshikawa1985ManipulabilityOR} have been complemented by modern reinforcement learning-based systems like QT-Opt~\revisedcitep{Kalashnikov2018QTOptSD}  and Transporter Networks~\revisedcitep{Zeng2020TransporterNR}. The field has also expanded to multi-task and long-horizon planning, where agents are required to complete sequences of interconnected tasks rather than isolated actions. A prominent framework in this area is Task and Motion Planning (TAMP)~\revisedcitep{Garrett2020IntegratedTA}, which integrates high-level symbolic reasoning (e.g., planning to "pick up the cup, open the door, and place the cup on the table") with low-level continuous motion planning (e.g., calculating joint angles and trajectories to execute those actions). This enables agents to operate in complex and dynamic environments. Recent trends include uncertainty-aware planning~\revisedcitep{Stachniss2005InformationGE,Blanco2008ANM,georgakis2022uncertainty,Carlone2013ActiveSA,Pan2019RiskAR}, semantic mapping~\revisedcitep{rosinol2020kimera} and task planning in dynamic environments~\revisedcitep{Rosinol20rss-dynamicSceneGraphs}. Another fast growing field in robotics is autonomous driving~\revisedcitep{guan2024world,zhao2024bev}, where recent trends increasingly rely on Bird’s Eye View (BEV) representations, which transform multi-view sensor inputs into a unified top-down map using end-to-end learned models. BEV is a common intermediate representation used to simplify reasoning about spatial layouts, obstacles, lanes and other semantic elements. Transformer-based architectures like BEVFormer~\revisedcitep{li2024bevformer} and BEVFusion~\revisedcitep{liu2023bevfusion} have emerged as state-of-the-art, enabling spatially consistent, semantic-rich BEV maps that support tasks such as detection, planning and trajectory prediction. For example, LaneSegNet~\revisedcitep{li2024lanesegnet} trains an end-to-end system to predict lane segments from multi-view surrounding images. 
Unlike semantic maps that are discussed throughout this survey, BEV maps are typically short-lived local (per-frame) representations and specialized for driving scenarios. 
Overall, the advances in robotics collectively demonstrate a shift from reactive, single-task robots toward robust, general-purpose systems capable of operating autonomously in dynamic real-world environments.
}

\subsubsection{Embodied AI tasks}
\label{sec:embodied_ai_tasks}
%In this section we summarize various Embodied AI tasks.
% (\Cref{fig:embodied_ai_task}). 
% \input{chapters/figures/fig-embodied-ai}
\revision{
The field of embodied AI began to emerge prominently around 2017–2018, catalyzed by the availability of simulation environments, such as AI2-THOR~\citep{Kolve2017AI2THORAI}, Habitat~\citep{savva2019habitat}, etc., and benchmarks such as Vision-and-Language Navigation (VLN)~\citep{anderson2018vision}, Embodied Question Answering (EQA)~\citep{eqa_matterport}, etc. Historically, embodied AI diverged from robotics by focusing more heavily on learning-based agents interacting with simulated environments using vision, language, and action, often without relying on physical hardware. The divergence was motivated by the scalability and reproducibility challenges in real-world robotics, and the need to study cognition, planning, and multi-modal learning at scale. Broadly speaking, embodied AI positions itself at the intersection of computer vision, natural language processing and reinforcement learning, while classical robotics focuses on control, perception and physical interaction with the real world.
}

Embodied AI tasks vary depending on the type interaction of an agent with its environment.  Broadly, we can group embodied tasks into three groups -- \textit{Exploration} task~\citep{chaplot2020learning} requires an agent to efficiently explore its environments; \textit{Navigation} task~\citep{wijmans2019dd,batra2020objectnav} requires the agent to take actions in order to move around the environment; \textit{Manipulation} task~\citep{szot2021habitat,weihs2021visual} requires the agent to perform interactive actions to change the state of other objects in the environment.
The taxonomy of tasks can be further broken down by 
the target specification provided to the agent, which impacts the information that need to be retained. For instance, for navigation, the following are commonly studied.
In \textit{Point-Goal Navigation} \citep{wijmans2019dd} (PointNav), the agent is given a target coordinate relative to its starting position, whereas in \textit{Image-Goal Navigation} (ImageNav) it is given a target image \citep{chaplot2020neural}.
In the \textit{Object-Goal Navigation} (ObjectNav) task, the agent needs to navigate to any instance of an object category \citep{habitatchallenge2022}. An extension to the ObjectNav task is the \textit{Multi-Object Navigation} (MultiON)~\citep{wani2020multion} task where the agent is required to navigate to multiple objects in a particular sequence.
\textit{Vision-and-Language Navigation} (VLN)~\citep{anderson2018vision} requires the agent to find the target as specified by a natural language instruction. 
In \textit{Audio-Visual Navigation} task, the agent needs to navigate to an object emitting a particular sound in an indoor environment \citep{chen2020soundspaces,gan2019look}.  Depending on the type of task, it may be sufficient to store just the object category (e.g. for ObjectNav) in the map, or it is may be necessarily to retain more finer-grained information (e.g. VLN). In this survey, we will mainly focus on recent work on room-scale map building for navigation as these methods can be extended for maps for manipulation and used for exploration. 

\subsection{\revision{Simultaneous Localization and Mapping (SLAM)}}
\label{sec:slam}
\revision{
Although we briefly discuss robotics applications in \Cref{sec:embodied_tasks}, we address Simultaneous Localization and Mapping (SLAM)~\citep{thrun1998probabilistic,thrun2000real,ferris2007wifi,huang2011efficient,sh-p1-prelude} separately here due to its close connection with semantic mapping, particularly the Semantic SLAM literature. While both aim to enrich spatial representations with semantic information, they differ in scope and purpose. Rooted in robotics, Semantic SLAM focuses on building globally consistent, pose-aware maps augmented with semantics to improve localization, mapping accuracy and long-term robustness. In contrast, semantic mapping in embodied AI emphasizes representations that support high-level reasoning and decision-making, often abstracting away precise localization and low-level sensor noise in favor of adaptability.
}
% This is similar to robotics settings where landmark locations are known apriori, for example, when GPS is available or a prebuilt map exists, making SLAM unnecessary.

\subsubsection{\revision{Basics of SLAM}}
\label{sec:slam_basics}
\revision{SLAM (Simultaneous Localization and Mapping) enables a robot to leverage data from multiple sensors (cameras, LiDAR, IMU etc.) to perceive the environment and build a map while simultaneously localizing itself on the map.
Mathematically speaking, SLAM aims to estimate the robot’s trajectory and a map of the environment simultaneously, using noisy sensor data and imperfect motion estimates. 
A robot’s motion is modeled by a state transition function that predicts its next pose based on the previous pose and control inputs:
\begin{equation}
\label{eq:slam_motion_model}
    x_t = f(x_{t-1}, u_t) + w_t
\end{equation}
where, $x_t$ is the robot’s state at time $t$, $u_t$ is the control input (such as wheel odometry or IMU), and $w_t$ is process noise. Simultaneously, the robot makes observations of the landmarks in the environment, which is modeled by the measurement equation:
\begin{equation}
\label{eq:slam_obs_model}
    z_t^i = h(x_t, m_i) + v_t
\end{equation}
where $z_t^i$ is the observation of landmark $i$ at time $t$, $m_i$ is the position of the landmark and $v_t$ is the measurement noise. The difference between the predicted and the actual observations is the residual (error):
\begin{equation}
\label{eq:slam_residual}
    e_t(x) = z_t^i -  h(x_t, m_i)
\end{equation}
SLAM formulates the estimation problem as a nonlinear least-squares optimization, which aims at finding the optimal trajectory $x^*$ and landmark positions $m^*$ that best explain all measurements:
\begin{equation}
\label{eq:slam_opt}
    x^*, m^* = arg~\min_{x,m} \sum_t \sum_i e_t^T(x)~ \Omega_t e_t(x)
\end{equation}
where $\Omega_t$ is the information matrix (inverse covariance) that weights the measurements according to their uncertainty, i.e. more weight is given to a measurement that is very certain. $e_t^T(x)$ is the transpose of the error $e_t(x)$ and the term $e_t^T(x)~ \Omega_t e_t(x)$ represents weighted squared error.
In graph-based SLAM, this optimization corresponds to optimizing a factor graph~\citep{loeliger2004introduction, dellaert2017factor}, where nodes represent robot poses and edges represent spatial constraints, such that \Cref{eq:slam_opt} becomes:
\begin{equation}
\label{eq:slam_graph_opt}
    x^* = arg~\min_x \sum_{(i,j) \in \varepsilon} ||z_{ij}-\hat{z}_{ij} (x_i, x_j)||^2_{\Omega_{ij}}
\end{equation}
where, $(i,j) \in \varepsilon$ represent edges in the pose graph, $z_{ij}$ is the relative pose measurement between nodes $i$ and $j$
, $\hat{z}_{ij}$ is the predicted relative measurement from current estimates, and $||~.~||^2_{\Omega_{ij}}$ is the Mahalanobis distance (weighted squared error).
This global optimization ensures that the estimated map and trajectory are as consistent as possible with the entire history of noisy sensor data.}

\subsubsection{\revision{SLAM Techniques}}
\label{sec:slam_techniques}
\begin{figure}[ht]
\centering
\subfloat{\includegraphics[width=0.4\linewidth]{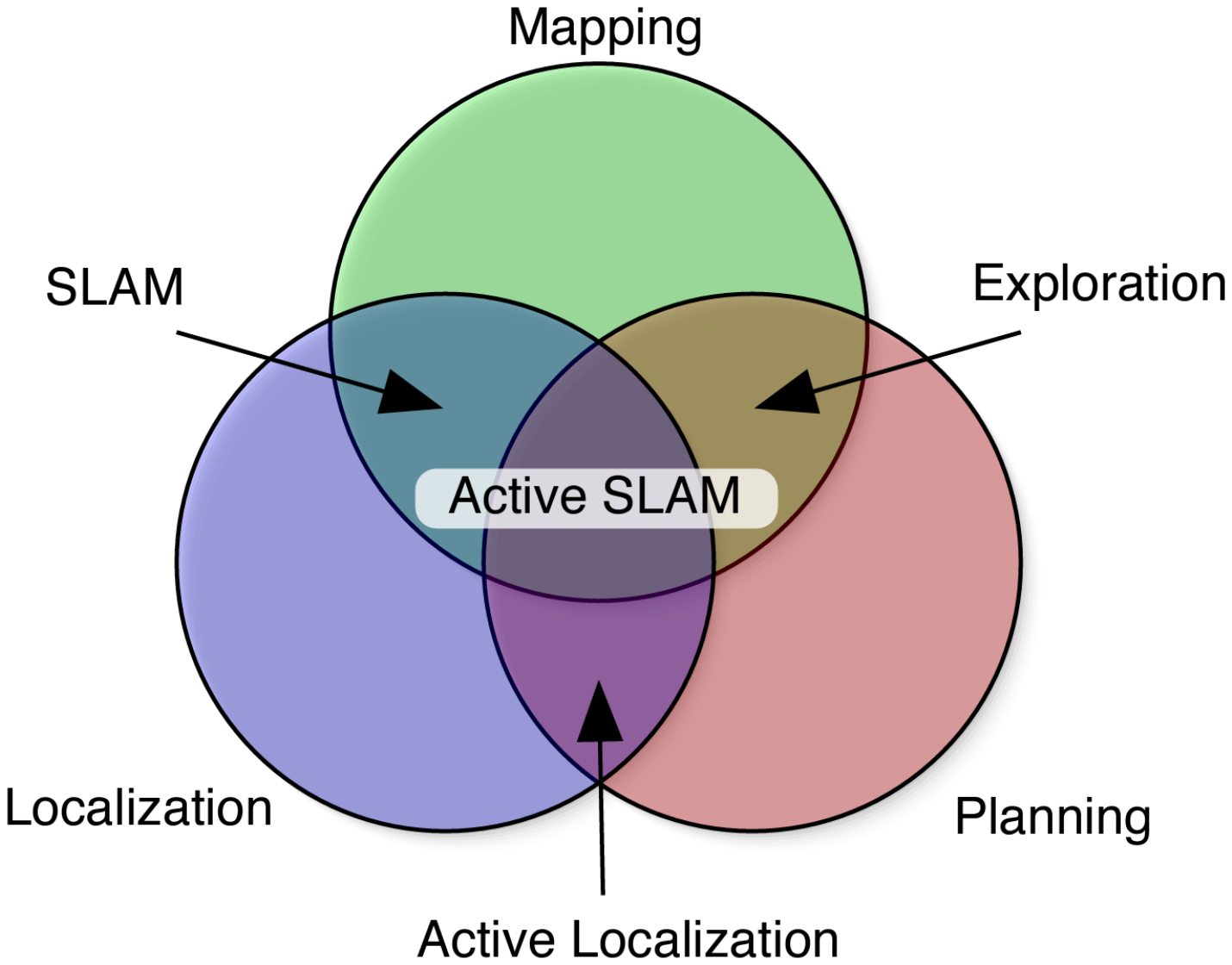}}
     \subfloat{\includegraphics[width=0.5\linewidth]{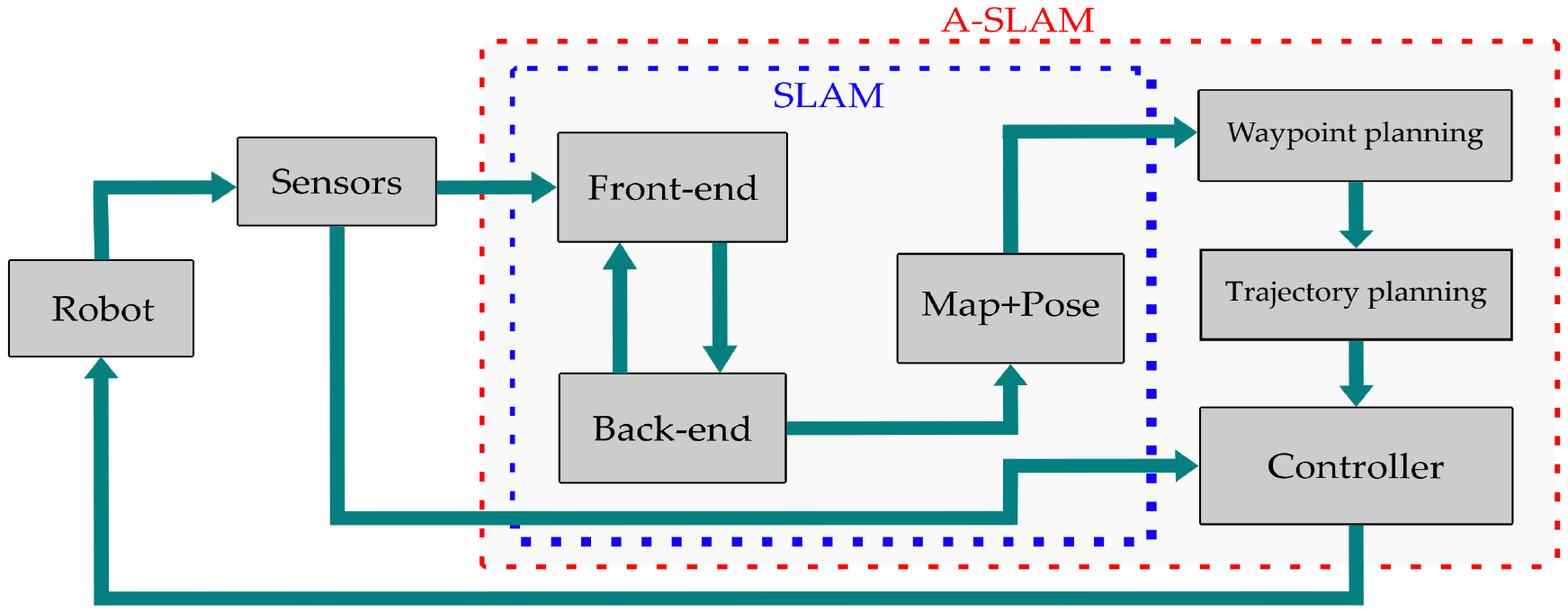}}
\caption{
\textbf{SLAM.} (left) At the core of classical mobile robotics lie three core tasks -- mapping, localization, and planning. These are often interdependent on each other and overlap to form other tasks such as SLAM (localization and mapping), exploration (mapping and planning), active localization (planning and localization) and active SLAM (mapping, localization, and planning)
[\textit{figure reproduced from \cite{fairfield2009localization}].}
\revision{(right) Architecture of SLAM includes the front-end to handle perception tasks and the back-end to estimate pose and the map. Active SLAM (A-SLAM) additionally includes planning and controller modules [\textit{Figure reproduced from \cite{ahmed2023active}].}
}}
\label{fig:active_slam}
\end{figure}

\revision{Active SLAM~\citep{ahmed2023active} enables an autonomous agent to actively choose its actions to improve its map and localization, rather than passively mapping the environment as it moves (\Cref{fig:active_slam}-left). The SLAM system~\citep{cadena2016simultaneous,pu2023visual} primarily consists of a front-end module and a back-end module (\Cref{fig:active_slam}-right). 
The front-end module computes feature extraction, data association and feature classification on the sensor observations. Moreover it applies algorithms such as Iterative Closest Point (ICP)~\citep{he2017iterative} that aligns sensor observations from subsequent frames into consistent 3D geometry as well as loop closure~\citep{tsintotas2022revisiting} that recognizes whether
the current observation matches with a previously visited area.
The backend module, on the other hand, is responsible for pose optimization and map estimation based on the data from the front-end. The back-end also provides feedback to the front-end for loop closure and verification.
While some SLAM methods use LiDARs~\citep{khan2021comparative}, others use camera as the primary sensor (visual SLAM or vSLAM), due to the availability of less expensive cameras and the advances in computer vision.
Visual SLAM methods often extract and match geometric features (points, lines, or planes)~\citep{mur2015orb,yang2017direct,kaess2015simultaneous,cai2021improved} from the image, with the help of feature detection algorithms such as SIFT~\citep{lowe2004distinctive}, SURF~\citep{bay2008speeded}, and ORB~\citep{rublee2011orb}. 
Others operate directly on the image pixel intensities~\citep{newcombe2011dtam,engel2014lsd} and thus retain all information about the image.
While these methods use RGB camera as the sensor, RGB-D SLAM~\citep{newcombe2011kinectfusion,kaess2012kintinuous,endres20133,whelan2016elasticfusion} uses RGB-D cameras to simultaneously collect color images as well as depth images.
In contrast, Visual-Inertial SLAM~\citep{mur2017visual,cheng2021improved} uses an additional IMU sensor to mitigate the effects of image blur and poor illumination from using camera sensor alone. 
} 
% While these works form the foundation for SLAM systems, they fail to meet the demands of autonomous navigation and planning.

% Traditional robotics addresses sensor uncertainty, whereas many embodied AI simulators simplify the problem by assuming ideal, noise-free sensors.

\subsubsection{\revision{Semantic SLAM}}
\label{sec:semantic_slam}
\revision{
While geometric maps in SLAM systems prove to be effective for simple navigation and control~\revisedcitep{shan2020lio,campos2021orb}, they provide limited understanding of the environment’s semantic content. 
As robotic systems grew more complex and began interacting with unseen unstructured dynamic environments, the need for higher-level understanding led to the emergence of Semantic SLAM.
It enriches spatial geometric maps with meaningful concepts like objects, rooms, and affordances, thus bridging the gap between perception and task-level reasoning in modern embodied agents. Early approaches do so by matching image features to object model databases~\revisedcitep{civera2011towards,salas2013slam++} or by incorporating semantic and geometric cues into structure-from-motion~\revisedcitep{bao2011semantic,bao2012semantic} and SLAM optimization processes~\revisedcitep{fioraio2013joint}.
With advances in deep learning, recent works on Semantic SLAM~\revisedcitep{mccormac2017semanticfusion,yin2020fusionlane} have improved semantic representation, real-time performances and dynamic environment modeling.
Some works show that using semantic maps improves the performance of the various modules in a SLAM system~\revisedcitep{xiang2017rnn,tateno2017cnn,qin2021semantic,qian2021semantic}, while others show that using SLAM to impose consistency constraints improves semantic segmentation~\revisedcitep{mozos2007supervised,lai2014unsupervised,pronobis2012large,pillai2015monocular,cadena2015fast}. Many also show that combining SLAM and semantic segmentation into joint frameworks enhances both mapping and object recognition~\revisedcitep{flint2011manhattan,bao2012semantic,hane2013joint,kundu2014joint,sengupta2015semantic,vineet2015incremental}.
}

\revision{
Moreover, semantic SLAM works can be categorized based on the structure and encoding of the semantic maps, in the same way as semantic mapping in embodied AI. For example, some works use sparse representations~\revisedcitep{bowman2017probabilistic,yang2019cubeslam,nicholson2018quadricslam,chen2020sloam,shan2020orcvio,atanasov2018unifying,feng2019localization,salas2013slam++}, while others use dense representations~\revisedcitep{rosinol2021kimera,grinvald2019volumetric,mccormac2018fusion++,chen2019suma++,maturana2018real,miller2021any,miller2022stronger}. Some works store explicit semantic concepts such as rooms~\revisedcitep{pronobis2012large} or objects~\revisedcitep{pillai2015monocular} in the map, while others store implicit neural features~\revisedcitep{xiang2017rnn,tateno2017cnn,qin2021semantic,shah2022lmnav,maggio2024clio,Werby-RSS-24}. Throughout this survey, we will revisit these papers to categorize them based on the categories we use for semantic mapping in embodied AI.}

\revision{For a more comprehensive reading on SLAM and semantic SLAM, we refer the readers to the following survey papers -- \revisedcite{Thrun2003RoboticMA,kostavelis2015semantic,cadena2017past,younes2017keyframe,taketomi2017visual,landsiedel2017review,saputra2018visual,sualeh2019simultaneous,chen2020survey,lluvia2021active,taheri2021slam,tourani2022visual,pu2023visual,racinskis2023constructing,sousa2023systematic,wang2024survey,chen2025semantic}.}

\subsection{\revision{System design strategies}}
\label{sec:system_design}
\revision{Designing systems for embodied agents involves a fundamental architectural choice between end-to-end learning and modular pipelines. While end-to-end approaches map raw sensory input directly to actions using a single neural network, modular systems decompose the task into interpretable components. Understanding this distinction is essential for situating semantic mapping within the broader system design, as it influences how maps are constructed, represented and utilized. This section briefly outlines these paradigms (\Cref{sec:end-to-end} and \Cref{sec:modular_methods}) and discuss trade-offs between them in \Cref{sec:design_tradeoffs}, to provide a background for the mapping methods discussed in the remainder of this survey.}

\begin{figure}[t]
\begin{center}
% \framebox[0.6\linewidth]{
% \fbox{\rule[-.5cm]{0cm}{4cm} \rule[-.5cm]{4cm}{0cm}
\includegraphics[width=0.7\linewidth]{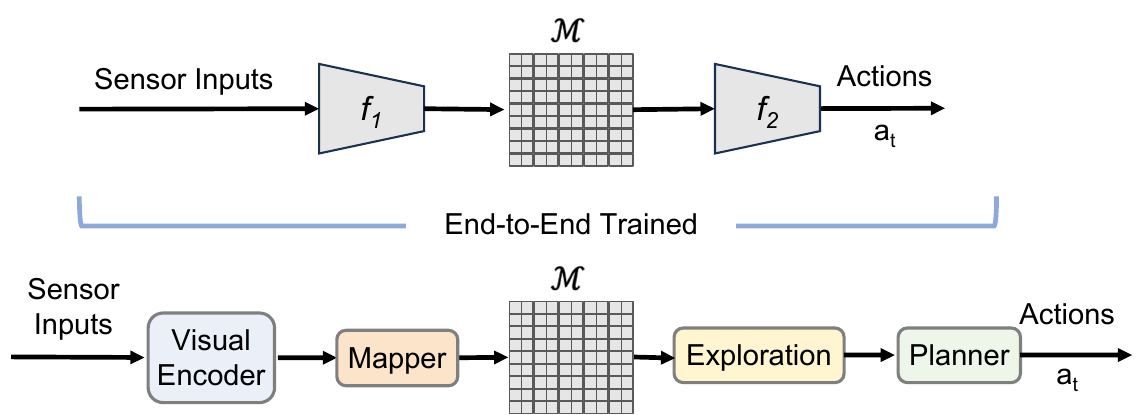}
% }
\end{center}
\vspace{-10pt}
\caption{
\textbf{End-to-end vs Modular.} 
(Top) End-to-end model is trained as a single pipeline which generates actions directly from sensory inputs.
(Bottom) Modular pipeline consists of various sub-modules, each with a specific function so that they can be trained independently of the others. 
% For example, the \textit{Mapper} is responsible for generating a map whereas, the \textit{Planner} is responsible to generate low-level actions.
% Most modular pipelines consist of a Visual Encoder to extract features which are then used by the Mapper along with the agent pose to produce a map. The Exploration module allows the agent to explore new places in the environment. The Planner generates and executes a plan to transport the agent to the target.
}
\label{fig:end_vs_modular}
\end{figure}

\subsubsection{End-to-end approaches}
\label{sec:end-to-end}
The embodied AI and robotics community has seen a lot of progress in training task-specific end-to-end models with reinforcement learning (RL)~\revisedcitep{tai2017virtual} that directly learns to predict discrete~\citep{wani2020multion} or continuous actions~\citep{kalapos2020sim} from visual observations (see \Cref{fig:end_vs_modular}). 
These methods may consist of an unstructured memory such as LSTM~\revisedcitep{mei2016listen}~\citep{dobrevski2021deep}. However, such representation lacks reasoning
about 3D space and geometry and fail to perform well in long-horizon path planning. This has lead to the development of approaches that build an intermediate map representation. 
Such a map can be implemented using differentiable operations so as to facilitate end-to-end training.
% \citep{gupta2017cognitive} learns to build an intermediate map representation by jointly training the mapper and the planner to predict actions. The mapper and the planner are implemented using differentiable operations and a differentiable value iteration~\cite{tamar2016value} respectively, so as to allow for end-to-end training.
% They use DAGGER~\cite{ross2011reduction} to optimize the network .
\citet{gupta2017cognitive} shows that an egocentric map built this way is beneficial for both PointNav and ObjectNav tasks, whereas \citet{henriques2018mapnet} learns a global map for the task of localization. While these methods are trained using supervised learning, \citet{wani2020multion} use RL to learn to predict actions based on an intermediate global map of the environment to address the complex MultiON task. 
\revision{
In autonomous driving and planning, end-to-end approaches aim to learn a direct mapping from raw sensor inputs (e.g., images, LiDAR) to control outputs or motion trajectories. These methods often leverage transformer-based architectures~\citep{lin2022survey} to jointly model perception, spatial reasoning and decision-making within a single framework. By optimizing the entire pipeline holistically, they promise improved robustness and efficiency, particularly in complex, dynamic environments. Recent work demonstrates that such end-to-end models can learn to implicitly reason about obstacles, goals, and traffic patterns. However challenges remain in interpretability, generalization and safety-critical validation.
}
However, irrespective of the map representation or the mode of training, these approaches need to be retrained every time the task definition changes and even the basic skills required to perform a task need to be learned from scratch.

\subsubsection{Modular pipeline}
\label{sec:modular_methods}

% Inspite of the progress in end-to-end approaches, such methods fail to generalize to other complex tasks and to unseen environments. 
Another line of work explores how to breakdown a complex navigation task into a set of basic skills that the agent needs to acquire. Such skills can then be learned independently of each other so that they can be leveraged across various tasks without the need to be retrained from scratch. \revision{While some use two modules~\revisedcitep{bansal2020combining}, one for high-level decision making and another for low-level planner, others~\citep{chaplot2020learning,chaplot2020object,gervet2023navigating,raychaudhuri2023mopa} further breakdown into sub-modules, such as visual encoder, mapper and exploration in addition to the low-level planner.}
% This has led to various works on modular pipelines \citep{chaplot2020learning,chaplot2020object,gervet2023navigating,raychaudhuri2023mopa}, where each module is responsible for a particular skill and the modules interact with each other to perform the entire task.
% In the modular approach, it is common to have a \textbf{visual encoder} that processes and encodes the visual information at each time step, a \textbf{mapper} that aggregates the encoded information into a map, a \textbf{exploration} module that determine what parts of the environment needs to be explored, and a \textbf{planner} that determines the low-level action to take. 
\Cref{fig:end_vs_modular} shows the difference between the end-to-end approach and the module approach, and we elaborate on the modules and popular design choices below.
%The basic set of modules (see \Cref{fig:end_vs_modular}) that are common in most modular pipelines are described below.

\mypara{Visual Encoder.} This module encodes agent observations to produce semantic visual features and predictions at every time step. Prior works have used visual features from pretrained backbones such as ResNet~\cite{he2016deep} or ViT~\citep{dosovitskiy2020image}, and often leveraging object detectors MaskRCNN~\cite{he2017mask} or FasterRCNN~\cite{ren2015faster}.  As we will see in \Cref{sec:open-vocab}, with the development of large pretrained vision-language models and open-vocabulary detectors, pretrained models such as CLIP~\citep{radford2021learning}, LSeg~\citep{li2022languagedriven}, DINO~\citep{caron2021emerging,oquab2023dinov2}, and others are increasingly popular as the basis for building large to open-vocabulary maps. The visual encoder used will determine the information captured in the features, and whether there are detected object instance bounding boxes or segmentations for integration into the mapping modules.  

\mypara{Mapper.} The mapper is responsible for building a semantic map of the environment from the encoded image features and agent pose. To build a global map over time, the mapper typically aggregates the current map with the map from previous step (see \Cref{sec:semantic-map-building} for details).  In this paper, we survey how recent methods structure the map (\Cref{sec:structure}) and what information can be encoded in it (\Cref{sec:representations}).  
\Cref{tab-modular-app} summarizes the type of information stored in the map by various methods. This can be occupancy information, explored area or semantic labels of the detected objects.

\mypara{Exploration.} This module enables the agent to explore its environment efficiently to either ensure the map is complete (by maximizing the covered area) or selecting unvisited areas where the target is likely to be. Typically, the exploration module selects a point or region to explore given the obstacle map built by the mapper and the current agent pose. Agents can use simple heuristics-based methods such as sampling a point at uniform~\citep{zhang2021novel,raychaudhuri2023mopa}, systematically sampling four corners of a grid centered at the agent~\citep{luo2022stubborn} or selecting a point from the unexplored frontier~\citep{yamauchi1997frontier}. To decide which frontier point the agent should explore, various strategies are employed, such as selecting the nearest point to the agent~\citep{gervet2023navigating} or the most promising point based on semantic reasoning. In the semantic reasoning based exploration methods, the agents may select the highest text-image relevance score~\citep{gadre2023cows,yokoyama2023vlfm} from a pretrained large vision-language model such as BLIP-2~\citep{li2023blip}, select the highest probabilistic output of the VLM directly~\citep{ren2024explore}, or leverage a LLM to extract common-sense knowledge~\citep{zhou2023esc}. 
Researchers have also used learned policies~\citep{chaplot2020learning,chaplot2020object}, where the agents are generally trained with RL using rewards, such as coverage~\citep{chen2019learning} or curiosity~\citep{pathak2017curiosity,mazzaglia2022curiosity}. Although there is less hand-crafted rules in learning-based methods, they need millions of training steps and careful reward engineering.

\mypara{Planner.} Once a map is built, a low level path-planning module is used to plan a path to the goal location from the agent's current location.  The path consists of low-level actions that can be executed by the agent to move to the goal. While this is implemented as a heuristics-based Fast Marching Method~\citep{sethian1996fast} in most of the prior works, a recent approach MOPA by~\citep{raychaudhuri2023mopa} has used a learned PointNav policy trained offline with DD-PPO~\citep{wijmans2019dd}.

\vspace{10pt} 
\noindent In \Cref{tab-modular-app}, we compare common modular approaches and look at how they leverage various heuristics-based or learned approach for each of the modules.
The advantages of a modular pipeline include its ability to leverage pretrained models from other tasks \citep{gervet2023navigating,raychaudhuri2023mopa} and its ability to transfer from simulation to real-world robots better \citep{gervet2023navigating}.

\begin{table}[ht]
\caption{
We show how prior works build map either as part of an end-to-end architecture or as a modular architecture consisting of four basic modules. For the modular approaches, we summarize the different module choices as well as what they store in their map.
}
\label{tab-modular-app}
\centering
    \resizebox{\linewidth}{!}{
    \begin{tabular}{llc|lp{3cm}p{2.5cm}l}
    \toprule
    \multirow{2}{*}{\bf Methods} 
    &\multirow{2}{*}{\bf Task}
    &\multirow{2}{*}{\bf End-to-End} 
    &\multicolumn{4}{c}{\bf Modular} 
    \\ 
    && &\bf Visual Encoder &\bf Map &\bf Exploration &\bf Planner \\
    
    \midrule

    ANS~\citep{chaplot2020learning} &Exploration &&ResNet18~\citep{he2016deep} & occupancy + explored &learned policy &Fast Marching \\

    NTS~\citep{chaplot2020neural} &ImageNav & &ResNet18~\citep{he2016deep} &topological map & learned policy & A* \\
    
    SemExp~\citep{chaplot2020object} &ObjectNav & & MaskR-CNN~\citep{he2017mask} & occupancy + explored + semantic labels & learned policy &Fast Marching %\cite{sethian1996fast} 
    \\
    ModLearn~\citep{gervet2023navigating}&ObjectNav & & MaskR-CNN~\citep{he2017mask} & occupancy + explored + semantic labels & learned SemExp &Fast Marching \\
    
    % \cite{chen2023train} &ObjectNav &MaskR-CNN & semantic pointcloud+2D occupancy map+spatial scene graph &Semantic Frontier policy &Fast Marching \\
    MOPA~\citep{raychaudhuri2023mopa} &MultiON && FasterRCNN~\citep{ren2015faster} & semantic labels & Uniform Sampling Exploration &PointNav\citep{wijmans2019dd} \\

    \midrule

    CMP~\citep{gupta2017cognitive} & PointNav, ObjectNav &\cmark \\

    MapNet~\citep{henriques2018mapnet} & Localization &\cmark &\multicolumn{4}{c}{--} \\
    
    MultiON~\citep{wani2020multion} & MultiON &\cmark\\

     \bottomrule
    \end{tabular}
    }
\end{table}

% The modular approach in \citep{gervet2023navigating} is based on \citep{chaplot2020object} which proposes a semantically-aware exploration method called SemExp 
% by extending on the modular approach for exploration~\cite{chaplot2020learning} where the agent predicts an exploration goal based on the built semantic map of the environment and the target object. SemExp is trained using RL and is shown to be able to learn semantic priors such that the agent explores areas of the environment where the target is most likely to be located.\citep{chen2023train} also proposes a modular approach towards ObjectNav on Gibson~\cite{GIBSONENV} scenes.

% Some of the recent state-of-the-art methods in ObjectNav are trained end-to-end using Imitation Learning (IL)~\cite{ramrakhya2022habitat}, two stages of training with IL followed by RL finetuning~\cite{yadav2022OVRL, ramrakhya2023pirlnav} and by using advanced architecture of vision transformers~\cite{yadav2023ovrl}. However, \citep{gervet2023navigating} shows that a modular approach transfers better from simulation to real-world robots than end-to-end trained methods for ObjectNav. In this section we do a deep dive on the commonalities of different modular approaches and their limitations.

\subsubsection{\revision{Trade-offs}}
\label{sec:design_tradeoffs}
\revision{
Both the design choices have their own merits and limitations. End-to-end systems are simple to train and deploy, and they directly optimize task performance. However, they often lack interpretability, generalize poorly to new scenarios, and make component reuse difficult. In contrast, modular pipelines offer greater transparency, reusability, and flexibility to combine learned and classical methods, making them well-suited for complex tasks involving long-horizon planning and semantic reasoning, despite the potential for error propagation and suboptimal integration of modules.
}

\section{Semantic map}
\label{sec:semantic_map}
\revision{In this section, we introduce semantic maps (\Cref{sec:semantic-map-overview}), and provide an overview of their structural representations  (\Cref{sec:semantic-map-structure}), semantic encodings (\Cref{sec:semantic-map-encoding}), and the process of building them (\Cref{sec:semantic-map-building}).}

\subsection{What are semantic maps?} 
\label{sec:semantic-map-overview}
While maps, capturing the geometry of the space, can help an agent avoid obstacles, they are not sufficient for the demands of more complex reasoning tasks. An enhanced map that goes beyond geometry to capture meaning and context in its environment aligns with how humans perceive and navigate its surroundings. 
These are \emph{semantic maps}, which provide a richer and more nuanced understanding about the objects and places in the environment and are indispensable for performing complex tasks such as navigating to a specific room (kitchen)~\citep{narasimhan2020seeing}, rearranging objects~\citep{trabucco2022simple} or performing an action on a specific object (sitting on a couch)~\citep{Peng2023OpenScene}.
% Semantic maps can also be structured in different ways that goes beyond just grids.
% A semantic map captures not just the physical space of the environment but also semantic information about the environment, such as names of identified objects and regions, key features and other attributes relevant to how the agent navigates or interacts with the environment. It can also store spatial and functional relationships between the objects and regions.
% %and thus goes beyond traditional geometric maps storing obstacle information.
A physical or a virtual agent perceives the environment through sensors (camera, LiDAR, etc.), uses cognition to classify the objects and regions it perceives~\citep{ren2015faster,he2017mask,liu2024grounding,kirillov2023segment,zhang2023recognize}, stores them in a structured semantic map which can be queried for efficient reasoning. 
% Semantic maps enable the robot to reason about the environment so that it can efficiently interact with the environment in downstream tasks such as navigation, instruction-following and object manipulation. 
% Suppose an agent is tasked with finding an apple and putting it in the refrigerator. Let's say that the agent has seen the refrigerator first and then the apple. After picking up the apple, it would be efficient for the agent to retrace its steps, if it has memorized the location where it had seen the refrigerator.

% \todo{The following paragraph is somewhat redundant with the above, try to merge and trim, maybe can even be removed}
% To build a semantic map, a robot first perceives its environment through its sensors. 
% Advanced computer vision and deep learning techniques \citep{ren2015faster,he2017mask,liu2023grounding,kirillov2023segment,zhang2023recognize} are then used to identify and classify objects (door, bed, refrigerator, etc.) and regions (kitchen, office, bedroom etc.). This semantic information is then integrated into a spatial structure to create a multidimensional semantic map. Moreover, this map is dynamically updated as the robot learns from new experiences and receives new sensory data. This allows for accumulation of semantic knowledge over a longer horizon and enable the robot to perform complex tasks.

\subsection{What is the structure of this map?}
\label{sec:semantic-map-structure}
A semantic map can be structured as a \textit{\spatial}, \textit{\topological}, \textit{\revision{dense geometric}} or a \textit{\hybrid} (\Cref{fig:mapping_overview}). A \spatial is a top-down grid where each grid cell represents an area in the physical environment. So if an object is at a certain location $(X,Y,Z)$ in a 3D scene, the semantic map will contain information about that object at the corresponding grid cell $(x,y)$, where $x$ and $y$ are the row and column numbers respectively such that there is a direct mapping from $(X,Y,Z)$ to $(x,y)$. For navigation, most spatial maps are 2D, such that a grid cell ignores the $Z$-axis (up direction)  by aggregating the semantic information across the up axis. However they can also be 3D where the $Z$-axis is divided into discrete bins.
% $\forall x \in N_r$ and $\forall y \in N_c$ where $N_r$ and $N_c$ are the total number of rows and columns in the grid respectively.
On the other hand, a \topological is a graph-like structure where nodes represent objects or important landmarks in the scene and edges represent relationship (distance, spatial relation, etc.) between them.
It is also possible to store semantic information on a \revision{dense geometric}, which can be viewed as a 3D map with varying density.  In a \revision{dense geometric}, information is associated with each point $(x,y,z)$ corresponding to 3D location $(X,Y,Z)$ in the physical space.  Unlike the voxel-grid, which is regularly spaced, points can be sampled at varying densities.
Some works combine two or more of the above structures to form a \hybrid since each structure has its own advantages and limitations. We discuss each of these in detail in \Cref{sec:structure}.

\subsection{What encoding is stored in this map?}
\label{sec:semantic-map-encoding}
The semantic map stores information about a particular 3D location $(X,Y,Z)$ in the physical environment. This information can either be \textit{\explicit} or \textit{\implicit}. Explicit encodings have clear specific meanings assigned to each value.  For instance, each cell $(X,Y,Z)$ can store information about whether there are any obstacles at that position, whether that location has been explored by the agent, the category of the object present there and so on. 
On the other hand, an implicit encoding is a feature encoding capturing information derived from the sensory input (e.g. images) that the agent observes at that particular location $(X,Y,Z)$. The features are typically extracted from pre-trained encoders. Depending on whether the feature encoder was pre-trained on a set of images from limited categories or a large internet-scale dataset of image and language data, the implicit encoding can be either \textit{closed-vocabulary} or \textit{open-vocabulary}. The term \textit{closed-vocabulary} is used to indicate only a limited set of object categories is recognized, while in a \textit{open-vocabulary} setting, the features extractors can theoretically identify `any' object.

\begin{figure}[t]
\centering
% \framebox[0.9\linewidth]{
% \fbox{\rule[-.5cm]{0cm}{4cm} \rule[-.5cm]{4cm}{0cm}
\includegraphics[width=0.6\linewidth]{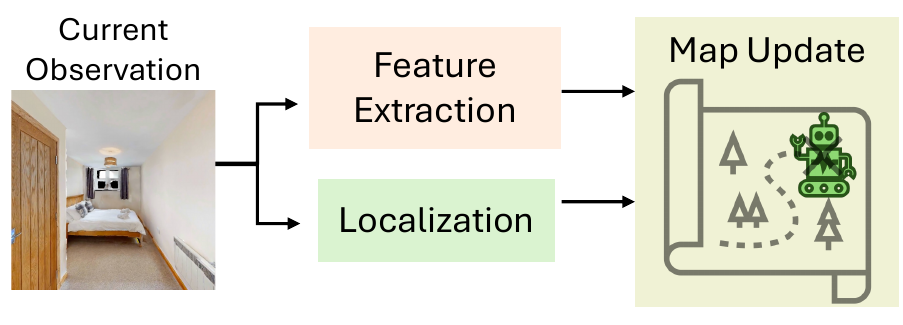}
% }
\caption{
Map building involves \textit{localization} (where the agent is on the map), \textit{feature extraction} (extracting useful semantic information from the observations), 
% \textit{projection} (projecting the features on to the map) 
and \textit{map update} (building the map by aggregating the semantic information over time).
}
\label{fig:map_building_steps}
\end{figure}

\subsection{How is the map built?}
\label{sec:semantic-map-building}
Creating accurate and detailed semantic maps requires integrating data from various sources and sensors such as camera, LiDAR and depth sensors.
More specifically, map building consists of having an agent navigate about a space, and accumulating observations $O_t$ at time step $t$ into the appropriate map structure $m_t$.
To build an accurate map, the agent first needs to have an estimate of where it is (\emph{localization}). Next it extracts semantic information from an observation $F(O_t)$ (\emph{feature extraction}), 
% projects the features onto the map (\emph{projection}), 
and combines the features into a common map over time (\emph{accumulation}) (refer to \Cref{fig:map_building_steps}). 
While building a spatial grid map, an additional step is to project the features onto the map (\emph{projection}).
It is common to group the last three steps into \emph{map building} and study it jointly with \emph{localization} in Simultaneous Localization and Mapping (SLAM). We discuss SLAM methods briefly in \Cref{sec:slam}. 

\mypara{Localization.} 
Localization can be challenging due to noisy sensors and actuators.  To simplify the problem, it is common in the embodied AI community to either assume perfect localization is given at each time step~\citep{cartillier2020semantic} or to localize the agent with respect to its starting position in an episode~\citep{henriques2018mapnet} assuming perfect actuation. The latter is more easily adapted to the real-world setting since it doesn't require the exact knowledge about the agent's pose. Instead the relative displacement of the agent with respect to its starting pose is enough to build the map eventually.

\mypara{Feature extraction.}
Feature extraction is a crucial part of building a semantic map. Ideally these features should be representative of the objects present in the map. We discuss this topic at length in section \Cref{sec:representations}.

\mypara{Projection.} 
%A popular map building technique~\citep{cartillier2020semantic} to build spatial grid maps 
An important step in building a spatial grid map is taking the 2D observations and project them into 3D.   Typically, this relies on having depth information and known camera parameters in order to 
convert 2D pixel coordinates to 3D world coordinates. 
%and then to 2D top-down grid coordinates. 
%This method is based on the well known projective camera geometry from computer vision.
To project a particular pixel in the camera frame, first a ray is shot from the camera center through the image pixel $(i,j)$ to the depth $d_{i,j}$ to get a 3D point in the camera coordinate frame. Next the camera coordinates are converted to the world coordinates $(X,Y,Z)$. For a 2D spatial map, the 3D coordinate$(X,Y,Z)$ is mapped to the grid cell indices $x$ and $y$ in the spatial map.
The transformation for the standard pinhole camera with known camera pose (3D rotation $R$ and 3D translation $t$) and intrinsics ($K$) can be written as: 
\begin{gather}
 \begin{bmatrix} X \\ Y \\ Z \end{bmatrix}
 = d_{i,j} R^{-1} K^{-1} \begin{bmatrix} i \\ j \\ 1 \end{bmatrix} - t
\end{gather}
and the orthographic projection can be written as, 
\begin{gather}
 \begin{bmatrix} x \\ y \\ 0 \\ 1 \end{bmatrix}
 = P_v \begin{bmatrix} X \\ Y \\ Z \\ 1 \end{bmatrix}
\end{gather}
where where $P_v$ is known orthographic projection matrix to convert 3D world coordinates into 2D grid cell indices.
If more than one point are projected to the same grid cell in the spatial map, they are accumulated into the cell using an function to aggregate the features or predictions.
%they can be aggregated in various ways, for example by retaining only the point with maximum height, etc.

% Some maps that capture the geometric layout of the environment, are constructed by leveraging projective camera geometry with known camera parameters that allows conversion from 2D pixel coordinates to 3D world coordinates and then to 2D top-down grid coordinates.
% \todo{vague sentence - references? - done} 
% The observed image are first converted to 3D coordinates in the camera space with known intrinsic camera parameters, which are then converted to world coordinates with known camera pose. Finally the 3D world coordinates are voxelized and projected on a top-down 2D grid with a known projection matrix by summing over the height dimension.

\mypara{Accumulation.} 
During \textit{map update} there are many ways to aggregate features or prediction into the map including 1) overwriting the map with the latest observations ($m_t=m_{t-1}$), 2) performing mathematical operations such as max ($m_t=\max(m_t,m_{t-1})$) or mean ($m_t=\mean(m_t,m_{t-1})$), and 3) using a learned neural network. 
For learned aggregation functions, it is common to use a recurrent network (LSTM, GRU)  ($m_t=\GRU(m_t,m_{t-1})$).

\vspace{10pt}
\noindent During the process of building the map, there are several other important aspects to consider.

\mypara{Egocentric vs allocentric.}
There are also choices in the reference frame used for map-building, to either maintain a map with an \emph{egocentric} coordinate frame that is relative to the agent (e.g. +y coordinate to the front of the agent) or \emph{allocentric} (e.g. world) coordinate frame.  

\mypara{Tracking visited areas.} For the map to be complete, it is important for the agent to be able to determine whether it has already visited a location or not, and whether there are unexplored locations.  For a specific embodied task, it may not always be necessary for the agent to built a complete map if the task can be accomplished. 

\mypara{View point selection.} In the case of the embodied setting, the agent is also limited in the possible viewpoints it can observe, and must accumulate into the map, observations in a sequential manner.  This is in contrast to the non-embodied setting, where there can be more freedom in selection of viewpoints, and observations can be first collected and then analyzed together.  

\mypara{Online vs offline map building.} It is possible for the agent to build a map by exploring an environment first. After the map has been built, the agent can then start to perform the specific task. In this scenario, the agent builds the map and performs the task in two separate phases, a process known as \emph{offline} method of map building. Although this method saves compute time during the actual task, there is the overhead of an extra exploration phase for the agent to familiarize itself with a new environment.  This approach can be appropriate when the agent is expected to be reused in the same environment repeatedly.  However, since the map is a static snapshot and if it is not updated during the task, there can be mismatch between the actual state of the environment vs what was precomputed.  For instance, it might happen that the agent ends up at a location which has not been captured in the map. This might lead to the task failure.
Moreover, in real-life applications where a robot is expected to perform a task in an unseen environment, such as search-and-rescue operations, it's not ideal for it to spend extra time exploring the environment first and then performing the task.
In contrast a better way is to build or update the map during the task or \emph{online} so as to keep it updated at all times.

\mypara{Map building in real world.} 
Maps built in simulation in embodied AI tasks are often noisy due to unrealistic assumptions that limit its usage in the real-world.
% While the basic process of building a semantic map in embodied AI and real-world robots is the same, the embodied AI community makes some unrealistic assumptions that have made map building methods difficult to transfer well to real-world robots.
% The reason for this is well motivated in the embodied AI community since map building has been mostly studied as a sub-module in conjunction to solving more complex high-level reasoning tasks and abstracting out some real-world issues allows one to study the high-level task independently.
Map building has been mostly studied in the community as a sub-module in conjunction to solving more complex high-level reasoning tasks. Researchers have thus tried to investigate what type of maps are useful for which tasks and decoupling the issue of noisy sensors from map building enables them to do exactly that.
The most prominent of the assumptions is that of noiseless sensors. For example, sometimes the community assumes \textit{perfect localization} (agent's current location and orientation) at all times during navigation, which is unrealistic in real-world. This is mainly because GPS and Compass sensors are generally noisy, whenever available. However, in most indoor spaces, GPS might not even be available. SLAM methods which work really well in real-world robots operate under the assumption that GPS is not available, and relies on the onboard sensors to estimate its location on the map.
Another example of the noiseless sensor assumption is that of a perfect actuation, which means that when an agent initiates an action to move forward by 25cm, it will end up exactly at a location 25cm ahead of its current position. But real-world actuators are noisy and affected by varying friction on different surfaces, which results in significant drifts over time.
SLAM systems are inherently capable of addressing such issues by operating under uncertainty in the robot's pose estimation. Loop Closure is a sub-algorithm of SLAM which identifies previously visited locations and then uses them to correct accumulated errors in pose estimation. 
In general SLAM systems build a more consistent and accurate map of the environment in a real-world setting than the current mapping techniques in embodied AI.

\begin{figure}[h]
\centering
\includegraphics[width=\linewidth]{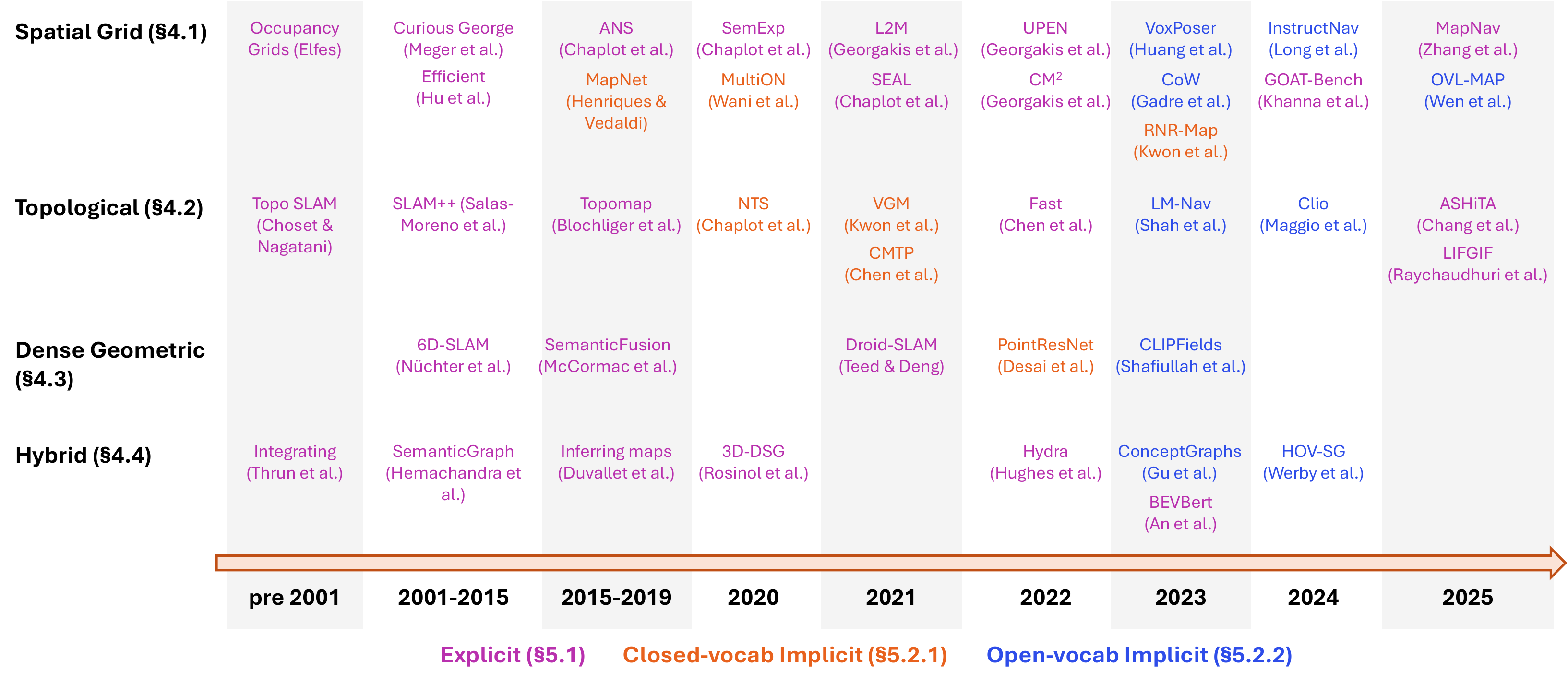}
\caption{
This timeline highlights how semantic mapping methods have progressed over the years, with increasing diversity in map structure and encoding. While various map structures have long been explored, the last few years have seen a shift toward open-vocabulary semantic maps, driven by advances in large language and vision-language models.
}
\label{fig:timeline}
\end{figure}

\section{Map structure}
\label{sec:structure}
In this section we will look at various map structures that have been used in prior works (see \Cref{tab-sem-map-grouped}) in more depth.
A semantic map can be structured in various ways: \textit{\spatial}, \textit{\topological}, \textit{\revision{dense geometric}} or a \textit{\hybrid}. 
% (see \Cref{fig:spatial_vs_topo_overview}). 
Spatial grid maps are metric maps of the environment such that its dimensions align to that of the environment and they can be structured as either 2D or 3D grids. Topological maps, on the other hand, represent the environment through a set of landmarks represented as nodes and relation between adjacent landmarks represented as edges in the form of a graph.
\revision{Dense geometric} maps are the densest form of 3D maps whose all three dimensions align to the 3D space such that each 3D point in the scene is captured in the map. Two or more of these types of maps are sometimes combined together to form Hybrid maps.
% Various approaches across different downstream tasks store different types of values in the map and use various types of aggregation function (see \Cref{tab-sem-map}).

\begin{figure*}[t]
\centering
% \framebox[0.9\linewidth]{
% \fbox{\rule[-.5cm]{0cm}{4cm} \rule[-.5cm]{4cm}{0cm}
% \includegraphics[width=0.9\linewidth]{chapters/figures/images/spatial_overview.pdf}
\includegraphics[width=\linewidth]{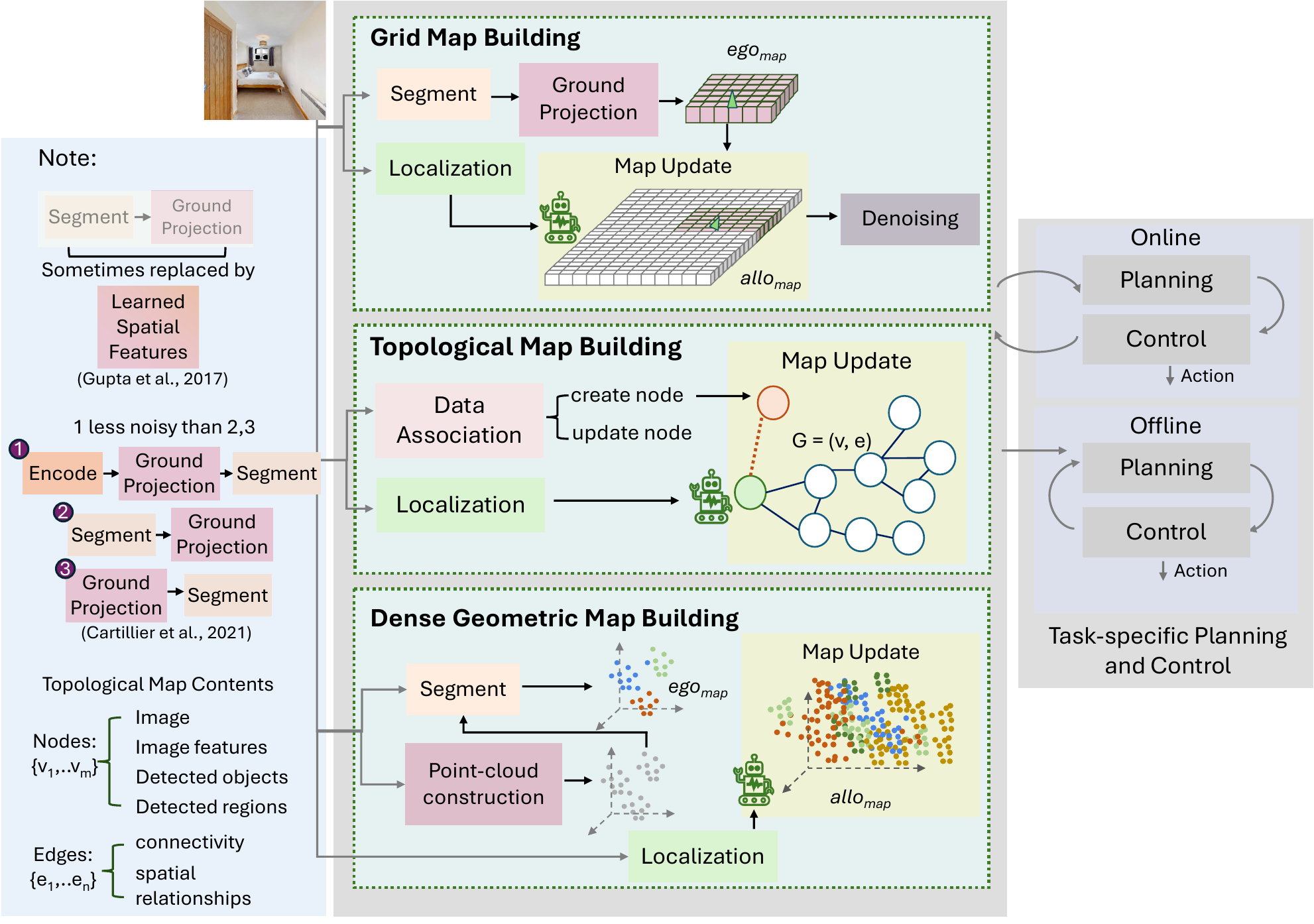}
% }
\caption{
\textbf{Grid map building.} 
A spatial grid map has dimensions ($M\times N \times K$) where $M$ and $N$ are spatial dimensions and $K$ is the number of semantic channels. The common pipeline to build the map is to \textit{segment} the input image, then \textit{ground project} into an egocentric map $ego_{map}$, which is then registered to the allocentric map $allo_{map}$ via \textit{map update} using the \textit{localized} agent pose. A \textit{denoising} step generally follows and the map is built online along with the task planning and control. While \cite{gupta2017cognitive} learns a spatial representation without segmenting and ground projecting, \cite{cartillier2020semantic} observes that encoding followed by ground projecting and then segmenting reduces noise in the produced map.\\
\textbf{Topological map building.}
A topological map is a graph-like structure, $G=(v,e)$ with nodes ($v=\{v_1,...,v_m\}$) and edges ($e=\{e_1,...e_n\}$), that can be built \textit{Online} or \textit{Offline} with or before task-specific planning and control respectively. Based on current observation, the agent performs \textit{localization} on the graph and then performs matching (\textit{Data Association}) with the current node, based on which either a new node is created or an existing node is updated. The nodes and edges may contain various types of semantic information thus enabling decision making in a task.\\
\textbf{Dense geometric map building.} A dense geometric map accumulates semantic information directly onto the 3D geometry of a scene, most popularly using a \emph{point-cloud}.
% This figure compares different spatial map building techniques.
% (top left) An egocentric Spatial feature representation may be directly learned from RGB images by using a neural network.
% (top right) The input image encoding is projected on to a ground plane and then transformed to an allocentric map via `Registration'~\citep{henriques2018mapnet}. 
% (bottom left) The input image is segmented and converted to 3D world coordinates first using known camera parameters and then projected on to the ground plane after discretization and summing over height dimension~\citep{chaplot2020object}. 
% (bottom right) In contrast to the previous method, encoding followed by projecting and then segmenting on the 2D map space is shown to produce noise-free spatial maps~\citep{cartillier2020semantic}.
% \textit{(Figures reproduced from paper.)}
}
\label{fig:map_building_compact}
\end{figure*}

\subsection{Spatial grid map}
\label{sec:spatial_map}
A spatial grid map $m_t$ is a ($M \times  N \times  K$) matrix where $M$ and $N$ are the spatial dimensions of the map and $K$ denotes the number of channels to store semantic information at that location. It is a grid like structure where each cell has a width and a height which correspond to a certain area in the physical environment. 
Current research on indoor embodied AI make use of environments from datasets such as Matterport3D (MP3D)~\citep{chang2017matterport3d} and Habitat-Matterport3D (HM3D)~\citep{ramakrishnan2021hm3d} which are 3D reconstructions of real-world spaces. Compared to MP3D, HM3D contains around 10x more scenes with high visual fidelity and lesser reconstruction artifacts. 
% \todo{expand this a bit - done}. 
These environments are typically for houses or office spaces with a total area of $ < 1000\text{m}^2$. A grid map with each cell representing an area of $400-900\text{cm}^2$ is found to be good enough to represent such spaces~\citep{wani2020multion,raychaudhuri2023mopa}. 
At the start of each episode, the spatial map is initialized with a tensor of size ($M \times  N \times  K$), and gradually built as the agent moves around the environment. These are often 2D top-down maps~\citep{gupta2017cognitive,henriques2018mapnet,narasimhan2020seeing,cartillier2020semantic} with the first two dimensions corresponding to the spatial dimensions of the environment. However, some build 3D spatial maps~\citep{chaplot2021seal} to capture the vertical dimension, in which case the map $m_t$ is a 4D tensor ($M \times  N \times P \times  K$). 
% \todo{explain that often these are 2D top-down maps, but can also have vertical dimension to create 3D maps - done}

A spatial map may be built in a number of ways depending on whether raw features are directly projected onto the map, or whether a semantic segmentation is used (see \Cref{fig:map_building_compact}). 
% There are also choices in the reference frame used for map-building, to either maintain a map with an \emph{egocentric} coordinate frame that is relative to the agent (e.g. +y coordinate to the front of the agent) or \emph{allocentric} (e.g. world) coordinate frame.  
One way is to learn an egocentric projection of the image features as in CMP~\citep{gupta2017cognitive} which forms the egocentric map. In CMP, the egocentric observations are first encoded with a learned image encoder network such as ResNet~\citep{he2016deep}), and then the network learns to predict an egocentric projection of the image features, without explicit supervision on the map. Instead, the mapper is trained end-to-end along with the planner to predict actions. Egocentric maps, however, not only fail to capture the global structure of the environment, but `forget' most of the past observations. Thus in long-horizon planning tasks, where the agent needs to `remember' its past observations for efficiency, egocentric maps fall short.
% \todo{elaborate on what CMP does and deficiencies of egocentric map vs allocentric - done}. 

To maintain an allocentric map, it is necessary to take the egocentric information at each time step and aggregate it into a global map.  One way to achieve this is to first obtain egocentric projection of image features and then aggregate to a global allocentric map of the environment via a process known as \emph{registration}. Registration allows the map to incorporate new observations on to specific grid cells. In case the grid cells are already occupied, the new observations are accumulated with the existing ones by employing an aggregation function.   This aggregation function can be as simple as taking the latest or the average, but can also be a learned network (see \Cref{sec:semantic-map-building}). MapNet~\citep{henriques2018mapnet} builds an allocentric map by projecting the egocentric image features using depth observations and known camera intrinsics on a 2D top-down grid.
This ground projection results in an egocentric projection on a spatial neighborhood around the camera.
Next it performs registration by first obtaining a stack of egocentric maps rotated $r$ times and then performing a dense matching with the allocentric map from the previous step to obtain the agent's current pose on the map. The dense matching is efficiently implemented with convolution operators. A LSTM then performs the aggregation of the current observations rotated by the current pose with the allocentric map from the previous step.  While an LSTM is used in this work for aggregation, other functions and neural architectures can be used for aggregation as well (see \Cref{tab-spatial-sem-map} for a summary of aggregation methods used in different works).

Another way to build an allocentric spatial map is to first convert the image pixels to 3D coordinates in the camera space with known camera intrinsics. The camera coordinates are then converted to world coordinates with known camera pose. Finally the 3D world coordinates are voxelized and projected on a top-down 2D grid with a known projection matrix by summing over the height dimension. This approach is followed in Semantic MapNet \citep{cartillier2020semantic} and MOPA \citep{raychaudhuri2023mopa}. Spatial maps built this way might be noisy due to noisy sensors and need an additional denoising step. Semantic MapNet uses a learned denoising network while MOPA employs a heuristic approach by selecting the centroid of a noisy cluster to obtain a clean map.
Following on the last technique, Semantic MapNet~\citep{cartillier2020semantic} shows that first encoding the image, followed by projecting on to the ground plane and finally performing segmentation on the 2D map reduces noise in the spatial map thereby eliminating the need for an additional denoising step.
Irrespective of the approach, it may happen that multiple image features are projected on to the same grid cell. In such cases, it is important to have a scheme for aggregating the features.  Some common approaches to aggregation is to take the maximum  \citep{henriques2018mapnet,wani2020multion,cartillier2020semantic}, mean~\citep{huang2023visual}, or the sum of the feature values \citep{chaplot2020object}.

\begin{table}[ht]
\caption{
\textbf{Spatial grid maps.} Prior works build spatial grid maps (2D or 3D) for various embodied tasks. Information is aggregated onto the map over time in many different ways such as learned recurrent networks and replacing with most recent information among others.  
}
\label{tab-spatial-sem-map}
\centering
    \resizebox{\linewidth}{!}{
    \begin{tabular}{lllp{2.5cm}l}
    \toprule
    \bf Method 
     &\bf Task 
    &\bf Environment &\bf Dataset 
    &\bf Aggregation
     % &\bf Implicit &\bf Explicit
    % &\bf Training 
    % &\bf Stored Values
    % &\bf Value Aggregation
    \\
    
    \midrule
    Efficient~\citeyearpar{hu2013efficient} &3D scene analysis &robot &VMR-Oakland~\citeyearpar{xiong20113}, Freiburg~\citeyearpar{behley2012performance} &most recent \\
    CMP~\citeyearpar{gupta2017cognitive} &PointNav, ObjectNav &Custom simulator & S3DIS~\citeyearpar{armeni20163d} & weighted mean \\
    
    MapNet~\citeyearpar{henriques2018mapnet} &Mapping  &Doom~\citeyearpar{mahendran2016researchdoom} &Active Vision Dataset~\citeyearpar{ammirato2017dataset} &LSTM \\
    
    ANS~\citeyearpar{chaplot2020learning} & Exploration  
    & Habitat~\citeyearpar{savva2019habitat} & Gibson~\citeyearpar{GIBSONENV}, MP3D~\citeyearpar{chang2017matterport3d} &channel-wise max-pool \\
    
    MultiON~\citeyearpar{wani2020multion} &MultiON &Habitat~\citeyearpar{savva2019habitat} &MP3D~\citeyearpar{chang2017matterport3d} &element-wise max-pool \\
    
    SemExp~\citeyearpar{chaplot2020object} & ObjectNav 
    & Habitat~\citeyearpar{savva2019habitat} & Gibson~\citeyearpar{GIBSONENV}, MP3D~\citeyearpar{chang2017matterport3d} &channel-wise max-pool \\
    
    SemanticMapNet~\citeyearpar{cartillier2020semantic} &ObjectNav, Visual QA  &Habitat~\citeyearpar{savva2019habitat} &MP3D~\citeyearpar{chang2017matterport3d} &GRU \\

    \revision{L2M}~\citeyearpar{georgakis2021learning} &ObjectNav &Habitat~\citeyearpar{savva2019habitat} &MP3D~\citeyearpar{chang2017matterport3d}, ObjectNav~\citeyearpar{batra2020objectnav} &most recent \\
    \revision{SEAL}~\citeyearpar{chaplot2021seal} &ObjectNav &Habitat~\citeyearpar{savva2019habitat} &custom with Gibson~\citeyearpar{GIBSONENV} scenes &channel-wise max-pool \\

    \revision{CM$^2$}~\citeyearpar{georgakis2022cross} & Vision and Language Nav &Habitat~\citeyearpar{savva2019habitat} &VLN-CE~\citeyearpar{krantz2020beyond} &most recent \\
    
    \revision{RNR-Map}~\citeyearpar{kwon2023renderable} & Visual Nav &Habitat~\citeyearpar{savva2019habitat} &Gibson~\citeyearpar{GIBSONENV} & mean \\
    \revision{Le-RNR-Map}~\citeyearpar{taioli2023language} & Visual Nav &Habitat~\citeyearpar{savva2019habitat} &Gibson~\citeyearpar{GIBSONENV} & mean\\
    \revision{VoxPoser}~\citeyearpar{huang2023voxposer} &Table-top manipulation &SAPIEN~\citeyearpar{xiang2020sapien}, robot & custom & - \\
    MOPA~\citeyearpar{raychaudhuri2023mopa} &MultiON  & Habitat~\citeyearpar{savva2019habitat} & HM3D~\citeyearpar{ramakrishnan2021hm3d} &most recent  \\

    GOAT-Bench~\citeyearpar{khanna2024goat} & Multimodal ObjectNav & Habitat~\citeyearpar{savva2019habitat} &HM3D-Sem~\citeyearpar{yadav2023habitat} &most recent   \\
    
    \revision{Instruction-guided}~\citeyearpar{wang2025instruction} &Vision-language navigation &Habitat &VLN-CE~\citeyearpar{krantz2020beyond} & max-pool \\
    
    \bottomrule
    \end{tabular}}
\end{table}
\mypara{Summary.}
Spatial grid maps capture dense information about the environment. Such representations are useful for the agent to better reason about the spatial structure of the environment. However, the spatial maps need to be initialized with a certain width and height and as such is hard to scale if the environment size changes. Moreover, it consumes a lot of memory which might affect agent performance in the task. 

% Most of the prior works~\citep{chaplot2020object,wani2020multion,chaplot2020learning, raychaudhuri2023mopa} assume perfect  odometry.

% \Cref{tab-spatial-sem-map} summarizes various implicit as well as explicit methods.
% \input{chapters/tables/tab-spatial-map}
% \input{chapters/figures/fig-latent-map-cmp}
% \input{chapters/figures/fig-latent-map-mapnet}

% \input{chapters/figures/fig-topological-overview}
\subsection{Topological map}
\label{sec:topo_map}
Compared to the high-precision grid maps, topological maps are graph-like structures with nodes connected to each other by edges. This essentially abstracts a large space into significant areas (nodes) where the agent can take decisions and connections or paths between them (edges)~\citep{johnson2018topological}. This enables parsing the environment into a local and a global structure such that the agent can plan locally in the small space represented as nodes while navigating the large space through graph search following the edges.
This way of planning and navigating is inspired from how humans navigate in an unseen environment in that they identify and memorize significant landmarks and find paths to reach those landmarks~\citep{janzen2004selective,foo2005humans,object_important,epstein2014neural}.
\revision{Moreover, in instruction-following tasks, where a robot follows instructions (e.g., `go to the yellow chair in front of you'), the language instruction is often represented as a graph with high initial uncertainty. As the robot moves and observes its environment, it uses a probabilistic model to update this graph, leading to improved task performance~\revisedcitep{tucker2019learning,patki2019inferring,arkin2020multimodal,raychaudhuri2025zeroshotobjectcentricinstructionfollowing}.}
% The nodes encode semantic information about locations in the environment while the edges store relationship among the nodes. The main idea is to capture semantics of important landmarks in an environment. 
% When humans need to navigate in an unseen environment, it is common for them to identify and memorize landmarks and plan a task based on them~\citep{janzen2004selective,foo2005humans,object_important,epstein2014neural}. 
% \todo{add citation - done} 
Thus topological maps have been a popular choice in traditional robotics research \revision{such as SLAM~\citep{thrun2006graph,rosinol2020kimera,campos2021orb}, manipulation~\revisedcitep{arkin2020real,patki2020language}, object search~\revisedcitep{aydemir2011search,aydemir2013active,lorbach2014prior}, language grounding~\revisedcitep{macmahon2006walk,Chen_Mooney_2011,tellex2011understanding,matuszek2013learning,duvallet2013imitation,gong2018temporal,paul2018efficient}} as well as in embodied AI research~\citep{savinov2018semi,chen2021topological,chaplot2020neural,kwon2021visual,conceptgraphs,mehan2024questmaps,garg2024robohop,an2024etpnav,yang20243dmem3dscenememory,tang2025openin}.

\begin{table*}[ht]
\caption{
\textbf{Topological maps.} 
Various works in indoor embodied AI build topological map either in an exploration phase or online while performing the task. The nodes often store explicit, learned features about the observation or temporal information (visitation timestep), while edges may store relative poses between a pair of nodes or types of edges.
}
\label{tab-topo-map}
\centering
    \resizebox{\linewidth}{!}{
    \begin{tabular}{llp{2.6cm}p{3.8cm}p{2.6cm}p{3cm}}
    \toprule
    \bf Method &\revision{\bf Task}
    &\bf Map Building Phase
    &\bf Node values
    &\bf Node Feature Encoder
    &\bf Edge values
    \\
    
    \midrule

    \revision{Topo SLAM}~\citeyearpar{choset2001topological} &SLAM &Online &Pose &\xmark &\xmark \\
    \revision{SLAM++}~\citeyearpar{salas2013slam++} &SLAM &Online &Pose, object type &\xmark &\xmark \\
    \revision{Imitation}~\citeyearpar{duvallet2013imitation} &Instruction following &Online &Pose &\xmark &\xmark \\
    \revision{TopoMap}~\citeyearpar{blochliger2018topomap} &SLAM &Online &free space &\xmark &\xmark \\
    SPTM~\citeyearpar{savinov2018semi} & ImageNav 
    &Pre-Exploration &Image features &ResNet18~\citeyearpar{he2016deep} & \xmark \\

    \revision{Compact}~\citeyearpar{patki2019inferring} & Instruction following &Online  &Semantic attributes & \xmark & \xmark \\
    
    \revision{Multimodal}~\citeyearpar{arkin2020multimodal} &Instruction following &Online &Semantic attributes & \xmark & \xmark \\
    
    NTS~\citeyearpar{chaplot2020neural} & ImageNav 
      & Online  &Image features &ResNet18 & Relative pose in polar coordinates \\

    CMTP~\citeyearpar{chen2021topological} &Instruction following &Pre-Exploration &Image features  &ResNet152 & Relative pose in discrete polar coordinates (8 directions, 3 distances (0-2m,2-5m,>5m) \\
    
    VGM~\citeyearpar{kwon2021visual} &ImageNav & Online &Image features, visitation timestep &ResNet18  & \xmark \\

    \revision{Fast}~\citeyearpar{chen2022fast} &Path planning &Pre-Exploration &free space &\xmark &\xmark \\
    
    TSGM~\citeyearpar{kim2023topological} &ImageNav & Online & ImageNode stores image features; ObjectNode stores features for detected objects & pretrained image and object encoders &\xmark \\
    
    \revision{LM-Nav}~\citeyearpar{shah2022lmnav} &Instruction following &Online & Image features &CLIP~\citeyearpar{radford2021learning} & \xmark \\
    RoboHop~\citeyearpar{garg2024robohop} &Language querying & Online & Image features for each image segment & CLIP, DINOv2~\citeyearpar{oquab2023dinov2} & edge types denoting inter- and intra-image connectivity \\
    
    \revision{Clio}~\citeyearpar{maggio2024clio} & Mobile manipulation &Online &object semantics & CLIP &\xmark  \\
    \revision{LIFGIF}~\citeyearpar{raychaudhuri2025zeroshotobjectcentricinstructionfollowing} & Instruction following &Online  &Semantic attributes & \xmark & \xmark \\
    \revision{ASHiTA}~\citeyearpar{chang2025ashitaautomaticscenegroundedhierarchical} &Task planning &Online &Task, sub-task, objects & MobileCLIP~\citeyearpar{vasu2024mobileclip} &\xmark \\

    \bottomrule
    \end{tabular}}
\end{table*}

A key design decision during topological map building is what should be represented as nodes and what should be edges.
Generally speaking, the nodes encode semantic information about locations in the environment such that the agent can make a decision whereas the edges store relationship or connection between the nodes.  
For indoor navigation, the landmarks for the nodes are typically objects in the environments.  They can also be openings or intersections~\citep{fredriksson2023semantic}, locations the agents has visited~\citep{chaplot2020neural}, and other regions of interest~\citep{kim2023topological,shah2022lmnav,garg2024robohop}. 
For navigation, two nodes are connected with an edge if it is possible to navigate from one node to another. Some methods also store spatial relationships between the nodes~\citep{conceptgraphs} in the edges to enable better reasoning.

One way to construct a topological map (see \Cref{fig:map_building_compact}) is during an exploration phase previous to the actual task and then use the graph to plan a path to the node most similar to the target, for example, in Semi-Parametric Topological Memory (SPTM) \citep{savinov2018semi}. 
% builds a topological map of the environment in the first phase of exploration which is then used to navigate the environment. Each node in this graph stores an observation at a specific location and two nodes are connected if they are adjacent or the observations are similar. They then use a deep network (Retrieval Network) to retrieve a waypoint node from the graph based on the current observations during navigation. Another network called Locomotion Network is then responsible to generate an action given the waypoint. Both retrieval and locomotion networks are trained using supervised learning.
During exploration the agent follows multiple random trajectories for each environment to form a node for every visited location and add an edge between the current node and the previous one to encode connectivity or reachability between them.
A common post-processing step includes trimming out redundant nodes and edges to form a sparse graph~\citep{chen2021topological}. When the graph of one environment is collected from multiple random trajectories, it is also common to merge these graphs into one. However, a topological map generated this way in a pre-exploration phase is still sparse meaning that some observations in the environment might not have been captured by the graph. This affects the agent performance in the downstream task. Moreover, they need a pre-exploration phase which makes them unsuitable for unseen environments. 

To mitigate this issue some works construct the topological map online while the agent is navigating during performing the task as is the case with Neural Topological SLAM (NTS)~\citep{chaplot2020neural}. NTS consists of several modules -- `Graph Update' to update the topological map from observations, `Global Policy' to sample subgoals on the map and `Local Policy' which outputs discrete navigation actions to reach the subgoal. The `Graph Update' method gradually updates the nodes and edges in the graph from the current observations and agent poses. It first attempts to localize the agent on the graph from the previous timestep. If the agent gets localized in an existing node, it adds an edge between that node and the node from the last timestep. It also stores the relative pose between the two nodes represented as $(r, \theta)$ where $r$ is the relative distance between the nodes and $\theta$ is the relative direction. If the agent is unable to be localized, a new node is added to the graph. 

Another important aspect in the topological map creation is how to determine if two observations are similar to each other, in which case the two are mapped to the same node. If they are not similar, two different nodes exist for the two observations.
This requires the map building methods to compare RGB images. The goal here is to classify two images as similar if (1) they are exactly the same or (2) there is a slight change in direction or distance between the two. 
Traditionally this is the problem of \textit{data association} in SLAM-based systems, where an incoming observation could be matched to multiple landmarks (nodes) and either the best match is selected or a new node is created to mark a new landmark~\citep{bowman2017probabilistic,dellaert2000feature}.
In embodied AI some works use a pretrained classifier network to implement this. The network is trained to classify whether two images are from the same area. NTS uses MLP trained with a cross-entropy loss in a supervised manner to predict whether are similar. This however needs annotated pairs of training data.
Cross-Modal Transformer Planner (CMTP) \citep{chen2021topological}, on the other hand, uses an oracle `Reachability Estimator' to first obtain the geodesic distance between the two underlying locations based on the traversibility of the 3D mesh. If the distance is below a threshold, it maps them to the same node.
Visual Graph Memory (VGM) \citep{kwon2021visual} also uses a pretrained network to determine if two images are similar. But they learn an unsupervised representation of the observations which are then projected onto an embedding space. The idea is to have the embeddings of observations coming from nearby areas clustered together because they are likely to have similar appearances. The training data in this case consist of randomly sampled
observations from the training environments, thus eliminating the need for manual annotations. 
\cite{kim2023topological} on the other hand use semantic similarity score obtained from a pretrained network~\citep{li2021prototypical} between two images to determine whether they are the same nodes. A similar approach is taken in LM-Nav~\citep{shah2022lmnav} where they use CLIP to calculate the cosine similarity between image features.
\Cref{tab-topo-map} compares different methods that build topological maps.

\mypara{Summary.}
In summary, topological maps are convenient to build and maintain due to their concise and condensed representation when compared to spatial maps. They are memory-efficient and can easily be scaled as the environment size increases by simply adding more nodes to the graph. However, they capture only certain landmarks in the environment and as such lack dense global information. This might lead to overlooking visual cues in a cluttered indoor scene that could be helpful for the agent to carry out spatial reasoning.

\subsection{\revision{Dense geometric map}}
% \todo{rethink name: 3D scene map?}}
\label{sec:dense_map}

\revision{
Semantic information can also be directly accumulated onto the 3D geometry of a scene using triangle meshes~\revisedcitep{valentin2013mesh}, surfels~\revisedcitep{stuckler2014multi} or point clouds~\citep{conceptfusion}, with point cloud maps being the most widely used due to simplicity and ease of use. While these representations are not always traditionally considered ``maps'', they effectively function as semantic maps by coupling geometric structure with semantic content.
Dense geometric map representations, particularly point clouds, have long been widely used in robotics for tasks such as mapping, localization and navigation, as well as in broader 3D scene understanding tasks~\citep{Peng2023OpenScene,xu2024esam}.
Recently, there has been a surge of interest in applying these representations to embodied AI for downstream reasoning and decision-making~\citep{conceptgraphs}. 
We further categorize dense geometric map representations into \emph{point cloud maps} (\Cref{sec:point_cloud}) and \emph{neural fields} (\Cref{sec:neural_fields}). Point cloud maps store 3D points with associated semantic labels whereas neural fields represent scenes as continuous functions using neural networks (see \Cref{tab-dense-map} for a summary of prior works).
}

\begin{table*}[ht]
\vspace{-10pt}
\caption{
\revision{\textbf{Dense geometric maps} allow storing semantics onto the 3D geometry of a scene. It can be further categorized into point-cloud maps and neural fields. Here, VL is visual-and-language, Str is Structure, PC is point-cloud, SF is surfel, GS is gaussian splat and NF is neural fields.
}
}.
\label{tab-dense-map}
\centering
    \resizebox{\linewidth}{!}
    {
    \begin{tabular}{lp{3.5cm}p{2.5cm}lp{3.5cm}l}
    \toprule
    \bf Method &\bf Task &\bf Env
    &\bf Str &\bf Encoding & \bf Train 
    \\
    \toprule
    6D-SLAM~\citeyearpar{nuchter20076d} &SLAM  & robot &PC &geometry &\xmark \\
    Towards semantic maps~\citeyearpar{nuchter2008towards} &SLAM &robot &PC &3D location, category label, robot pose &\cmark \\
    Robust 3D mapping~\citeyearpar{may2009robust} &SLAM & robot &PC &geometry &\xmark \\
    3D object-class map~\citeyearpar{stuckler2012semantic} &SLAM &robot &SF & category label &\cmark \\
    Parsing~\citeyearpar{triebel2012parsing} &Semantic segmentation &robot &PC &3D location, category label &\xmark \\
    Street Scenes~\citeyearpar{floros2012joint} &Semantic segmentation &simulator &PC &3D location, category label &\xmark \\
    Dense 3D~\citeyearpar{hermans2014dense} &3D reconstruction &simulator &PC &3D location, category label & \cmark \\
    SemanticFusion~\citeyearpar{mccormac2017semanticfusion} &SLAM &robot &SF &3D location, normal, category label &\cmark  \\
    Droid-SLAM~\citeyearpar{teed2021droid} &SLAM &robot &PC &geometry &\cmark \\
    Voldor+ SLAM~\citeyearpar{min2021voldor+} &SLAM & robot &PC &geometry &\xmark \\
    Semantic-NeRF~\citeyearpar{zhi2021place} & Scene understanding &Habitat, Replica &NF &category label &\cmark \\
    NeSF~\citeyearpar{vora2021nesf} &Semantic segmentation &Kubric &NF & category label &\cmark \\
    3D aware ObjNav~\citeyearpar{zhang20223d} & Object navigation &Habitat & PC &3D location, semantic category, consistency information &\cmark \\
    ConceptFusion~\citeyearpar{conceptfusion} &Rearrangement, autonomous driving, semantic segmentation &Replica, robot &PC &3D location, normal, confidence, color, pixel aligned VL features &\xmark \\
    OpenScene~\citeyearpar{Peng2023OpenScene} &Open-vocab scene understanding & ScanNet, MP3D, nuscenes &PC &CLIP &\xmark \\
    LERF~\citeyearpar{kerr2023lerf} & Open-vocab scene understanding &robot &NF &learned neural VL features &\cmark \\
    LGM~\citeyearpar{shen2023distilled} &  Open-vocab mobile manipulation &robot & GS &CLIP dense patch features & \cmark \\ 
    CLIP-Fields~\citeyearpar{shafiullah2023clip} &Object navigation &Habitat, robot &NF &learned neural VL features &\cmark  \\
    LangSplat~\citeyearpar{qin2024langsplat} &Open-vocab semantic segmentation &LERF &GS &VL features from CLIP, SAM &\cmark \\
    SemanticGaussians~\citeyearpar{guo2024semantic} &Open-vocab scene understanding &ScanNet, LERF &GS &VL features from CLIP, SAM, LSeg &\cmark \\
    GaussNav~\citeyearpar{lei2025gaussnav} & Visual navigation &Habitat & GS &category label &\xmark \\
    GeFF~\citeyearpar{qiu2024learning} &  Open-vocab mobile manipulation &robot & GS &MaskCLIP~\citeyearpar{zhou2022extract} &\cmark \\
    SGS-SLAM~\citeyearpar{li2024sgs} &SLAM &robot &GS & category label &\cmark \\
    GaussianGrasper~\citeyearpar{zheng2024gaussiangrasper} &Open-vocab tabletop manipulation &robot &GS & CLIP &\cmark \\
    GeomGS~\citeyearpar{lee2025geomgs} &Localization &robot &GS &geometric confidence &\cmark \\
    \bottomrule
    \end{tabular}
    }
% \vspace{-10pt}
\end{table*}

\revision{\subsubsection{Point cloud maps}
\label{sec:point_cloud}
Point clouds have become increasingly popular over the years due to their ease of use. Given a 3D point cloud, semantic labels can be associated with each point and thus by coupling geometric structure with semantic content, they serve as a dense semantic map.
It is also possible to spatially extend each point to a 3D Gaussian with a covariance matrix.  Such representations are called Gaussian splats, and allow for differentiable rendering.
Point clouds (and Gaussian splats more recently) have been adopted in a variety of robotics tasks as well as in 3D scene understanding tasks, gaining popularity in the embodied AI community in recent years.
}

\revision{
In robotics, point cloud-based mapping has undergone significant evolution, driven by advances in sensing, computational power and SLAM algorithms. Early efforts~\citep{nuchter20076d,may2009robust} use 3D sensor scans to build geometrically consistent representations of large-scale environments by emphasizing accuracy and loop-closure. As real-time visual SLAM has matured, dense point cloud generation from RGB-D sensors became feasible, enabling detailed indoor reconstructions~\citep{whelan2015elasticfusion,dai2017bundlefusion,schops2019bad,min2021voldor+}. 
More recently, learning-based SLAM systems like Droid-SLAM~\citep{teed2021droid} integrate deep feature extraction and end-to-end optimization to improve SLAM.
CoFusion~\citep{runz2017co} introduce systems for dense SLAM with semantic fusion, pushing toward maps that are not only geometrically accurate but also semantically meaningful. 
This trend has been accelerated recently with ConceptFusion~\citep{conceptfusion}, 
% and Clio~\citep{maggio2024clio}, 
which aligns open-vocabulary visual features (e.g., CLIP) with 3D reconstructions to create task-agnostic and queryable maps. These efforts represent a shift from purely geometric maps toward dense, semantically-rich, open-vocabulary representations that support a wide range of downstream tasks.
Moreover, Gaussian splatting has recently emerged as a powerful alternative to traditional point cloud representations in robotics, enabling high-fidelity scene reconstruction and differentiable 3D representations useful for tasks such as visual localization, SLAM, and manipulation. Recent works such as SGS-SLAM~\citep{li2024sgs} demonstrate how to include semantic labels onto Gaussian splats to effectively perform SLAM, and  GaussianGrasper~\citep{zheng2024gaussiangrasper} demonstrates how these representations can be fused with semantics to support open-vocabulary grasping.
}

\revision{
In embodied AI, maintaining point cloud maps typically prove to be memory-intensive and not ideal for real-time use. However some efforts have been made to improve efficiency. For instance, \cite{zhang20233d} builds a 3D semantic scene representation based on an online point cloud-based construction algorithm~\citep{zhang2020fusion}, made efficient by using a tree-based dynamic data structure. This method is very memory efficient and a lot faster when applied to the ObjectNav task. Recent efforts also use Gaussian splatting in navigation tasks. For example, \cite{lei2025gaussnav}, based on Gaussian splatting, achieves state-of-the-art performance on the ImageNav task.
}

\revision{
In 3D scene understanding, point cloud representations have become a cornerstone due to their flexibility and fine spatial resolution. Early learning-based approaches such as PointNet~\citep{qi2017pointnet} and PointNet++~\citep{qi2017pointnet++} enable direct processing of unordered point sets for tasks like semantic segmentation and object classification.
These were followed by more expressive models like KPConv~\citep{thomas2019kpconv}, which introduce learnable convolutions, and Point Transformer~\citep{zhao2021point}, which leverage attention mechanisms to capture long-range dependencies across points. More recent work integrates multi-modal and language supervision into point cloud understanding\citep{Peng2023OpenScene,xu2024esam}, thus enabling fine-grained inference such as semantic segmentation, affordance prediction and 3D object search. }
OpenScene~\citep{Peng2023OpenScene}, for example, predicts dense 3D features so that they are co-embedded with the corresponding text and the image in the CLIP embedding space. This allows the association of each 3D point in the scene with semantic information such that the scene can be queried using text to infer physical properties, affordances, etc.
CLIP2Scene~\citep{chen2023clip2scene} also uses CLIP to perform a 3D point cloud segmentation on outdoor scenes for application in autonomous driving.
However, because the optimization happens per scene in many of these methods, they are typically expensive and not suitable for real-time use in embodied applications.

\mypara{Summary.}
\revision{
Point cloud maps remain a widely used representation in the research community due to their simplicity, high fidelity and direct correspondence with sensor data.
However, challenges persist in handling noise, achieving scalability in large environments and maintaining consistent semantics over time. Future research should aim to integrate multi-modal features and improve robustness to dynamic scenes in point cloud maps, all the while maintaining efficiency.
}
% \vspace{-10pt}

\revision{\subsubsection{Neural fields}
\label{sec:neural_fields}
}
\revision{
Neural fields are continuous functions that map spatial coordinates (and sometimes viewing directions) to signals such as color, occupancy or semantic features. Unlike point clouds that discretely store information, neural fields encode entire 3D scenes as compact, continuous functions.
}
Often, a neural network (typically MLP) is used to obtain the features associated with each point given the position (x,y,z) as input~\citep{mildenhall2021nerf,mescheder2019occupancy}.  
% Such neural functions that transform coordinates to real-valued vectors are also known as neural fields.  These neural fields can produce dense 3D semantic maps where each 3D point in the scene is captured and represented in the map.
There are typically two strategies to train neural fields 1) use distillation~\citep{kerr2023lerf,qiu2024learning} to provide features that are similar to a pretrained 2D backbone, such as CLIP~\citep{radford2021learning}, and 2) use of differentiable renderer to match the rendered semantics in addition to color~\citep{zhi2021place,vora2021nesf}.

\revision{
In robotics, recent works~\citep{sucar2021imap,zhu2022nice,rosinol2023nerf} have integrated neural fields into SLAM systems for real-time mapping. While these enable compact, high-fidelity 3D reconstruction and view synthesis, they are more computationally intensive and less interpretable than traditional methods. Thus challenges remain in training speed, memory demands and dynamic scene handling. In contrast, neural Signed Distance Fields (SDFs) are a specific type of neural field that encode the signed distance to the closest surface, enabling precise geometric reasoning. Neural SDFs have been shown to be particularly suited for tasks like mapping, collision checking, and planning due to their ability to represent detailed geometry with differentiable distance metrics~\citep{ortiz2022isdf,camps2022learning,talha2025differentiable}.
}
In embodied AI, a representative work using neural field maps include CLIP-Fields~\citep{shafiullah2023clip}, which learns a continuous, neural implicit representation of 3D environments by aligning 2D image features from CLIP with 3D points in space using volumetric rendering. This enables the agent to build an open-vocabulary, queryable 3D map where each spatial point can be semantically queried using natural language, without needing task-specific supervision. 

\revision{
In 3D scene understanding, neural fields have emerged as a powerful representation from early works on NeRF~\citep{mildenhall2021nerf} to incorporating semantics into neural fields, enabling tasks such as semantic segmentation, object discovery and scene parsing in 3D~\citep{fu2022panoptic}. More recent advances~\citep{wang2022clip,kerr2023lerf} introduce language-supervised neural fields by integrating CLIP features, enabling open-vocabulary semantic reasoning and text-driven queries over 3D scenes. 
}

% OpenScene~\citep{Peng2023OpenScene}, for example, predicts dense 3D features so that they are co-embedded with the corresponding text and the image in the CLIP embedding space. This allows the association of each 3D point in the scene with semantic information such that the scene can be queried using text to infer physical properties, affordances, etc.

% CLIP2Scene~\citep{chen2023clip2scene} also uses CLIP to perform a 3D point cloud segmentation on outdoor scenes for application in autonomous driving. %In Language-Assisted Feature Learning~\citep{zhang2023language}, linguistic-based object features are learned for the set of 3D points in the scene from a text query by using a language parser.  
% However, because the optimization happens per scene in many of these methods, they are typically expensive and not suitable for real-time use in embodied applications.
% A representative work that uses  neural field maps for embodied AI include CLIP-Fields~\citep{shafiullah2023clip}.
% \todo{more citations similar to clip-fields}
% \cite{zhang20233d}, on the other hand, builds a 3D semantic scene representation based on an online point cloud-based construction algorithm~\citep{zhang2020fusion} made efficient by using a tree-based dynamic data structure. This method is very memory efficient and a lot faster when applied to the ObjectNav task. 
% Similarly, \cite{lei2024gaussnavgaussiansplattingvisual} shows that a 3D representation based on Gaussian Splatting is able to achieve state-of-the-art performance on the ImageNav task.

\mypara{Summary.}
\revision{
Neural fields are gaining popularity in robotics and embodied AI for building continuous 3D scene representations that capture both geometry and semantics. However, challenges remain in real-time training and deployment in dynamic environments. An important future direction would be to focus on making neural fields more efficient, adaptable, and compatible with foundation models for open-world semantic reasoning.
}

\subsection{Hybrid maps}
\label{sec:hybrid_map}
So far we have seen how prior works structure maps as either a spatial grid or a landmark based scene graph or a more dense \revision{geometric map}. However there is a more recent effort on combining these different structures into a single map representation. This helps to capture information at various granularity and perform different types of reasoning on the environment.

The combination of metric information from grid-based maps together with a topological map is also known as a \textit{topometric} map~\citep{thrun1998integrating,tomatis2001combining,blanco2008toward,konolige2011navigation,ko2013semantic,an2023bevbert,mozos2007supervised,vasudevan2008bayesian,zender2008conceptual,pronobis2010multi,hemachandra2011following,walter2013learning,walter2014framework,hemachandra2015learning,duvallet2016inferring,10878387}.
\citet{thrun1998integrating} proposes a single statistical mapping algorithm that first constructs a coarse topological map and uses it to construct a fine-grained grid map. They show that the topological map solves a global alignment problem by correcting large odometry errors, while the grid map solves a local alignment problem by producing high-resolution maps.
\citet{tomatis2001combining}, on the other hand, proposes a compact environmental model where corners and hallways are represented by a topological map and rooms are represented by a grid map, both of which are connected in a single representation. When the robot is moving in hallways, it creates and updates the global topological map, and as soon as it enters a room, it creates a new local metric map\footnote{They differentiate hallways and rooms by using laser sensor such that thin long open spaces are considered as hallways whereas other open spaces are considered as rooms.}. They argue that the robot will only need to be precise inside rooms (e.g. manipulating objects, etc.) which justifies the need for the fine-grained precise metric maps, while the topological map is used to simply maintain global consistency in indistinguishable spaces such as long hallways and transitioning between significant places. BEVBert~\citep{an2023bevbert} is a more recent method that constructs hybrid maps offline and then learn a multimodal map representation to perform better spatial reasoning in the complex task of language-guided navigation.

There are also hybrid maps that combines grids, \revision{dense geometric}, and topological maps.
StructNav~\citep{chen2023structnav} builds such a hybrid map where the spatial grid stores occupancy information, a scene graph stores landmarks with their connectivity, and a 3D semantic point cloud where each 3D point in the environment has a semantic label associated with it. Thus the spatial grid allows for obstacle avoidance and low-level path planning, the scene graph allows high-level reasoning about the relationship among the landmarks, and the 3D point cloud allows for a more dense semantic and spatial matching. 

It is also possible to build maps that capture the semantic hierarchy of scenes at various levels of abstraction~\revisedcitep{galindo2005multi,pronobis2012large,hemachandra2014learning,Werby-RSS-24} that allows for various levels of reasoning. \citet{armeni20193d} represent a static scene at multiple levels of hierarchy (buildings, rooms and objects), where entities are represented as nodes in a hierarchical graph and connected by edges representing coordinate frame transformations. \cite{tang2025openin} (OpenIn) build a similar hierarchical scene graph to track objects in dynamically changing indoor environments. 
\citet{Rosinol20rss-dynamicSceneGraphs} (Dynamic Scene Graphs) too build layered scene graphs to track moving agents and objects in addition to building a dense 3D metric spatial map by modeling spatio-temporal relations between objects and agents.
% They build separate scene graphs for various concepts such as objects, rooms, buildings, as well as dynamic elements such as moving humans. Such a hierarchical structure allows for various levels of reasoning.
\citet{hughes2024foundations} demonstrate that hierarchical scene representations scale better than flat representations in a large environments and thereafter introduces a system called Hydra that incrementally builds 3D scene graphs from sensor data in real-time.
\citet{fischer2024multi} introduces a multi-level scene graph representation of large-scale dynamic urban environments from a set of images captured from moving vehicles and proposes a new view synthesis benchmark for urban driving scenarios.

\mypara{Summary.}
Different types of map structures have their own strengths and limitations. While the coarse topological map is useful to represent significant landmarks in an environment, fine-grained metric maps are useful to represent its precise geometry and \revision{dense geometric map} is useful to represent an even more dense 3D geometry of objects in the scene. Therefore it's crucial to combine two or more of these structures in order to create better representations of the environment, preferably at varying levels of semantic abstraction.
While hybrid maps have been explored in robotics, they still remain under-explored in embodied AI.
However, weaknesses of these maps need to be considered carefully before combining them. For example, topological and \revision{dense geometric maps} scale better than a grid map with larger environments, and \revision{dense geometric map} needs the most and topological map needs the least memory to be stored.

\section{Map encoding}
\label{sec:representations}
\begin{figure}[ht]
\centering
% \framebox[0.9\linewidth]{
% \fbox{\rule[-.5cm]{0cm}{4cm} \rule[-.5cm]{4cm}{0cm}
\includegraphics[width=0.8\linewidth]{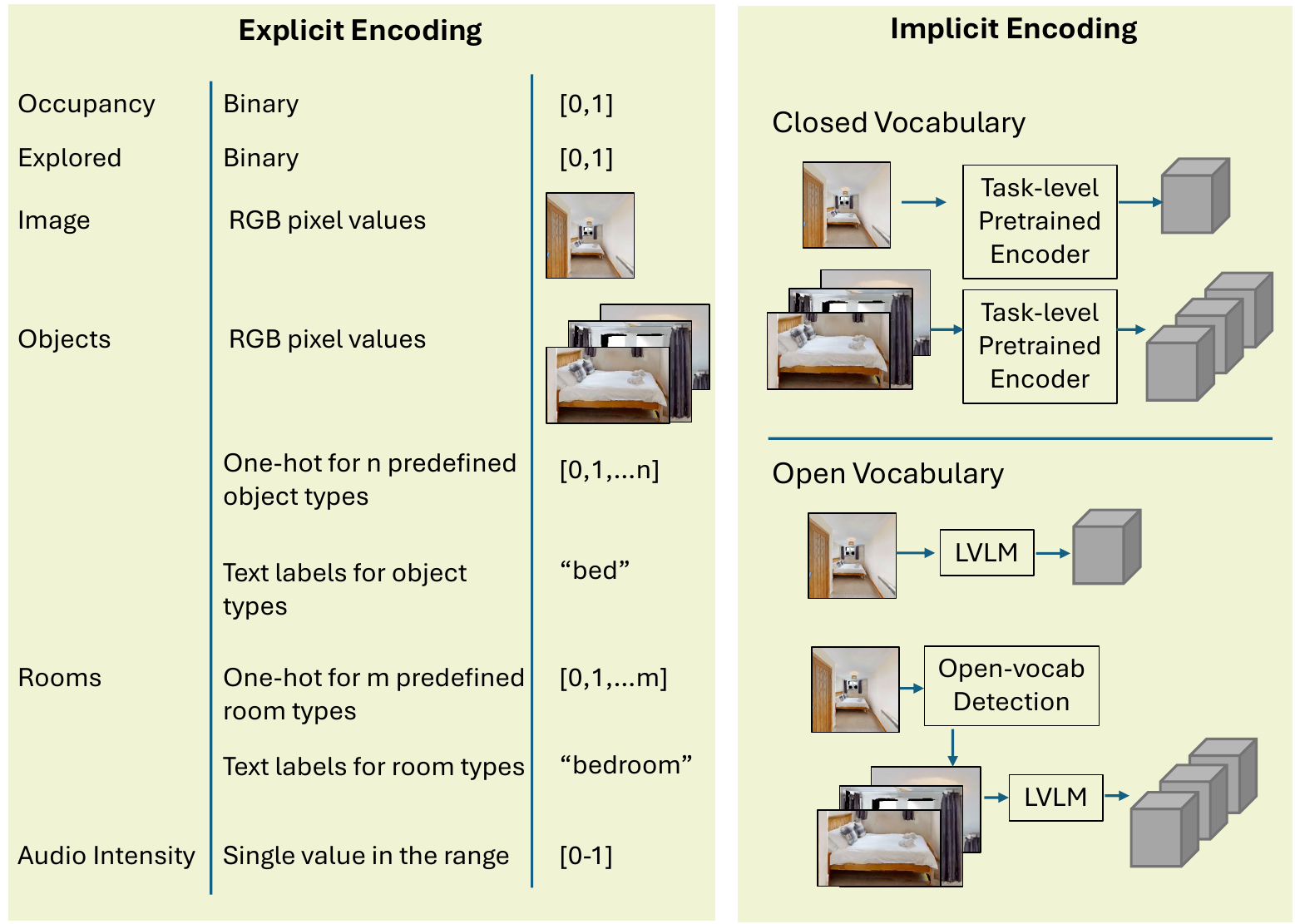}
% }
\caption{
\textbf{Map Encoding} refers to the values stored in the map and can be either \textit{explicit} or \textit{implicit}, depending on whether the information is hand-selected or a learned feature representation of the observation.
}
\label{fig:map_encoding}
\end{figure}

% \todo{let's rename `map representation' to `map encoding'.  The text has been somewhat updated but should be checked for consistency.  The term representation can apply for both structure and encoding}
In this section we will discuss different ways information is encoded and stored into semantic maps.
Irrespective of how the map is structured, the map encoding, \ie the values stored in the map can be either \textit{explicit} or \textit{implicit}. An explicit map encoding is one where the type of information stored is clearly known. An implicit encoding, on the other hand, uses a feature embedding that may not be directly interpretable. We now summarize various works that explore these two types of encodings (\Cref{fig:map_encoding}).

\subsection{Explicit encoding}
\label{sec:explicit}
% In order for the map to be useful across a variety of tasks, it needs to be built as part of a \mapping module in a modular pipeline (refer to \Cref{sec:modular_methods}). This allows the map to be learned independently of the other modules. 
% In such cases, the map stores explicit information about the semantics of its observed environment, thus allowing for better interpretability.
Many prior works store explicit information in the map, such as \textit{occupancy}, which has proven beneficial in obstacle avoidance~\revisedcitep{30720,may2009robust,chaplot2020learning,georgakis2022uncertainty}.
% Prior works have found that almost every embodied AI task benefits from the information about obstacles present in the environment and as such maintaining a spatial map that has this \textit{occupancy} information helps the agent learn to avoid obstacles on its path.
In such cases, the spatial map stores a binary occupancy value of $1$ or $0$ depending on whether the corresponding location in the physical environment is `occupied' by objects or not.
% The type of semantic information stored in a spatial map depends on what type of information is useful for the agent to perform the downstream tasks. 
% \input{chapters/figures/fig-map-active-neural-slam}

In an exploration task, however, the agent needs to maximize the explored area in an environment while being efficient, in which cases, the information of whether a location has already been explored encourages the agent to explore unexplored areas more. Active Neural SLAM~\citep{chaplot2020learning} 
% (\Cref{fig:map_active}) 
stores the \textit{explored} (binary) information in addition to the occupancy information. 
Active Neural SLAM consists of several modules connected together to perform the task of Exploration to maximize the explored area. The `Neural SLAM' module takes as inputs the visual observations and agent pose and outputs a top-down egocentric spatial map by learning to predict occupancy and explored information using a binary cross-entropy loss. 
All the modules are jointly trained with different losses. 
VLMNav~\citep{goetting2024end} similarly stores \emph{explored} information in a top-down voxel map to demonstrate its generalizability to downstream navigation tasks.
% \todo{describe other modules}
% \input{chapters/figures/fig-map-semexp}

However, in more complex tasks such as ObjectNav, which requires the agent to navigate to a particular semantic category in its environment, storing occupancy and explored information alone might not be sufficient. In such cases, it is important for the agent to identify the semantic category of the object~\revisedcitep{meger2008curious,georgakis2021learning,chaplot2021seal,raychaudhuri2023mopa,khanna2024goat,zhang2025mapnav}. Goal-oriented Semantic Exploration (SemExp) by \citet{chaplot2020object} additionally stores the semantic class labels of the objects that the agent identifies through its visual observations. SemExp uses a MaskRCNN \citep{he2017mask} on the RGB observations to predict the semantic categories of the objects and then project these on the map using depth observations. It aggregates the occupancy and explored information using element-wise max pooling whereas the semantic categories are overwritten by the latest prediction. A denoising network is then used to get the final map.
This work demonstrates that using this semantic map to predict a long-term goal helps the agent find the goal object category more efficiently. The mapping module is trained using supervised learning with cross-entropy loss on the predicted semantic map.
Similarly, a semantic categories map also helps a more complex longer-horizon task of MultiON. MOPA~\citep{raychaudhuri2023mopa} shows that maintaining a memory with objects that the agent observes while moving around is crucial to perform this task efficiently. 
However, a map built by segmenting the image first and then projecting may result in ‘label splattering’ \ie noisy category labels splattered across multiple grid cells in a spatial map. This arises mostly due to noisy depth observations and might negatively affect agent performance in a task.
% in the tasks where success is measured by how close the agent stops from the goal object.
% that happens when there are prediction errors around the object boundaries. 
Semantic MapNet \citep{cartillier2020semantic} 
% (\Cref{fig:map_semmapnet}) 
finds that projecting encoded features on to a map and then segmenting produces a more noise-free map. They show that this map can be then be applied to two different tasks effectively. However this method requires an additional exploration phase where the agent effectively explores the environment to build the map first and then use that map in the downstream task.
A recent work GOAT~\citep{chang2023goat,khanna2024goat} shows that having an object instance map helps to navigate to a goal specified by either language, image or a category label. They achieve this by storing raw images and later using CLIP~\citep{khandelwal2022simple} for image-to-language matching and SuperGlue~\citep{sarlin2020superglue} for image-to-image matching.
MapNav~\citep{zhang2025mapnav} builds an annotated semantic map by storing text labels of the segmented objects and shows that this helps a VLM to ground objects better.

While a map with semantic categories help in object navigation, an acoustic map storing audio intensity is found to be useful in the Audio-Visual Navigation task~\citep{chen2020learning}. Here the map is aggregated by averaging the intensity.
In the more complex Interactive Question Answering task, \citet{gordon2018iqa} find that storing object detection probabilities in a spatial map helps agent performance. They use a GRU recurrent memory to aggregate the current map with the previous one.

While the above works build explicit spatial maps, some works also build explicit topological maps~\revisedcitep{choset2001topological,blochliger2018topomap,chen2022fast,salas2013slam++,duvallet2013imitation,patki2019inferring,arkin2020multimodal,raychaudhuri2025zeroshotobjectcentricinstructionfollowing,chang2025ashitaautomaticscenegroundedhierarchical}. \citet{kwon2021visual} store the visitation timestep in the graph node in order to encode a temporal relation between visited locations. This information is then replaced by the latest visitation timestep while aggregating the map.

\paragraph{Summary.}
The advantage of explicit map encoding is its interpretability and the fact that it allows investigating the type of information that is beneficial for various downstream tasks. However, the type of semantic information to be stored is a design choice based on the task.
Moreover, the above approaches require a predefined set of categories to be mentioned beforehand to the mapper. This restricts the map to store only a limited number of object categories.

% Prior works on longer-horizon tasks such as MultiON \citep{wani2020multion} also finds that a map-based agent trained end-to-end with RL performs better than a no-map agent and that a semantic map that stores object labels is more useful than an occupancy map.

\subsection{Implicit encoding}
\label{sec:implicit}
Implicit maps store latent features in a semantic map. While most of the prior works use extracted features from a vision model pre-trained on a closed-vocabulary set of object categories, recent methods use features extracted from a pre-trained large vision-language model thus producing flexible open-vocabulary queryable maps.  It is also possible to store features that are not necessarily queryable with language, but captures the visual information at that location.  One example is RNR-Map~\citep{kwon2023renderable} which uses a grid map with latent codes that corresponds to a neural field that can be used to render possible views at that location.

\subsubsection{Closed-vocabulary encoding}
\label{sec:closed-vocab}
These features can be learned from scratch during training. For example, \citep{wani2020multion} learn image features using CNN blocks and use them to build a global spatial map of the environment. This is trained end-to-end to predict actions in the MultiON task.
On the other hand, the features may also be extracted from a vision model such as ResNet, pre-trained on the ImageNet~\citep{deng2009imagenet} data to encode RGB images.
For example, \citep{gupta2017cognitive} introduces Cognitive Mapper and Planner (CMP) that uses a pretrained ResNet-50 model to encode the egocentric RGB images and then projecting on the map using a differentiable mapper module. In CMP, the learned map encoding is not explicitly supervised but learned in conjunction with a differentiable planner. This enables the mapper to learn to store information that is most useful for the planner to perform the task efficiently.
Also the map is accumulated over time meaning that the map from one navigation step is integrated into the next using a differentiable warp. 
% However a global allocentric map is found to be more beneficial than an egocentric map to localize the agent with respect to a global representation of its environment. This also allows the agent to reason about the spatial structure of its environments.
Similarly MapNet \citep{henriques2018mapnet} also uses a pretrained ResNet-50 model to extract image features but they build an allocentric global map of the environment instead of an egocentric map. They do so by first ground projecting the features and then registering these into a global allocentric map, updating the map at every navigation step. This model is learned end-to-end on the task of localization and trained with a series of RGB-D observations and the corresponding ground-truth positions and orientations.

While the above methods build a spatial map, other methods use a similar approach to store implicit features in the nodes of a topological map. Each node stores encoded features from the observations at a particular location in the environment. 
% The visual observations at a particular location is passed through an image encoder to obtain visual features which can be stored in the appropriate node. 
Here too, using a pretrained ResNet encoder to extract RGB image features is a popular choice among prior works~\citep{chaplot2020neural,chen2021topological}. 

\mypara{Summary.}
The advantage of using a pretrained ResNet model over learning from scratch is that it has already been trained to encode useful features and is thus sample efficient.
%The advantage of an implicit map encoding is that it allows the model to learn to store features inside the map, which are the most useful for the task. 
% However since they are learned end-to-end for a particular task, they do not transfer to other tasks without having to be re-trained. 
% \todo{how much training is needed}
% However these maps are difficult to interpret. It is hard to make sense of what the map actually captures. The above works simply assume that the map encodes free space.
However, using a pre-trained ResNet is limited by the number of object categories it was trained on.

\begin{table*}[ht]
\caption{
\textbf{Open-vocabulary maps.} Various works build open-vocabulary semantic map using trained as well as off-the-shelf pretrained models in both simulation and real-world robots. These methods use either heuristics-based or LLM-based planner to perform the downstream taskss.
% \todo{some details about aggregation}
}
\label{tab-open-vocab-map}
\centering
    \resizebox{\linewidth}{!}{
    \begin{tabular}{llllp{2.2cm}p{3.2cm}}
    \toprule
    \bf Method 
    &\bf Environment 
    % &\bf Dataset 
     % &\bf Spatial &\bf Topological
    &\bf Training 
    % &\bf Evaluation
    &\bf VL Encoder
    &\bf Task Planner
    &\bf Aggregation
    \\
    
    \midrule
    CoW~\citeyearpar{gadre2023cows} &Habitat, RoboTHOR 
    % &Pasture 
    % &DD-PPO for 60M steps 
    &\cmark & CLIP & A* & highest similarity score \\
    
    VLMap~\citeyearpar{huang2023visual}  &Habitat, robot 
    % & MP3D 
    & \xmark & LSeg & A* & mean \\

    NLMap~\citeyearpar{chen2023open} &robot 
    % &Custom 
    & \xmark &CLIP, ViLD & LLM-based & multi-view fusion \\

    ConceptGraphs~\citeyearpar{conceptgraphs} &AI2Thor, robot 
    % &Custom 
    & \xmark &CLIP, DINO & GPT-4 & highest similarity score\\

    VoxPoser~\citeyearpar{huang2023voxposer} &Sapien, robot 
    % & Custom 
    & \xmark &OWL-ViT & GPT-4 & - \\

    VLFM~\citeyearpar{yokoyama2023vlfm} &Habitat, robot & \xmark &BLIP-2 &pretrained PointNav & highest similarity score\\

    CLIP-Fields~\citeyearpar{shafiullah2023clip} &Habitat, robot 
    % &Custom, HM3D 
    &\cmark &CLIP &SLAM & weighted mean \\

    \revision{LM-Nav}~\citeyearpar{shah2022lmnav} &robot &\xmark &CLIP &modified Dijkstra &CLIP similarity \\
    \revision{ConceptFusion}~\citeyearpar{conceptfusion} &AI2Thor, robot &\xmark &CLIP & LLM-based &weighted mean \\

    \revision{Le-RNR-Map}~\citeyearpar{taioli2023language} & Habitat &\cmark & CLIP &Random walk exploration &latest \\
    
    \revision{HOV-SG}~\citeyearpar{Werby-RSS-24}  &Habitat, robot &\xmark & CLIP &A* &weighted mean \\
    \revision{RoboHop}~\citeyearpar{garg2024robohop} &Habitat, robot &\xmark &CLIP &Dijkstra & mean \\
    \revision{Clio}~\citeyearpar{maggio2024clio} &robot &\xmark &CLIP &Dijkstra & mean  \\
    OneMap~\citeyearpar{busch2024mapallrealtimeopenvocabulary} &Habitat, robot &\xmark &SED &A* & weighted sum with uncertainty-based weights \\

    \revision{OVL-Map}~\citeyearpar{wen2025ovl} &Habitat &\cmark &LSeg &DD-PPO~\citeyearpar{wijmans2019dd} &weighted mean \\

    \bottomrule
    \end{tabular}}
\end{table*}
\subsubsection{Open-vocabulary encoding}
\label{sec:open-vocab}
\begin{figure*}[h]
\begin{center}
\includegraphics[width=\linewidth]{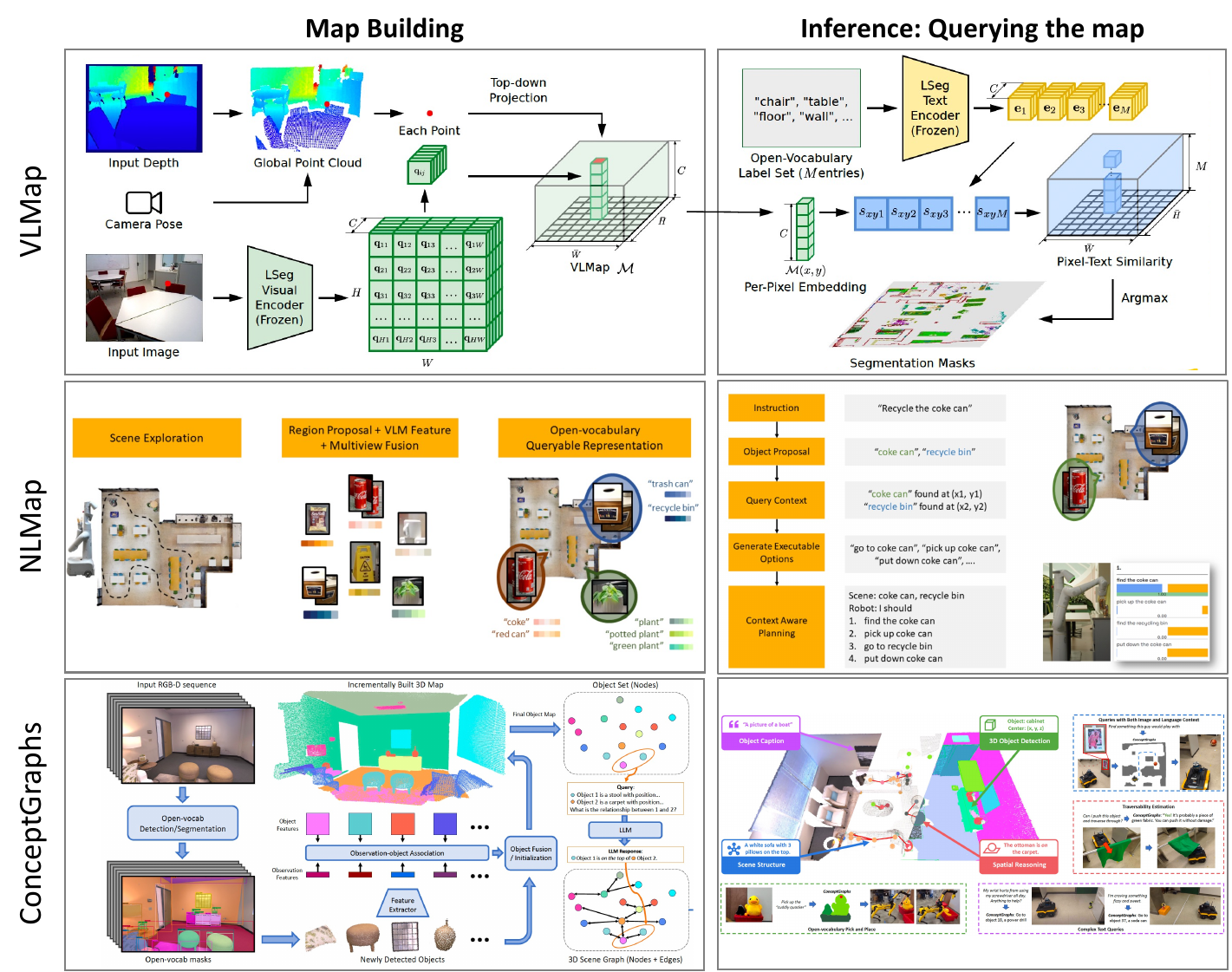}
\end{center}
\vspace{-10pt}
\caption{
\textbf{Open-Vocabulary map building.} 
There has been a growing interest to build flexible open-vocabulary maps which can be built once and then used in various downstream tasks during inference. VLMap~\citep{huang2023visual} and NLMap~\citep{chen2023open} structure their maps as spatial maps while ConceptGraphs~\citep{conceptgraphs} build a topological map.
\textit{(Figures reproduced from paper.)}
}
\label{fig:open_vocab}
\end{figure*}

% The map building methods discussed so far are restricted by a predefined set of object categories. However 
The limitation of the closed-vocabulary encoding can be mitigated by extracting features from a Large Vision-Language Model (LVLM) that was jointly trained on a vast amount of internet data of images and their text captions, such as CLIP~\citep{radford2021learning}. This allows the map to store information about 
% and the growing interest in developing general-purpose robotic agents needs for them to identify and reason about
`any' object in the environment and eventually be queried via an open-vocabulary text query~\revisedcitep{wen2025ovl} not limited to a predefined set of object categories. 
% Such objects are described via a natural language instruction. 
For example, an agent may be asked to `find a \textit{red and blue striped zebra toy} in the children's room'. It is likely that the agent has not seen a `red and blue striped zebra toy' during training but can leverage a LVLM to reason about its prior knowledge about zebras, colors and rooms in general.
Moreover, recently large language models (LLMs), such as GPT-4~\citep{openai2023gpt4}, have been shown to be able to perform complex task planning. 
% are now capable of such open-vocabulary language understanding and general purpose semantic reasoning.
Thus open-vocabulary maps built using LVLM along with LLM-based planners~\revisedcitep{taioli2023language,huang2023voxposer,longinstructnav} have led to a recent line of works in embodied agents (\Cref{tab-open-vocab-map}, \Cref{fig:open_vocab}). 
% This allowed current researchers to explore how to use the pretrained LLMs in mapping, planning and instruction following in embodied agents. 
% \Cref{tab-open-vocab-map} summarizes various methods that attempts at building such open-vocab maps.
% which eliminates the need for training separate models for the downstream tasks for millions of steps. 
% Language-driven zero-shot object navigation (L-ZSON) by \citep{gadre2023cows} requires the agent to navigate to an object described in a natural language instruction.
% A popular large vision-language model is CLIP \citep{radford2021learning} which is trained on a wide variety of (image, text) pairs from the internet. 

A pretrained CLIP model can be used to compute similarity scores between an input image and a natural language description with the highest score corresponding to the most likely image-text match. 
% CLIP has been effective to detect `any' object in an environment.
CLIP has been successfully used in map building where the map stores the similarity scores between each image that the agent observes and the language instruction describing an object. Popularly, these maps are called \emph{value maps}.
CoW \citep{gadre2023cows} shows that such a 2D value map can be successfully applied in the downstream task of language-driven ObjectNav in a zero-shot manner without any re-training. The planner in this method plans a path to the object when the stored similarity score exceeds a certain threshold. 
VLFM~\citep{yokoyama2023vlfm} follows a similar strategy to perform ObjectNav task by using the BLIP-2~\citep{li2023blip} 2D value map to semantically explore the environment. 
InstructNav~\citep{longinstructnav} extends this idea to enhance semantic value maps with multi-sourced value maps encoding actions, landmarks, and navigation history, thus improving generic instruction following.
VoxPoser~\citep{huang2023voxposer}, on the other hand, builds a similar 3D value map to efficiently perform table-top robot manipulation.
Such map encoding is quite powerful compared to the previous encodings that stores the semantic labels for a predefined set of objects. However, this particular map still lacks the ability to perform spatial reasoning because it performs similarity matching to the entire input image ignoring the semantic information about individual objects in the scene.

To mitigate this issue, researchers need to develop methods that identify where objects are located in the image, and then extract features for those objects.
% The focus of the current research is on building general purpose queryable map representations which can be used in a variety of downstream Embodied AI tasks.
% To do this, the semantic map needs to store dense features corresponding to `all' the objects the agent sees in its current visual image which can later be used to compute similarity scores based on a language query in order to select the object with the highest score.
% There are two challenges here -- (1) to understand where the objects are present in an image, and (2) what kind of features will be beneficial for the agent to be stored.
% One way to achieve both 
One way to achieve this is to store feature embeddings for each pixel in the input image. 
% such that the information about all objects present in the image are stored in the map. 
This can be done using  LSeg~\citep{li2022languagedriven} which outputs pixel-level embeddings given an image as is done in VLMaps~\citep{huang2023visual}. The embeddings are then projected into a 2D spatial map using depth observations. When multiple points are projected onto the same grid cell on the map, VLMaps aggregates by averaging the features. During inference, they extract object names from the language query and calculate pixel-text similarity scores on the map to retrieve the objects of interest.
However, these pixel-level embeddings are very dense and can be redundant since not all pixels in an image contain objects. Second, some information might be lost while averaging the pixel-level embeddings. A third issue is that it does not encode object-level semantics, thus ignoring information about spatial relationships.
OneMap~\citep{busch2024mapallrealtimeopenvocabulary} stores patch-level features extracted from SED~\citep{xie2024sed} with a hierarchical encoder-based backbone which has shown to capture spatial information better than the transformer-based architecture. This alleviates the issues of using pixel-based features.

However, a better way is to first identify all objects present in an image, which can be done by using a class-agnostic region proposal network to propose regions of interest (objects) in an image. This approach is used in NLMap~\citep{chen2023open} which uses ViLD~\citep{gu2021open} as its class agnostic region proposal network. For each region of interest, they extract image embeddings using an ensemble of CLIP and ViLD. These features are then stored in a 3D spatial map along with the 3D location and estimated size of the object. They show that this spatial map representation can be applied to any downstream task by performing a natural language based query. This is achieved by extracting object names from the query and using them to select the object from the map with the highest similarity scores. 
% This is then used by an LLM planner to generate executable code to plan and execute robotic actions.
% VLMaps~\citep{huang2023visual}, build open-vocabulary semantic maps. They use pixel-level embeddings using the visual encoder from the pretrained LSeg~\citep{li2022languagedriven} model, which maps an image to language in a way that each pixel embedding lies in the CLIP~\citep{radford2021learning} feature space. They store these pixel-level embeddings along with the corresponding 3D location in a 2D topdown grid map by projecting them using the depth observation. They average the pixel embeddings when multiple pixels are projected on to the same grid and also across multiple views. During inference, they first convert the language instruction to obtain a list of landmarks and generate a series of executable Python robot code using an LLM. They calculate the similarity of the landmark embeddings with the stored map embeddings to localize the landmark on the map. 

% \input{chapters/figures/fig-open-vocab-conceptgraphs}
While NLMap builds a 3D spatial map, 
% Open-vocabulary maps can also be structured as a graph-like topological map which has an advantage of consuming less memory than the dense point-cloud based NLMap. This topological map has nodes encoding geometric and semantic features of the objects in the scene and edges encoding their relations with other objects. 
many~\revisedcitep{shah2022lmnav,garg2024robohop} build a topological map following a similar approach. For instance, ConceptGraphs~\citep{conceptgraphs} is a recent approach that retrieves the objects of interest from the input image by using the class-agnostic 2D segmentation model, Segment-Anything (SAM)~\citep{kirillov2023segment} to obtain candidate masks. These objects then form the nodes of the topological map.
The image features for each object can then be extracted by CLIP and DINO \citep{oquab2023dinov2} to be stored in the corresponding node. Additional information about the objects can also be stored in each node. For example, ConceptGraphs store the point cloud of the proposed masks and a caption for each object as obtained by using LLaVA \citep{liu2023visual} and GPT-4 \citep{openai2023gpt4} along with a point cloud obtained by projecting the object mask proposed by SAM into 3D space.
After the nodes are formed, edges between the nodes can be constructed depending on whether the objects are spatially related. In ConceptGraphs, spatial relation between two objects is determined by whether the point clouds of the respective objects have a geometric similarity or overlap. In other words, when a certain proportion of points in point cloud of one object lie within a distance threshold of that of the second object, the objects can be said to be spatially related to each other and an edge is constructed between the corresponding nodes. The edges additionally store a spatial relationship description obtained from LLM.
When a newly detected object is found to be similar to a node, it is updated with averaged object features, union of point clouds and the latest object caption. In case it is not similar, a new node is added to the graph.
% These nodes and edges form the topological map for the entire scene and can be queried during inference in the downstream tasks. 
ConceptGraphs show that a single map representation built in this manner can be successfully and effectively used in object grounding, robot navigation and robot manipulation tasks.

\mypara{Summary.}
The advantage of an open-vocabulary map encoding is that it can be built once and then transferred to several different downstream tasks.
It can be queried using an open-vocabulary text effectively and is highly interpretable.
However one current limitation is that the computational costs of using large foundation models can be significant. 

% ConceptGraphs~\citep{conceptgraphs} also proposes an open-vocabulary scene representation which can be queried using language instructions. This choice is motivated by the huge memory consumption of a dense 3D map representation and lack of a structure that allows dynamic scene updates. Similar to \citep{chen2023open}, they first obtain candidate masks using a class-agnostic 2D segmentation model, Segment-Anything (SAM)~\citep{kirillov2023segment} which form the nodes of the graph. 
% Next they use a large vision-language model (LVLM) to caption each node and an LLM to obtain relationships between neighboring nodes, which form the edges in the graph. Each mask is next associated with visual semantic features obtained from CLIP image encoder.
% and DINO~\citep{oquab2023dinov2}. Next these masks are projected to 3D point clouds and clustered into regions using DBSCAN clustering.
% Aggregation - The objects are then captioned by feeding object crops from 10 different views to a large vision-language model (LVLM) with a prompt to describe the object. For this they use LLaVA~\citep{liu2023grounding}. They use an LLM, GPT-4~\citep{openai2023gpt4} (gpt-4-0613), to summarize this caption for accuracy. During inference, they use the ConceptGraphs along with an LLM to localize the object in the query by passing the object information (3D bounding box location and caption) in a json format to the LLM asking to find the most relevant object.

\section{Evaluation}
\label{sec:evaluation}
In the embodied AI and robotics literature, very little focus has been given to evaluating the built map \revision{(\emph{intrinsic evaluation})}. Instead the focus has been to evaluate the agent performance on the downstream tasks \revision{(\emph{extrinsic evaluation})} through various metrics. 
This is because these works build semantic maps as an intermediate step while performing downstream tasks, such as localization, navigation, exploration, manipulation and so on. 
Moreover, most of these works focus on a single task and hence it suffices for them to evaluate the task performance directly without caring about how well the map representation is.
However, we think that it is crucial to do intrinsic evaluation based on the \textit{accuracy}, \textit{completeness}, \textit{consistency} and \textit{robustness} of the built map, in addition to evaluating the task performance.
% its \textit{utility} on the downstream tasks. 
We next discuss the evaluation metrics (both extrinsic and intrinsic) in detail.

\subsection{\revision{Extrinsic evaluation}}
\revision{
This section discusses extrinsic (or task-level) evaluation metrics commonly used to assess agent performance across embodied tasks. In navigation, success is not simply reaching the goal but signaling it intentionally, often via a dedicated action like ‘Stop’ or ‘Found’~\citep{anderson2018evaluation}. Accordingly, Success Rate (\emph{SR}) is used to evaluate whether the agent correctly signals goal completion, while Success weighted by inverse Path Length (\emph{SPL}) accounts for trajectory efficiency. These metrics are standard in navigation benchmarks~\citep{Beeching2020LearningTP,batra2020objectnav,chaplot2020object,chaplot2020neural,wijmans2019dd,li2021ion,krantz2022instance,georgakis2022uncertainty,taioli2024collaborative,yokoyama2024hm3d}. Additional metrics include Oracle SR (\emph{OSR}), which accounts for when the agent reaches the goal but fails to signal, and Navigation Error (\emph{NE}), which measures the final distance to the goal. In multi-step tasks or multi-goal navigation, metrics like \emph{Progress}~\citep{wani2020multion,khanna2024goat} or \emph{Goal-Condition Success}~\citep{ALFRED20,TEACh22} capture the proportion of successfully completed subgoals. For instruction following, normalized Dynamic Time Warping (\emph{nDTW}) is used to evaluate how well the agent’s trajectory aligns with the reference path~\citep{ilharco2019generalevaluationinstructionconditioned,krantz2020beyond,raychaudhuri2021language,raychaudhuri2025zeroshotobjectcentricinstructionfollowing}. Tasks involving classification or detection often use \emph{precision} (how often predictions are correct), \emph{recall} (how well all relevant targets are found), \emph{F1 score} (combines and balances precision and recall) and receiver-operating characteristic curve or \emph{ROC} curve (measures the tradeoff between sensitivity or true positive rate and specificity or false positive rate)~\citep{Chen_Mooney_2011,5152712,Matuszek2012LearningTP,Tellex2011UnderstandingNL,Wellhausen2020SafeRN} to evaluate prediction quality. In exploration tasks, \emph{coverage} is used to measure the proportion of the environment observed~\citep{chaplot2020learning}.
In manipulation tasks such as object rearrangement or tabletop operations, \emph{task success rate}, measuring the percentage of trials where the agent achieves the desired manipulation (e.g., grasping, placing), is the dominant metric~\citep{Kalashnikov2018QTOptSD,batra2020rearrangement,lin2020learning,Zeng2020TransporterNR,huang2023voxposer,gu2023maniskill2}. Some works also evaluate policy outputs using precision~\citep{Zeng2017RoboticPO}, or employ pose error metrics to assess deviations between the final and target object positions~\citep{Liu2012FastOL}.
}

\revision{\mypara{Summary.} Extrinsic evaluation metrics have been extensively studied across the literature~\citep{deitke2022retrospectives} and are continually evolving in response to the growing complexity and diversity of embodied tasks. As new challenges emerge in navigation, manipulation, and instruction following, the need for nuanced and task-specific performance metrics has driven the development of more robust and comprehensive evaluation frameworks. Although extrinsic metrics have dominated evaluation in the field, we believe it's time for the community to place greater emphasis on intrinsic metrics (we discuss this in next section). This shift is especially important given the recent progress in building open-vocabulary, flexible, and general-purpose maps, which require more nuanced assessment of map quality beyond task success alone.
}

% \mypara{Utility.} 
% Most works build semantic maps as an intermediate step while performing downstream tasks, such as navigation, exploration, manipulation and so on. In such cases, the map is exploited by the task planner to plan a path and generate low-level actions in order to complete the task. 
% Moreover, most of these works focus on a single task and hence it suffices for them to evaluate the task performance directly without caring about how well the map representation is.
% \cite{gupta2017cognitive,kim2023topological} build map for navigation tasks, while \cite{chaplot2020learning} builds map to efficiently explore the environment. 
% \citet{kim2023topological} builds a topological map based on which it learns to generate actions using an action policy network towards the ImageNav task.
% Although the open-vocabulary map in \citet{conceptgraphs} is pre-built once and used in multiple downstream tasks (navigation, manipulation, object segmentation etc.), they still evaluate the task performance and not the map itself.
% All these works evaluate the downstream tasks and not the map itself.
% The navigation tasks are evaluated on Success and SPL~\citep{anderson2018evaluation}, while the exploration tasks are evaluated on the coverage area (Cov) and the percentage of area explored (\% Cov)~\citep{chaplot2020learning}.

\subsection{\revision{Intrinsic evaluation}}
\label{sec:intrinsic_eval}

\revision{
In this section, we discuss intrinsic evaluation or map-level evaluation which evaluates the map directly.
While most robotic and embodied AI systems are evaluated based on downstream task success, a few focus on directly evaluating the quality of the maps built during these tasks. However, there is no widely agreed-upon metric suite for evaluating the map representation itself. We next list the various dimensions across which we believe the map can be evaluated and discuss the gaps and future scope for each.
}

\mypara{Accuracy.}
Map accuracy refers to how accurately the map captures geometric or semantic information when compared against the ground truth. However obtaining the ground truth map can be challenging in most cases.
\revision{Some works~\citep{georgakis2022uncertainty,ramakrishnan2020occupancy} evaluate the 2D occupancy maps by using \emph{map accuracy} to measure how much of that area matches with the ground-truth, and \emph{IoU} to measure intersection over union between the built and the ground-truth map.}
While others~\citep{cartillier2020semantic,georgakislearning} evaluate the accuracy of the built 2D semantic map using semantic segmentation metrics such as \emph{pixel-wise labeling accuracy}, \emph{pixel-based F1 score}, \emph{IoU} score and a \emph{contour-based average boundary F1 score}. 
\revision{However, there is a lack of standardized metrics that can be used across various map types. For example, IoU is not directly applicable to topological maps.}
Moreover, while evaluating accuracy for semantic maps that store a fixed set of object categories is straightforward, it is tricky to evaluate semantic maps that store open-vocabulary features. As an alternative, OpenScene~\citep{Peng2023OpenScene} performs semantic segmentation on their open-vocabulary map for a fixed set of categories and thereafter evaluate it using accuracy and IoU. ConceptGraphs~\citep{conceptgraphs}, on the other hand, evaluate their open-vocabulary topological map by employing human evaluators on Amazon Mechanical Turk (AMT) to report the scene graph accuracy.
\revision{
OpenLex3D~\citep{kassab2025openlex3d} has been recently introduced to evaluate open-vocabulary 3D scene representations, by providing object labels at various granularity or precision. Similar evaluation techniques could benefit intrinsic evaluation in embodied setting as well.
The challenge to evaluate modern-day open-vocabulary maps lies in collecting precise ground-truth maps for large real-world environments, which is labor-intensive or infeasible. This limits the use of direct comparisons in order to calculate map accuracy. Moreover, accuracy in dynamic or cluttered environments is also rarely assessed. Both these provide possible future area of research.
}

% Similar metrics are used by \citet{huang2023visual} to evaluate their open-vocabulary spatial map and by \citet{Peng2023OpenScene} to evaluate their point-cloud on the task of semantic segmentation. 

\mypara{Completeness.}
Map completeness measures how completely a generated map represents an environment, encompassing both geometric and semantic coverage.
Geometric coverage refers to the fraction of environment that has been mapped whereas semantic coverage refers to the completeness of the semantic information captured in the map.
Completeness is particularly critical for tasks like search and rescue, where a thorough understanding of the surroundings is essential. Moreover, complete geometric and semantic mapping support better decision-making by reducing dependence on incomplete or inaccurate data. The extent of map completeness depends on how thoroughly the robot explores the environment during the downstream task and is closely tied to the `stopping criteria' -- a method that determines when to end the exploration process. In embodied AI, exploration typically ends when the task is deemed complete or when a predefined time budget is reached. However, reaching this time budget does not always guarantee that the map is fully complete, making the development of reliable stopping criteria an ongoing challenge in robotics as well as embodied AI research~\citep{placed2022enough,luperto2024estimating}.
Geometric coverage metric has been reported in prior works that performs the task of exploration~\citep{chaplot2020learning} in terms of the fraction of the environment explored by the robot. However, a lot of works that tackle other embodied AI tasks fail to report this metric even if exploration is a crucial part of their task~\citep{gervet2023navigating,raychaudhuri2023mopa,yokoyama2023vlfm}. Semantic coverage on the other hand relies on the presence of a detailed ground-truth map with semantic information and could be hard to obtain, as discussed in the previous paragraph `Accuracy'. 
\revision{
Due to the lack of detailed ground-truth semantic maps, semantic completeness is rarely quantified.
}
% Although some works \citep{cartillier2020semantic,georgakislearning} report semantic accuracy, none report semantic coverage of the built map.

\mypara{Consistency.}
\textit{Geometric consistency} of a map refers to how accurately the spatial structure (distances, angles, and relative positions of structures/objects) of the map represents the physical layout of the environment.
Accurate geometry ensures safe and efficient path planning and obstacle avoidance. This is crucial in classical robotics for identifying loop closure, a popular technique to recognize previously visited locations in SLAM, which refines the map and reduces drift. 
\revision{
Popular loop closure metrics~\citep{Li2021SALOAMSL,Qian2022TowardsAL} include Absolute Trajectory Error (\emph{ATE}) that evaluates the cumulative error, in addition to precision and recall.
}
However, in embodied AI systems, there is no drift due to absence of sensory and actuator noise and hence the generated maps are mostly consistent with the environment. Hence prior embodied AI works do not report geometric consistency metrics for the built map. That said, ensuring geometric consistency will be crucial in future where dynamic moving objects may disrupt the map’s structural fidelity. It can be measured by reporting the Root Mean Square Error (\emph{RMSE}) from the ground-truth geometry.
On the other hand, the \textit{semantic consistency} of a map refers to the alignment between the semantic information of structures/objects and their physical locations in the environment. Semantic consistency is particularly crucial to remain consistent over time despite changes in perspective, lighting, or environment dynamics when the robot moves around the environment. This could be measured using a temporal accuracy metric that measures how accuracy of the semantic information changes over time. However, none of the prior works measure semantic consistency and may be considered in future research.

\mypara{Robustness.}
Evaluating robustness in semantic maps is mostly crucial for assessing their reliability in unpredictable or dynamic environments. A robust map exhibits low uncertainty and high confidence in its semantic information, allowing the robot to adapt to errors, sensor noise, and environmental changes. 
\revision{In SLAM, some study system robustness to measure Absolute Trajectory Error (\emph{ATE}) and Relative Pose Error (\emph{RPE}) under noise perturbations~\citep{Prokhorov2019MeasuringRO,yang2025robustnesslidarbasedposeestimation}.}
On the other hand, since recent approaches use pretrained models, measuring the confidence of the model’s predictions could be beneficial to assess map robustness, with higher confidence indicating a more robust map. Alternatively, model uncertainty can be measured by assessing the variance in model predictions, reflecting epistemic uncertainty in the model, where low uncertainty will indicate a robust map. 
While some studies in embodied AI~\citep{georgakis2022uncertainty,raychaudhuri2025zeroshotobjectcentricinstructionfollowing} incorporate uncertainty into task planning, it has not been used as a formal metric in previous research, presenting an area for future exploration.
\revision{
We think that uncertainty-based reasoning in the SLAM and active robotic exploration can provide valuable insights into designing uncertainty-based map robustness metrics.
More specifically, \emph{joint entropy}, \emph{expected map information}, and \emph{uncertainty-aware mapping} all relate to reasoning under uncertainty in the robotics literature. Joint entropy quantifies the total uncertainty over both the robot’s trajectory and the map, while expected map information (often expressed as mutual information) captures how much that uncertainty in the map is expected to decrease after a future observation. These are widely used as internal utilities for planning~\citep{Stachniss2005InformationGE,Blanco2008ANM,georgakis2022uncertainty,Carlone2013ActiveSA}, guiding action selection in active SLAM and exploration by prioritizing trajectories that maximize expected information gain.  Uncertainty-aware mapping refers to the practice of explicitly representing and propagating uncertainty within the map itself, often using probabilistic models like occupancy grids with confidence estimates, Gaussian processes, or learned uncertainty models. This enables more robust perception and safer navigation, especially in dynamic or partially observed environments. While joint entropy and expected map information are used to decide where to go next, uncertainty-aware mapping focuses on how we represent what we’ve already seen.
However, these are rarely reported as evaluation metrics, as they are task-dependent, abstract, and computationally intensive to calculate precisely.}

\begin{table*}[ht]
\caption{
\revision{\textbf{Map evaluation.} While extrinsic or task-level evaluation has been extensively studied over the years, there has been little progress in intrinsic evaluation to assess the quality of the built map across various dimensions such as accuracy, completeness, consistency and robustness. In this study, we emphasize the need for a standardized suite of intrinsic metrics to be applied across different map structures and encodings.}
}
\label{tab-map-eval}
\centering
    \resizebox{\linewidth}{!}{
    \begin{tabular}{llll}
    \toprule
    \bf Evaluation 
    &\bf Dimensions 
    &\bf Metrics
    &\bf Task
    \\
    
    \midrule

    Extrinsic & Utility & Success rate, SPL, NE, OSR & Navigation~\citeyearpar{Beeching2020LearningTP,batra2020objectnav,wijmans2019dd,li2021ion,taioli2024collaborative,yokoyama2024hm3d}\\
    &&nDTW & Instruction following~\citeyearpar{krantz2020beyond,raychaudhuri2025zeroshotobjectcentricinstructionfollowing} \\
    &&QA accuracy, Nav accuracy &Embodied QA~\citeyearpar{eqa_matterport,Majumdar2024OpenEQAEQ}\\
    &&Precision, Recall, F1, ROC &Language grounding~\citeyearpar{Chen_Mooney_2011,Matuszek2012LearningTP,Tellex2011UnderstandingNL}\\
    &&RPE (translation, rotation) &Localization~\citeyearpar{Liu2019GlobalLW} \\

    \midrule
    
    Intrinsic &Accuracy &Map accuracy, IoU, Pixel F1, Boundary F1 & Semantic segmentation~\citeyearpar{cartillier2020semantic,georgakis2021learning} \\

    &Completeness & Geometric coverage &Exploration~\citeyearpar{chaplot2020learning} \\
    &Consistency &Precision, Recall, True Positives, ATE &Loop closure~\citeyearpar{Li2021SALOAMSL,Qian2022TowardsAL} \\
    &Robustness & ATE and RPE under noise perturbations &Visual SLAM~\citeyearpar{Prokhorov2019MeasuringRO} \\

    \bottomrule
    \end{tabular}}
\end{table*}
\subsection{Summary}
\revision{
In this section we summarize the key challenges and the future scope of evaluation (\Cref{tab-map-eval}).
Extrinsic evaluation metrics have been widely explored in the literature and continue to evolve with the emergence of increasingly complex tasks.
However, evaluating the quality of maps (intrinsic) in semantic SLAM and embodied AI continues to be understudied and presents two key challenges. First, collecting accurate, fine-grained, open-vocabulary ground-truth annotations for large, real-world environments is inherently difficult. This limitation makes it particularly challenging to assess important properties of the built maps such as semantic accuracy, consistency, and completeness of the maps. Without reliable ground truth, these evaluations often remain coarse or dataset-specific. Second, there is currently no standardized suite of evaluation metrics that can be consistently applied across different types of map structures (topological, spatial grid, dense geometric, etc.) or across different encoding schemes (explicit and implicit features). Together, these challenges highlight a critical gap in the field and point toward the need for developing scalable, generalizable map evaluation frameworks, offering an important direction for future research.
}

% There is considerable scope for future research to develop better metrics that evaluate the built semantic map and not just the downstream task performance. This can further be improved by developing a standardized evaluation framework for semantic maps.

\section{\revision{Challenges}}
\label{sec:challenges}

\revision{Despite significant progress in semantic map building, numerous challenges remain that warrant further attention and improvement. This section outlines key open problems in the field across multiple dimensions.}

\revision{\mypara{Efficiency.}
As semantic maps grow richer by incorporating fine-grained semantic information, they become increasingly data-heavy, posing significant challenges for memory efficiency and compact storage, particularly on robots with constrained hardware. The underlying map structure plays a crucial role in determining storage and computational efficiency. Different representations offer trade-offs among semantic richness, spatial fidelity, and scalability. For example, spatial grids uniformly capture space but are memory-intensive and difficult to scale in large, high-resolution environments. Dense geometric maps provide high spatial fidelity and support per-point semantic features (e.g., color, normals), yet are often redundant and storage-heavy. In contrast, topological maps are lightweight and scalable but lack the geometric detail needed for tasks like manipulation or precise localization. Hybrid topometric maps attempt to balance these aspects by combining the spatial accuracy of grids with the efficiency of topological graphs, though managing and updating both layers increases complexity and computational load. Map encoding methods also impact efficiency. Learned high-dimensional representations (e.g., neural features) offer compressed yet expressive encodings, but they are often harder to interpret or update, and computationally expensive to query. Overall, achieving efficient storage and memory usage in semantic mapping remains an open and active research challenge.
}

\revision{\mypara{Scalability.}
As semantic mapping systems are deployed in increasingly large, dynamic, and diverse environments, scalability emerges as a critical challenge. It refers to a system’s ability to accommodate growth in spatial extent (e.g., large physical spaces), semantic complexity (e.g., diverse object categories and scene types), and temporal evolution (e.g., long-term and lifelong operation), while maintaining efficiency, accuracy, and robustness. High-resolution, globally consistent maps require significant computational and memory resources, which grow rapidly with the size and richness of the environment. For example, a home robot navigating multiple cluttered rooms over extended periods must retain detailed spatial layouts and object-level information without exhausting onboard storage or processing capacity. The challenge intensifies in dynamic settings, such as when furniture is rearranged or objects are moved, requiring the map to adapt continuously. Moreover, as maps expand, core operations like loop closure, re-localization and map querying become increasingly expensive in both simulation and real robots. This underscores the need for lightweight, adaptable semantic mapping methods that can scale with environmental complexity while staying efficient on resource-constrained hardware.}

\revision{\mypara{Real-time processing.} 
Real-time processing is a critical challenge in semantic map building, especially in applications such as autonomous driving and real-time human-robot interaction, where decisions must be made within strict latency bounds. Semantic mapping in these domains involves constructing maps with semantic and spatial integrity at frame rates fast enough to support safe and responsive behavior. However, this is extremely challenging. For instance, an autonomous vehicle navigating an urban environment must simultaneously detect traffic signals, update the map with moving vehicles or pedestrians, and plan a safe trajectory in milliseconds. Any lag in perception or map update could lead to disastrous decisions. Balancing the trade-off between semantic richness and processing speed, often on memory-constrained hardware, remains an open research challenge. Advances in model compression, edge computing, and efficient data representations are being explored, but ensuring high-fidelity semantic understanding under tight latency requirements continues to be a major hurdle for real-world deployment.}

\revision{\mypara{Noise and uncertainty.}
Handling noise and uncertainty is a fundamental challenge in semantic map building, particularly when operating in real-world environments characterized by sensor imperfections, dynamic changes and ambiguous semantics. Noise from cameras, LiDARs, or depth sensors can lead to inaccurate observations, while errors in object detection, segmentation, or classification models may introduce semantic inconsistencies in the map. For example, a robot navigating a cluttered indoor environment might repeatedly misclassify similar-looking objects (mistaking a stool for a chair) or fail to detect partially occluded objects or objects in poor lighting. Over time, these errors can compound, leading to semantic drift or contradictions in the map. Furthermore, the uncertainty in sensor data and learned representations is often not explicitly modeled or propagated, which limits the system's ability to reason about the confidence of its predictions. This becomes especially critical in safety applications where overconfident yet incorrect map entries can mislead planning and decision-making. While some approaches use Bayesian filters or probabilistic representations to quantify uncertainty in robotics~\citep{Blanco2008ANM,georgakislearning}, these are computationally expensive and may not scale well. Moreover, semantic mapping methods in embodied AI often assume noiseless sensors in simulated environments, and thus do not transfer well to real robots. Thus, building robust semantic maps that explicitly model and manage uncertainty remains a key open problem in both robotics and embodied AI.
}

\revision{\mypara{Multi-modal fusion.}
Recent semantic map building approaches increasingly rely on fusing diverse sensory inputs, including vision, depth, audio, natural language, speech etc., to construct rich interpretable representations of the environment. However, robust multi-modal fusion remains a core challenge, with alignment and integration between the modalities a non-trivial problem. For instance, visual inputs provide spatial and appearance cues, while natural language and speech often provide high-level semantic cues that may be ambiguous, indirect or context-dependent (e.g., `go to the room where the baby is sleeping'). Moreover such grounding is often compounded by noisy or incomplete observations. Audio cues such as footsteps, conversations or object sounds offer valuable environmental context but are transient and hard to spatially localize or persistently encode.
Moreover, multi-modal fusion needs to be done in a way that supports downstream reasoning and generalization in unseen environments. For instance, enabling agents to query the semantic map for complex spatial reasoning (e.g. `the cup behind the fruit bowl', `the vase between the kettle and the dish') and affordances (e.g. `where can I sit', `which objects are fragile').
The emergence of multi-modal foundation models~\citep{alayrac2022flamingo,Driess2023PaLMEAE,openai2023gpt4} offers new promise in addressing some of these challenges. These models are capable of aligning and reasoning across modalities, enabling semantic parsing of complex instructions, visual understanding and basic spatial reasoning. However, they are not trained explicitly for map construction or long-term memory integration, limiting their direct applicability in real-time embodied settings. Moreover, deploying such large models on robot hardware poses computational and energy constraints, particularly in latency-sensitive domains like autonomous driving. Thus the challenge lies in building efficient multi-modal fusion strategies that also support reliable and flexible querying, making this an open research problem.
}

\revision{\mypara{Lifelong learning.}
Lifelong learning poses significant challenges for semantic mapping, particularly in real-world applications where robots are expected to operate continuously over extended periods in dynamic environments. Unlike static maps that assume a fixed world, real-world settings are inherently non-stationary (furniture is rearranged, objects are moved, and scenes evolve with time and seasons). To support lifelong learning and long-term autonomy, semantic maps must be capable of continuous adaptation, i.e. updating outdated information, integrating new observations and distinguishing between transient and persistent changes. This requires temporal reasoning, mechanisms to prevent catastrophic forgetting and the ability to resolve conflicting data collected across different times or viewpoints.
The importance of long-term autonomy lies in its potential for practical deployment, i.e. robots that can adapt to changing homes, warehouses or outdoor environments without manual reconfiguration. It is a foundational requirement for truly intelligent agents that not only act within their world but grow and evolve alongside it, building richer and more useful semantic representations over time. However, most current systems are not yet equipped with robust mechanisms for continuous semantic learning and remain limited in handling evolving, open-world environments.
}

\revision{\mypara{Standardized evaluation framework.}
A significant challenge in semantic map building lies in the absence of standardized evaluation frameworks that can consistently benchmark performance across diverse tasks, environments and map representations, as we discuss at length in \Cref{sec:intrinsic_eval}. While the community has focused more on task-specific extrinsic evaluation, limited progress have been made towards developing intrinsic measures to evaluate the map quality. This lack of standardized intrinsic evaluation hinders progress, as it becomes unclear whether improvements in downstream task success stem from better semantic understanding or other factors like better control policies. Developing general, interpretable and task-agnostic evaluation metrics is crucial, not only to track progress but also to facilitate cross-domain generalization. As open-vocabulary multi-modal queryable general-purpose semantic maps become increasingly central to reasoning in embodied AI and real-world robots, creating a unified benchmark suite that reflects both structural and semantic fidelity is essential for advancing the field.
}

\revision{\mypara{Summary.} 
Several critical challenges remain unaddressed in semantic mapping approaches across various dimensions. Key open problems include building efficient, scalable and real-time maps, which are robust to noise and uncertainty. Building multi-modal maps, that are reliable and flexible, too remains an open challenge. Maps that are able to handle lifelong learning also remain under-explored. Moreover, lack of a standardized evaluation framework require further research and innovation.
}

\section{Future research direction}
\label{sec:future}

In this section we highlight the current challenges in semantic map building and outline potential directions for future research.
Although semantic map building has advanced significantly over the past decades, several challenges and opportunities for improvement remain. The field is evolving toward creating maps that are flexible, general-purpose, open-vocabulary and queryable, enabling the same map representation to support a wide range of downstream tasks. This shift aims to make maps more versatile and suitable for complex, multi-task robotic systems.
Moreover, to enable efficient reasoning about the spatial and semantic structure of the environment, the focus is also on developing dense, scalable, and memory-efficient maps. Such maps should maintain high resolution and detailed spatial information while being computationally efficient and consistent across dynamic and large-scale environments. Achieving this balance is critical for applications that require real-time processing or operate in resource-constrained settings. 
Furthermore, the emphasis has largely been on evaluating agent performance in downstream tasks using various metrics, with limited attention given to assessing the quality of the built maps. It is however essential to evaluate semantic maps beyond their utility for specific tasks, focusing on metrics such as accuracy, completeness, consistency, and robustness to ensure they are reliable and effective for broader applications.
Next, we provide an in-depth discussion of potential future directions that we believe are most critical for advancing research in semantic mapping.

% Although research in semantic map building has seen a lot of progress, there is still room for improvement. In recent times, the community has been moving towards building a map which is flexible, general-purpose, open-vocabulary and query-able, such that the same map representation can be used in several downstream tasks. Moreover, in order to better reason about the spatial structure of the environment, the map needs to be dense but at the same time scalable, memory-efficient, consistent and computationally efficient. We present a few future directions that are the most crucial in semantic mapping research.

\subsection{General-purpose maps}
Creating general-purpose semantic maps in robotics and embodied AI is crucial for enabling robots to perform a wide variety of tasks in diverse environments with minimal reconfiguration.
% The research in semantic mapping is moving towards building a flexible general-purpose open-vocabulary and queryable structured representation so that it can be built independently and then used to plan for the various downstream tasks. 
% Current approaches have various limitations including pretrained class-agnostic object detectors being limited in detecting small and thin objects.
% Moreover, current approaches use simple text queries to query the open-vocabulary map. A more systematic investigation is needed to understand whether the current representation allows for complex spatial reasoning queries.
The idea is to design a general-purpose semantic map that serves as a single comprehensive representation of the environment, combining spatial geometry and semantic information. This eliminates the need for task-specific maps, making it easier to reuse the same map for different tasks such as navigation, object manipulation, and scene understanding.
To enable this, the maps need to be open-vocabulary that allow the robots to understand and integrate previously unseen objects using natural language descriptions. This capability broadens the scope of the downstream tasks the robot can perform, especially in unstructured or novel environments. Such semantic maps provide an opportunity to thoroughly evaluate their ability to handle textual queries involving complex spatial and semantic reasoning. Despite the recent progress towards achieving this, it still remains an open research problem.  
Open-vocabulary maps are currently limited by the pretrained class-agnostic object detectors used in building such maps. For example, these detectors often struggle with detecting small, thin or obscure objects, thus limiting the semantic maps that rely on them. Moreover, open-vocabulary object detectors that incorporate unseen classes through textual descriptions are still not perfect and lack robustness. Thus improving open-vocabulary object detectors that can recognize new objects without extensive retraining could improve the quality of semantic maps that rely on them and hence present a future research avenue.
Another challenge that general-purpose map building face is that they are computationally expensive and memory-intensive due to the rich semantic and geometric data stored in them. Balancing map detail with resource efficiency is a challenging task and impacts the ability of robots to process large areas or continuously update the map in real-time without overwhelming computational resources. This presents a future research direction worth pursuing.

\subsection{Dense yet efficient maps}
Following our discussion on various map structures, we find it crucial that the semantic map be able to capture dense visual cues to allow for complex spatial reasoning among objects.
For example, beyond addressing straightforward queries like `Where is the table?', the semantic maps can be assessed on more intricate spatial reasoning queries such as `Can you retrieve my phone from my desk beside my laptop?'. To perform such reasoning, the semantic map needs to capture fine-grained detail about the spatial arrangement of the objects (phone and laptop on the table) in the scene.
While at one end of the spectrum topological maps are too sparse failing to capture the dense semantics of a scene, at the other end the \revision{dense geometric} maps are too dense capturing redundant empty space information. Although spatial top-down 2D maps exist somewhere in the middle, they fail to capture the semantic information in the third dimension (height). This is crucial where the navigation is in 3D space, for example in drones. Hence there is still a need of a dense-enough map to capture the 3D space in its entirety at the same time intelligently ignoring empty space information. 
Additionally, a dense map will consume more memory and will be difficult to update.
So a dense map representation which is still scalable, memory and computation efficient is a research direction worthwhile to pursue.

\subsection{Dynamic maps}
Current mapping techniques in indoor environments assume that the objects present in the environment are static and only the agent is moving. Although this assumption is reasonable in an indoor environment, it is unrealistic in an outdoor setting in the presence of moving vehicles and people. This entails investigating how well current map building approaches capture moving objects effectively in a dynamic environment and focus on building efficient dynamic maps.
Building such maps involves continuous tracking and updating of objects in real-time, as they may move unpredictably. Sensor fusion, which integrates data from multiple sensors like LiDAR and cameras, is often used to detect and track these objects. However, real-time updates can be computationally intensive, particularly in high-traffic areas. Thus, efficiently storing and representing dynamic data in a way that is both memory and computationally scalable remains a key challenge and an ongoing area of research.
Moreover, the dynamic nature of these maps complicates their use in downstream tasks. For instance, a robot may need to navigate around a moving pedestrian or vehicle, which requires understanding the object's trajectory and predicting its future movements to avoid collisions. Efficiently integrating this dynamic data into decision-making processes is a significant challenge in autonomous navigation.

\subsection{Hybrid map structure}
A spatial map is able to capture the geometry of a 3D space, which helps to reason about complex spatial relations among objects and areas.
A topological map, on the other hand, lacks such geometric spatial understanding but is able to explicitly capture semantic relationships (edges) among objects (nodes).
Since both structures have different merits, there have been research to explore a `hybrid' map structure that leverages the geometric accuracy of spatial maps with the semantic and relational power of topological maps, providing a more comprehensive and efficient tool for complex reasoning tasks. For example, in a large-scale outdoor environment, a robot could use the topological map for long-range navigation to reach one building from another, while switching to a spatial map for close-range navigation to avoid obstacles or interact with objects in a room. A hybrid approach can also help balance the computational load. Topological maps are less resource-intensive and can provide high-level guidance, while spatial maps can be used for precise actions in local regions of interest, reducing the need for continuous, high-cost processing across the entire environment. Moreover, maintaining a separate level of hierarchy to track dynamic objects can also reduce computation load of frequent real-time updates. Although several approaches to hybrid mapping have been proposed in recent years, one of the main challenges in creating hybrid maps is effectively integrating both spatial and topological representations without sacrificing the quality of either. Moreover, combining the two representations requires the robot to make intelligent decisions about when to transition from one to another. Hence significant research is still needed in optimizing hybrid map building, ensuring scalability for large-scale, real-time applications and intelligent algorithms to transition between the different maps.

\subsection{Devising evaluation metrics}
As we discuss in \Cref{sec:evaluation}, the evaluating semantic maps in embodied AI research has received limited attention compared to assessing agent performance in downstream tasks. However, we believe that advancing the field requires a stronger emphasis on map evaluation using metrics such as accuracy, completeness, consistency, and robustness. Regardless of the downstream task, maps should be assessed on how well they capture semantic information (accuracy), their geometric and semantic coverage (completeness), their spatial and semantic reliability in dynamic environments (consistency), and their confidence and ability to handle uncertainty and noise (robustness). Establishing standardized evaluation metrics and frameworks for semantic maps remains a critical challenge with substantial opportunities for future research.

\subsection{\revision{Summary}}
\revision{
This section outlines promising future directions, as derived from the current challenges in semantic mapping in \Cref{sec:challenges}. As the field moves toward creating flexible, general-purpose, and queryable maps, researchers face open problems in balancing semantic richness with efficiency, ensuring scalability and robustness in dynamic environments, and developing better evaluation metrics that go beyond task performance to assess map quality directly. These provide a multitude of future research directions, which will help advance the field.
}

\section{Conclusion}
\label{sec:conclusion}

\revision{
In this survey, we review a wide range of semantic map-building approaches and categorize them based on their underlying map structure, such as spatial grids, topological graphs, dense geometric and hybrid representations, as well as the nature of semantic encoding, such as explicit or implicit. 
This perspective is timely in the light of recent advances in foundation models and the increasing need for general-purpose, multi-modal, open-vocabulary and queryable map representations.
This survey helps identify key challenges in current semantic mapping paradigms and present promising directions for future research in semantic mapping.
}

% In this survey, we explore various approaches to semantic map building in the Embodied AI literature, focusing on indoor environments. 
Existing works primarily employ spatial maps to capture geometric layouts or topological maps to model landmark-based relationships. 
While many robotics studies have explored hybrid maps that combine spatial and landmark information, this approach remains under-explored in embodied AI research. It presents a promising avenue for future work, particularly in leveraging such maps to enhance performance on complex spatial reasoning tasks.
We also discuss how dense geometric maps, created by associating semantic information with point clouds, triangle meshes or surfels, offer potential for embodied AI tasks but remain under-explored. While promising for spatial reasoning, their high memory demands, computational inefficiency, and the presence of mostly empty 3D space in indoor environments pose significant challenges.
% Both structures have their advantages and hence it is worthwhile to explore whether a `hybrid' map structure, capturing both the geometry of the space and the landmark information, is better suited for navigation and spatial reasoning.
% With recent advances in large foundation models, the focus has shifted towards building open-vocabulary, queryable map representations that are task-agnostic. 
% Despite these advances, learning to map an indoor 3D scene in simulation rely on unrealistic assumptions such as perfect localization and noiseless sensors, leading to a significant sim-to-real gap. 
Moreover, most current mapping approaches assume static environments, limiting their applicability in dynamic settings, for example in scenes where furniture is rearranged or humans move around. 

\revision{
We hope that the insights presented in this survey serve to guide and inspire further advancements in the field by the research community.}

% Building on the insights from this survey, we highlight the limitations in existing semantic map-building methods and explore promising directions for future research. We aim for this discussion to not only summarize the current state of the field but also inspire the research community to advance it further.
% We hope that this discussion will not only provide a survey of the existing literature but also encourage the community to drive the research forward. 

% In this report we discuss Spatial and Topological map structures that are commonly used across Embodied AI tasks. We look at various representations and approaches that exist in the literature and also discussed their advantages and limitations. We also discuss a recent line of work that studies open-vocabulary map representations including 2D spatial, 3D spatial and 3D scene graphs and how they are used in multiple downstream Embodied AI tasks.
% We finally present a discussion on future works that are worthwhile to pursue in this regard.

\paragraph{Acknowledgements}
The authors were supported by a Canada CIFAR AI Chair grant and an NSERC Discovery Grant. We also thank 
% \href{https://msavva.github.io/}{Dr. Manolis Savva} and \href{https://yasu-furukawa.github.io/}{Dr. Yasutaka Furukawa} 
Manolis Savva, Yasutaka Furukawa, Tommaso Campari, Austin T. Wang, Xingguang Yan, Bernadette Bucher, Duy Ta and Sachini Herath, and the anonymous reviewers for their valuable feedback on our paper.

\bibliography{bibliography}

\begin{thebibliography}{363}
\providecommand{\natexlab}[1]{#1}
\providecommand{\url}[1]{\texttt{#1}}
\expandafter\ifx\csname urlstyle\endcsname\relax
  \providecommand{\doi}[1]{doi: #1}\else
  \providecommand{\doi}{doi: \begingroup \urlstyle{rm}\Url}\fi

\bibitem[Achour et~al.(2022)Achour, Al-Assaad, Dupuis, and
  El~Zaher]{achour2022collaborative}
Abdessalem Achour, Hiba Al-Assaad, Yohan Dupuis, and Madeleine El~Zaher.
\newblock {Collaborative mobile robotics for semantic mapping: A survey}.
\newblock \emph{Applied Sciences}, 12\penalty0 (20):\penalty0 10316, 2022.
\newblock URL \url{https://www.mdpi.com/2076-3417/12/20/10316}.

\bibitem[Ahmed et~al.(2023)Ahmed, Masood, Fremont, and
  Fantoni]{ahmed2023active}
Muhammad~Farhan Ahmed, Khayyam Masood, Vincent Fremont, and Isabelle Fantoni.
\newblock Active slam: {A} review on last decade.
\newblock \emph{Sensors}, 23\penalty0 (19):\penalty0 8097, 2023.
\newblock URL \url{https://www.mdpi.com/1424-8220/23/19/8097}.

\bibitem[Alayrac et~al.(2022)Alayrac, Donahue, Luc, Miech, Barr, Hasson, Lenc,
  Mensch, Millican, Reynolds, et~al.]{alayrac2022flamingo}
Jean-Baptiste Alayrac, Jeff Donahue, Pauline Luc, Antoine Miech, Iain Barr,
  Yana Hasson, Karel Lenc, Arthur Mensch, Katherine Millican, Malcolm Reynolds,
  et~al.
\newblock {Flamingo: a visual language model for few-shot learning}.
\newblock \emph{Advances in Neural Information Processing Systems},
  35:\penalty0 23716--23736, 2022.
\newblock URL
  \url{https://proceedings.neurips.cc/paper_files/paper/2022/file/960a172bc7fbf0177ccccbb411a7d800-Paper-Conference.pdf}.

\bibitem[Ammirato et~al.(2017)Ammirato, Poirson, Park, Ko{\v{s}}eck{\'a}, and
  Berg]{ammirato2017dataset}
Phil Ammirato, Patrick Poirson, Eunbyung Park, Jana Ko{\v{s}}eck{\'a}, and
  Alexander~C Berg.
\newblock A dataset for developing and benchmarking active vision.
\newblock In \emph{IEEE International Conference on Robotics and Automation},
  pp.\  1378--1385. IEEE, 2017.
\newblock URL \url{https://ieeexplore.ieee.org/abstract/document/7989164}.

\bibitem[An et~al.(2023)An, Qi, Li, Huang, Wang, Tan, and Shao]{an2023bevbert}
Dong An, Yuankai Qi, Yangguang Li, Yan Huang, Liang Wang, Tieniu Tan, and Jing
  Shao.
\newblock {BEVBert: Multimodal Map Pre-training for Language-guided
  Navigation}.
\newblock \emph{Proceedings of the IEEE/CVF International Conference on
  Computer Vision}, 2023.
\newblock URL \url{https://github.com/MarSaKi/VLN-BEVBert}.

\bibitem[An et~al.(2024)An, Wang, Wang, Wang, Huang, He, and
  Wang]{an2024etpnav}
Dong An, Hanqing Wang, Wenguan Wang, Zun Wang, Yan Huang, Keji He, and Liang
  Wang.
\newblock {ETPNav}: Evolving topological planning for vision-language
  navigation in continuous environments.
\newblock \emph{IEEE Transactions on Pattern Analysis and Machine
  Intelligence}, 2024.
\newblock URL \url{https://ieeexplore.ieee.org/abstract/document/10495141}.

\bibitem[Anand et~al.(2013)Anand, Koppula, Joachims, and
  Saxena]{anand2013contextually}
Abhishek Anand, Hema~Swetha Koppula, Thorsten Joachims, and Ashutosh Saxena.
\newblock Contextually guided semantic labeling and search for
  three-dimensional point clouds.
\newblock \emph{The International Journal of Robotics Research}, 32\penalty0
  (1):\penalty0 19--34, 2013.
\newblock URL
  \url{https://journals.sagepub.com/doi/full/10.1177/0278364912461538}.

\bibitem[Anderson et~al.(2018{\natexlab{a}})Anderson, Chang, Chaplot,
  Dosovitskiy, Gupta, Koltun, Kosecka, Malik, Mottaghi, Savva,
  et~al.]{anderson2018evaluation}
Peter Anderson, Angel Chang, Devendra~Singh Chaplot, Alexey Dosovitskiy,
  Saurabh Gupta, Vladlen Koltun, Jana Kosecka, Jitendra Malik, Roozbeh
  Mottaghi, Manolis Savva, et~al.
\newblock On evaluation of embodied navigation agents.
\newblock \emph{arXiv preprint arXiv:1807.06757}, 2018{\natexlab{a}}.
\newblock URL \url{https://arxiv.org/abs/1807.06757}.

\bibitem[Anderson et~al.(2018{\natexlab{b}})Anderson, Wu, Teney, Bruce,
  Johnson, S{\"u}nderhauf, Reid, Gould, and van~den Hengel]{anderson2018vision}
Peter Anderson, Qi~Wu, Damien Teney, Jake Bruce, Mark Johnson, Niko
  S{\"u}nderhauf, Ian Reid, Stephen Gould, and Anton van~den Hengel.
\newblock Vision-and-language navigation: Interpreting visually-grounded
  navigation instructions in real environments.
\newblock In \emph{Proceedings of the IEEE/CVF Conference on Computer Vision
  and Pattern Recognition}, pp.\  3674--3683, 2018{\natexlab{b}}.
\newblock URL
  \url{http://openaccess.thecvf.com/content\_cvpr\_2018/html/Anderson\_Vision-and-Language\_Navigation\_Interpreting\_CVPR\_2018\_paper.html}.

\bibitem[Arkin et~al.(2020{\natexlab{a}})Arkin, Park, Roy, Walter, Roy, Howard,
  and Paul]{arkin2020multimodal}
Jacob Arkin, Daehyung Park, Subhro Roy, Matthew~R Walter, Nicholas Roy,
  Thomas~M Howard, and Rohan Paul.
\newblock Multimodal estimation and communication of latent semantic knowledge
  for robust execution of robot instructions.
\newblock \emph{The International Journal of Robotics Research}, 39\penalty0
  (10-11):\penalty0 1279--1304, 2020{\natexlab{a}}.
\newblock URL
  \url{https://journals.sagepub.com/doi/abs/10.1177/0278364920917755}.

\bibitem[Arkin et~al.(2020{\natexlab{b}})Arkin, Paul, Park, Roy, Roy, and
  Howard]{arkin2020real}
Jacob Arkin, Rohan Paul, Daehyung Park, Subhro Roy, Nicholas Roy, and Thomas~M
  Howard.
\newblock Real-time human-robot communication for manipulation tasks in
  partially observed environments.
\newblock In \emph{Proceedings of the International Symposium on Experimental
  Robotics}, pp.\  448--460. Springer, 2020{\natexlab{b}}.
\newblock URL
  \url{https://link.springer.com/chapter/10.1007/978-3-030-33950-0_39}.

\bibitem[Armeni et~al.(2016)Armeni, Sener, Zamir, Jiang, Brilakis, Fischer, and
  Savarese]{armeni20163d}
Iro Armeni, Ozan Sener, Amir~R Zamir, Helen Jiang, Ioannis Brilakis, Martin
  Fischer, and Silvio Savarese.
\newblock 3d semantic parsing of large-scale indoor spaces.
\newblock In \emph{Proceedings of the IEEE conference on computer vision and
  pattern recognition}, pp.\  1534--1543, 2016.
\newblock URL
  \url{http://openaccess.thecvf.com/content_cvpr_2016/html/Armeni_3D_Semantic_Parsing_CVPR_2016_paper.html}.

\bibitem[Armeni et~al.(2019)Armeni, He, Gwak, Zamir, Fischer, Malik, and
  Savarese]{armeni20193d}
Iro Armeni, Zhi-Yang He, JunYoung Gwak, Amir~R Zamir, Martin Fischer, Jitendra
  Malik, and Silvio Savarese.
\newblock {3D} scene graph: {A} structure for unified semantics, {3D} space,
  and camera.
\newblock In \emph{Proceedings of the IEEE/CVF International Conference on
  Computer Vision}, pp.\  5664--5673, 2019.
\newblock URL
  \url{http://openaccess.thecvf.com/content_ICCV_2019/html/Armeni_3D_Scene_Graph_A_Structure_for_Unified_Semantics_3D_Space_ICCV_2019_paper.html}.

\bibitem[Atanasov et~al.(2018)Atanasov, Bowman, Daniilidis, and
  Pappas]{atanasov2018unifying}
Nikolay Atanasov, Sean~L Bowman, Kostas Daniilidis, and George~J Pappas.
\newblock {A Unifying View of Geometry, Semantics, and Data Association in
  SLAM}.
\newblock In \emph{Proceedings of the International Joint Conference on
  Artificial Intelligence}, pp.\  5204--5208, 2018.
\newblock URL \url{http://erl.ucsd.edu/ref/Atanasov_SemanticSLAM_IJCAI18.pdf}.

\bibitem[Aydemir et~al.(2011)Aydemir, Sj{\"o}{\"o}, Folkesson, Pronobis, and
  Jensfelt]{aydemir2011search}
Alper Aydemir, Kristoffer Sj{\"o}{\"o}, John Folkesson, Andrzej Pronobis, and
  Patric Jensfelt.
\newblock Search in the real world: {A}ctive visual object search based on
  spatial relations.
\newblock In \emph{IEEE International Conference on Robotics and Automation},
  pp.\  2818--2824. IEEE, 2011.
\newblock URL \url{https://ieeexplore.ieee.org/abstract/document/5980495}.

\bibitem[Aydemir et~al.(2013)Aydemir, Pronobis, G{\"o}belbecker, and
  Jensfelt]{aydemir2013active}
Alper Aydemir, Andrzej Pronobis, Moritz G{\"o}belbecker, and Patric Jensfelt.
\newblock Active visual object search in unknown environments using uncertain
  semantics.
\newblock \emph{IEEE Transactions on Robotics}, 29\penalty0 (4):\penalty0
  986--1002, 2013.
\newblock URL \url{https://ieeexplore.ieee.org/abstract/document/6507635}.

\bibitem[Bansal et~al.(2020)Bansal, Tolani, Gupta, Malik, and
  Tomlin]{bansal2020combining}
Somil Bansal, Varun Tolani, Saurabh Gupta, Jitendra Malik, and Claire Tomlin.
\newblock Combining optimal control and learning for visual navigation in novel
  environments.
\newblock In \emph{Conference on Robot Learning}, pp.\  420--429. PMLR, 2020.
\newblock URL \url{http://proceedings.mlr.press/v100/bansal20a}.

\bibitem[Bao \& Savarese(2011)Bao and Savarese]{bao2011semantic}
Sid~Yingze Bao and Silvio Savarese.
\newblock Semantic structure from motion.
\newblock In \emph{Proceedings of the IEEE/CVF Conference on Computer Vision
  and Pattern Recognition}, pp.\  2025--2032. IEEE, 2011.
\newblock URL \url{https://ieeexplore.ieee.org/abstract/document/5995462/}.

\bibitem[Bao et~al.(2012)Bao, Bagra, Chao, and Savarese]{bao2012semantic}
Sid~Yingze Bao, Mohit Bagra, Yu-Wei Chao, and Silvio Savarese.
\newblock Semantic structure from motion with points, regions, and objects.
\newblock In \emph{Proceedings of the IEEE/CVF Conference on Computer Vision
  and Pattern Recognition}, pp.\  2703--2710. IEEE, 2012.
\newblock URL \url{https://ieeexplore.ieee.org/abstract/document/6247992/}.

\bibitem[Bao et~al.(2023)Bao, Hossain, Lang, and Lin]{bao2023review}
Zhibin Bao, Sabir Hossain, Haoxiang Lang, and Xianke Lin.
\newblock A review of high-definition map creation methods for autonomous
  driving.
\newblock \emph{Engineering Applications of Artificial Intelligence},
  122:\penalty0 106125, 2023.
\newblock URL
  \url{https://www.sciencedirect.com/science/article/pii/S0952197623003093}.

\bibitem[Batra et~al.(2020{\natexlab{a}})Batra, Chang, Chernova, Davison, Deng,
  Koltun, Levine, Malik, Mordatch, Mottaghi, et~al.]{batra2020rearrangement}
Dhruv Batra, Angel~X Chang, Sonia Chernova, Andrew~J Davison, Jia Deng, Vladlen
  Koltun, Sergey Levine, Jitendra Malik, Igor Mordatch, Roozbeh Mottaghi,
  et~al.
\newblock Rearrangement: A challenge for embodied ai.
\newblock \emph{arXiv preprint arXiv:2011.01975}, 2020{\natexlab{a}}.
\newblock URL \url{https://arxiv.org/abs/2011.01975}.

\bibitem[Batra et~al.(2020{\natexlab{b}})Batra, Gokaslan, Kembhavi, Maksymets,
  Mottaghi, Savva, Toshev, and Wijmans]{batra2020objectnav}
Dhruv Batra, Aaron Gokaslan, Aniruddha Kembhavi, Oleksandr Maksymets, Roozbeh
  Mottaghi, Manolis Savva, Alexander Toshev, and Erik Wijmans.
\newblock {ObjectNav} revisited: On evaluation of embodied agents navigating to
  objects.
\newblock \emph{arXiv preprint arXiv:2006.13171}, 2020{\natexlab{b}}.
\newblock URL \url{https://arxiv.org/abs/2006.13171}.

\bibitem[Bavle et~al.(2023)Bavle, Sanchez-Lopez, Shaheer, Civera, and
  Voos]{bavle2023s}
Hriday Bavle, Jose~Luis Sanchez-Lopez, Muhammad Shaheer, Javier Civera, and
  Holger Voos.
\newblock S-graphs+: {R}eal-time localization and mapping leveraging
  hierarchical representations.
\newblock \emph{IEEE Robotics and Automation Letters}, 8\penalty0 (8):\penalty0
  4927--4934, 2023.
\newblock URL \url{https://orbilu.uni.lu/handle/10993/54383}.

\bibitem[Bay et~al.(2008)Bay, Ess, Tuytelaars, and Van~Gool]{bay2008speeded}
Herbert Bay, Andreas Ess, Tinne Tuytelaars, and Luc Van~Gool.
\newblock Speeded-up robust features (surf).
\newblock \emph{Computer vision and image understanding}, 110\penalty0
  (3):\penalty0 346--359, 2008.
\newblock URL
  \url{https://www.sciencedirect.com/science/article/pii/S1077314207001555}.

\bibitem[Beeching et~al.(2020)Beeching, Dibangoye, Simonin, and
  Wolf]{Beeching2020LearningTP}
Edward Beeching, Jilles Dibangoye, Olivier Simonin, and Christian Wolf.
\newblock {Learning to plan with uncertain topological maps}.
\newblock In \emph{European Conference on Computer Vision}, pp.\  473--490.
  Springer, 2020.
\newblock URL \url{https://api.semanticscholar.org/CorpusID:220486391}.

\bibitem[Behley et~al.(2012)Behley, Steinhage, and
  Cremers]{behley2012performance}
Jens Behley, Volker Steinhage, and Armin~B Cremers.
\newblock {Performance of histogram descriptors for the classification of 3D
  laser range data in urban environments}.
\newblock In \emph{IEEE International Conference on Robotics and Automation},
  pp.\  4391--4398. IEEE, 2012.
\newblock URL \url{https://ieeexplore.ieee.org/abstract/document/6225003/}.

\bibitem[Blanco et~al.(2008{\natexlab{a}})Blanco, Fern{\'a}ndez-Madrigal, and
  Gonz{\'a}lez]{Blanco2008ANM}
Jos{\'e}-Luis Blanco, Juan-Antonio Fern{\'a}ndez-Madrigal, and Javier
  Gonz{\'a}lez.
\newblock {A Novel Measure of Uncertainty for Mobile Robot SLAM with
  Rao—Blackwellized Particle Filters}.
\newblock \emph{The International Journal of Robotics Research}, 27:\penalty0
  73 -- 89, 2008{\natexlab{a}}.
\newblock URL \url{https://api.semanticscholar.org/CorpusID:5143649}.

\bibitem[Blanco et~al.(2008{\natexlab{b}})Blanco, Fern{\'a}ndez-Madrigal, and
  Gonzalez]{blanco2008toward}
Jose-Luis Blanco, Juan-Antonio Fern{\'a}ndez-Madrigal, and Javier Gonzalez.
\newblock Toward a unified bayesian approach to hybrid metric--topological
  slam.
\newblock \emph{IEEE Transactions on Robotics}, 24\penalty0 (2):\penalty0
  259--270, 2008{\natexlab{b}}.
\newblock URL \url{https://ieeexplore.ieee.org/abstract/document/4472721/}.

\bibitem[Blochliger et~al.(2018)Blochliger, Fehr, Dymczyk, Schneider, and
  Siegwart]{blochliger2018topomap}
Fabian Blochliger, Marius Fehr, Marcin Dymczyk, Thomas Schneider, and Rol
  Siegwart.
\newblock Topomap: {T}opological mapping and navigation based on visual slam
  maps.
\newblock In \emph{IEEE International Conference on Robotics and Automation},
  pp.\  3818--3825. IEEE, 2018.
\newblock URL \url{https://ieeexplore.ieee.org/abstract/document/8460641/}.

\bibitem[Bowman et~al.(2017)Bowman, Atanasov, Daniilidis, and
  Pappas]{bowman2017probabilistic}
Sean~L Bowman, Nikolay Atanasov, Kostas Daniilidis, and George~J Pappas.
\newblock Probabilistic data association for semantic {SLAM}.
\newblock In \emph{IEEE International Conference on Robotics and Automation},
  pp.\  1722--1729. IEEE, 2017.
\newblock URL \url{https://ieeexplore.ieee.org/abstract/document/7989203/}.

\bibitem[Busch et~al.(2025)Busch, Homberger, Ortega-Peimbert, Yang, and
  Andersson]{busch2024mapallrealtimeopenvocabulary}
Finn~L Busch, Timon Homberger, Jes{\'u}s Ortega-Peimbert, Quantao Yang, and
  Olov Andersson.
\newblock {One Map to Find Them All: Real-time Open-Vocabulary Mapping for
  Zero-shot Multi-Object Navigation}.
\newblock In \emph{IEEE International Conference on Robotics and Automation},
  2025.
\newblock URL \url{https://arxiv.org/abs/2409.11764}.

\bibitem[Cadena et~al.(2015)Cadena, Dick, and Reid]{cadena2015fast}
Cesar Cadena, Anthony Dick, and Ian~D Reid.
\newblock A fast, modular scene understanding system using context-aware object
  detection.
\newblock In \emph{IEEE International Conference on Robotics and Automation},
  pp.\  4859--4866. IEEE, 2015.
\newblock URL \url{https://ieeexplore.ieee.org/abstract/document/7139874/}.

\bibitem[Cadena et~al.(2016)Cadena, Carlone, Carrillo, Latif, Scaramuzza,
  Neira, Reid, and Leonard]{cadena2016simultaneous}
Cesar Cadena, Luca Carlone, Henry Carrillo, Yasir Latif, Davide Scaramuzza,
  Jos{\'e} Neira, Ian~D Reid, and John~J Leonard.
\newblock {Simultaneous localization and mapping: Present, future, and the
  robust-perception age}.
\newblock \emph{IEEE Transactions on Robotics}, 32, 2016.
\newblock URL \url{https://core.ac.uk/download/pdf/130797919.pdf}.

\bibitem[Cadena et~al.(2017)Cadena, Carlone, Carrillo, Latif, Scaramuzza,
  Neira, Reid, and Leonard]{cadena2017past}
Cesar Cadena, Luca Carlone, Henry Carrillo, Yasir Latif, Davide Scaramuzza,
  Jos{\'e} Neira, Ian Reid, and John~J Leonard.
\newblock Past, present, and future of simultaneous localization and mapping:
  {T}oward the robust-perception age.
\newblock \emph{IEEE Transactions on robotics}, 32\penalty0 (6):\penalty0
  1309--1332, 2017.
\newblock URL \url{https://core.ac.uk/download/pdf/130797919.pdf}.

\bibitem[Cai et~al.(2021)Cai, Ye, Gao, Li, and Zhang]{cai2021improved}
Lecai Cai, Yuling Ye, Xiang Gao, Zhong Li, and Chaoyang Zhang.
\newblock An improved visual {SLAM} based on affine transformation for {ORB}
  feature extraction.
\newblock \emph{Optik}, 227:\penalty0 165421, 2021.
\newblock URL
  \url{https://www.sciencedirect.com/science/article/pii/S0030402620312572}.

\bibitem[Campos et~al.(2021)Campos, Elvira, Rodr{\'\i}guez, Montiel, and
  Tard{\'o}s]{campos2021orb}
Carlos Campos, Richard Elvira, Juan J~G{\'o}mez Rodr{\'\i}guez, Jos{\'e}~MM
  Montiel, and Juan~D Tard{\'o}s.
\newblock {ORB-SLAM3}: An accurate open-source library for visual,
  visual--inertial, and multimap {SLAM}.
\newblock \emph{IEEE Transactions on Robotics}, 37\penalty0 (6):\penalty0
  1874--1890, 2021.
\newblock URL \url{https://ieeexplore.ieee.org/abstract/document/9440682/}.

\bibitem[Camps et~al.(2022)Camps, Dyro, Pavone, and
  Schwager]{camps2022learning}
Gadiel~Sznaier Camps, Robert Dyro, Marco Pavone, and Mac Schwager.
\newblock {Learning Deep SDF Maps Online for Robot Navigation and Exploration}.
\newblock \emph{arXiv preprint arXiv:2207.10782}, 2022.
\newblock URL \url{https://arxiv.org/abs/2207.10782}.

\bibitem[Carlone et~al.(2013)Carlone, Du, Ng, Bona, and
  Indri]{Carlone2013ActiveSA}
Luca Carlone, Jingjing Du, Miguel Efrain~Kaouk Ng, Basilio Bona, and Marina
  Indri.
\newblock {Active SLAM and Exploration with Particle Filters Using
  Kullback-Leibler Divergence}.
\newblock \emph{Journal of Intelligent \& Robotic Systems}, 75:\penalty0 291 --
  311, 2013.
\newblock URL \url{https://api.semanticscholar.org/CorpusID:207173907}.

\bibitem[Carlone et~al.(2024)Carlone, Kim, Dellaert, Barfoot, and
  Cremers]{sh-p1-prelude}
Luca Carlone, Ayoung Kim, Frank Dellaert, Timothy Barfoot, and Daniel Cremers.
\newblock \emph{SLAM Handbook}, chapter~1.
\newblock Cambridge University Press, 2024.
\newblock URL
  \url{https://github.com/SLAM-Handbook-contributors/slam-handbook-public-release}.

\bibitem[Caron et~al.(2021)Caron, Touvron, Misra, J{\'e}gou, Mairal,
  Bojanowski, and Joulin]{caron2021emerging}
Mathilde Caron, Hugo Touvron, Ishan Misra, Herv{\'e} J{\'e}gou, Julien Mairal,
  Piotr Bojanowski, and Armand Joulin.
\newblock Emerging properties in self-supervised vision transformers.
\newblock In \emph{Proceedings of the IEEE/CVF international conference on
  computer vision}, pp.\  9650--9660, 2021.
\newblock URL
  \url{https://openaccess.thecvf.com/content/ICCV2021/html/Caron_Emerging_Properties_in_Self-Supervised_Vision_Transformers_ICCV_2021_paper}.

\bibitem[Cartillier et~al.(2021)Cartillier, Ren, Jain, Lee, Essa, and
  Batra]{cartillier2020semantic}
Vincent Cartillier, Zhile Ren, Neha Jain, Stefan Lee, Irfan Essa, and Dhruv
  Batra.
\newblock Semantic {MapNet}: Building allocentric semantic maps and
  representations from egocentric views.
\newblock In \emph{Proceedings of the AAAI Conference on Artificial
  Intelligence}, 2021.
\newblock URL \url{https://ojs.aaai.org/index.php/AAAI/article/view/16180}.

\bibitem[Center(1984)]{center1984shakey}
Artificial~Intellgence Center.
\newblock Shakey the robot.
\newblock \emph{wikipedia}, 1984.
\newblock URL \url{https://en.wikipedia.org/wiki/Shakey_the_robot}.

\bibitem[Chan et~al.(2012)Chan, Baumann, Bellgrove, and
  Mattingley]{object_important}
Edgar Chan, Oliver Baumann, Mark Bellgrove, and Jason Mattingley.
\newblock From objects to landmarks: The function of visual location
  information in spatial navigation.
\newblock \emph{Frontiers in Psychology}, 3:\penalty0 304, 08 2012.
\newblock \doi{10.3389/fpsyg.2012.00304}.
\newblock URL
  \url{https://www.frontiersin.org/articles/10.3389/fpsyg.2012.00304/full}.

\bibitem[Chang et~al.(2017)Chang, Dai, Funkhouser, Halber, Niebner, Savva,
  Song, Zeng, and Zhang]{chang2017matterport3d}
Angel Chang, Angela Dai, Thomas Funkhouser, Maciej Halber, Matthias Niebner,
  Manolis Savva, Shuran Song, Andy Zeng, and Yinda Zhang.
\newblock Matterport3{D}: Learning from {RGB-D} data in indoor environments.
\newblock In \emph{International Conference on 3D Vision}, 2017.
\newblock URL
  \url{https://collaborate.princeton.edu/en/publications/matterport3d-learning-from-rgb-d-data-in-indoor-environments}.

\bibitem[Chang et~al.(2023)Chang, Gervet, Khanna, Yenamandra, Shah, Min, Shah,
  Paxton, Gupta, Batra, et~al.]{chang2023goat}
Matthew Chang, Theophile Gervet, Mukul Khanna, Sriram Yenamandra, Dhruv Shah,
  So~Yeon Min, Kavit Shah, Chris Paxton, Saurabh Gupta, Dhruv Batra, et~al.
\newblock {GOAT}: Go to any thing.
\newblock \emph{arXiv preprint arXiv:2311.06430}, 2023.
\newblock URL \url{https://arxiv.org/abs/2311.06430}.

\bibitem[Chang et~al.(2025)Chang, Fermoselle, Ta, Bucher, Carlone, and
  Wang]{chang2025ashitaautomaticscenegroundedhierarchical}
Yun Chang, Leonor Fermoselle, Duy Ta, Bernadette Bucher, Luca Carlone, and
  Jiuguang Wang.
\newblock {ASHiTA: Automatic Scene-grounded HIerarchical Task Analysis}.
\newblock In \emph{Proceedings of the Computer Vision and Pattern Recognition
  Conference}, pp.\  29458--29468, 2025.
\newblock URL
  \url{https://openaccess.thecvf.com/content/CVPR2025/html/Chang_ASHiTA_Automatic_Scene-grounded_HIerarchical_Task_Analysis_CVPR_2025_paper.html}.

\bibitem[Chaplot et~al.(2019)Chaplot, Gandhi, Gupta, Gupta, and
  Salakhutdinov]{chaplot2020learning}
Devendra~Singh Chaplot, Dhiraj Gandhi, Saurabh Gupta, Abhinav Gupta, and Ruslan
  Salakhutdinov.
\newblock Learning to explore using active neural {SLAM}.
\newblock In \emph{International Conference on Learning Representations}, 2019.
\newblock URL \url{https://openreview.net/forum?id=HklXn1BKDH}.

\bibitem[Chaplot et~al.(2020{\natexlab{a}})Chaplot, Gandhi, Gupta, and
  Salakhutdinov]{chaplot2020object}
Devendra~Singh Chaplot, Dhiraj~Prakashchand Gandhi, Abhinav Gupta, and Russ~R
  Salakhutdinov.
\newblock Object goal navigation using goal-oriented semantic exploration.
\newblock In \emph{Advances in Neural Information Processing Systems},
  volume~33, pp.\  4247--4258, 2020{\natexlab{a}}.
\newblock URL
  \url{https://proceedings.neurips.cc/paper/2020/hash/2c75cf2681788adaca63aa95ae028b22-Abstract.html}.

\bibitem[Chaplot et~al.(2020{\natexlab{b}})Chaplot, Salakhutdinov, Gupta, and
  Gupta]{chaplot2020neural}
Devendra~Singh Chaplot, Ruslan Salakhutdinov, Abhinav Gupta, and Saurabh Gupta.
\newblock {Neural Topological SLAM for Visual Navigation}.
\newblock In \emph{Proceedings of the IEEE/CVF Conference on Computer Vision
  and Pattern Recognition}, 2020{\natexlab{b}}.
\newblock URL
  \url{http://openaccess.thecvf.com/content\_CVPR\_2020/html/Chaplot\_Neural\_Topological\_SLAM\_for\_Visual\_Navigation\_CVPR\_2020\_paper.html}.

\bibitem[Chaplot et~al.(2021)Chaplot, Dalal, Gupta, Malik, and
  Salakhutdinov]{chaplot2021seal}
Devendra~Singh Chaplot, Murtaza Dalal, Saurabh Gupta, Jitendra Malik, and
  Russ~R Salakhutdinov.
\newblock {SEAL}: Self-supervised embodied active learning using exploration
  and {3D} consistency.
\newblock \emph{Advances in Neural Information Processing Systems},
  34:\penalty0 13086--13098, 2021.
\newblock URL
  \url{https://proceedings.neurips.cc/paper/2021/hash/6d0c932802f6953f70eb20931645fa40-Abstract.html}.

\bibitem[Chen et~al.(2023{\natexlab{a}})Chen, Xia, Ichter, Rao, Gopalakrishnan,
  Ryoo, Stone, and Kappler]{chen2023open}
Boyuan Chen, Fei Xia, Brian Ichter, Kanishka Rao, Keerthana Gopalakrishnan,
  Michael~S Ryoo, Austin Stone, and Daniel Kappler.
\newblock Open-vocabulary queryable scene representations for real world
  planning.
\newblock In \emph{IEEE International Conference on Robotics and Automation},
  pp.\  11509--11522. IEEE, 2023{\natexlab{a}}.
\newblock URL \url{https://ieeexplore.ieee.org/document/10161534}.

\bibitem[Chen et~al.(2020{\natexlab{a}})Chen, Jain, Schissler, Gari, Al-Halah,
  Ithapu, Robinson, and Grauman]{chen2020soundspaces}
Changan Chen, Unnat Jain, Carl Schissler, Sebastia Vicenc~Amengual Gari, Ziad
  Al-Halah, Vamsi~Krishna Ithapu, Philip Robinson, and Kristen Grauman.
\newblock Sound{S}paces: Audio-visual navigation in {3D} environments.
\newblock In \emph{European Conference on Computer Vision}, pp.\  17--36.
  Springer, 2020{\natexlab{a}}.
\newblock URL
  \url{https://link.springer.com/chapter/10.1007/978-3-030-58539-6\_2}.

\bibitem[Chen et~al.(2020{\natexlab{b}})Chen, Majumder, Al-Halah, Gao,
  Ramakrishnan, and Grauman]{chen2020learning}
Changan Chen, Sagnik Majumder, Ziad Al-Halah, Ruohan Gao, Santhosh~Kumar
  Ramakrishnan, and Kristen Grauman.
\newblock Learning to set waypoints for audio-visual navigation.
\newblock In \emph{International Conference on Learning Representations},
  2020{\natexlab{b}}.
\newblock URL \url{https://openreview.net/forum?id=cR91FAodFMe}.

\bibitem[Chen et~al.(2020{\natexlab{c}})Chen, Wang, Lu, Trigoni, and
  Markham]{chen2020survey}
Changhao Chen, Bing Wang, Chris~Xiaoxuan Lu, Niki Trigoni, and Andrew Markham.
\newblock A survey on deep learning for localization and mapping: Towards the
  age of spatial machine intelligence.
\newblock \emph{arXiv preprint arXiv:2006.12567}, 2020{\natexlab{c}}.
\newblock URL \url{https://arxiv.org/abs/2006.12567}.

\bibitem[Chen \& Mooney(2011)Chen and Mooney]{Chen_Mooney_2011}
David Chen and Raymond Mooney.
\newblock {Learning to Interpret Natural Language Navigation Instructions from
  Observations}.
\newblock \emph{Proceedings of the AAAI Conference on Artificial Intelligence},
  25\penalty0 (1):\penalty0 859--865, Aug. 2011.
\newblock \doi{10.1609/aaai.v25i1.7974}.
\newblock URL \url{https://ojs.aaai.org/index.php/AAAI/article/view/7974}.

\bibitem[Chen et~al.(2023{\natexlab{b}})Chen, Li, Kumar, Ghanem, and
  Yu]{chen2023structnav}
Junting Chen, Guohao Li, Suryansh Kumar, Bernard Ghanem, and Fisher Yu.
\newblock {How To Not Train Your Dragon: Training-free Embodied Object
  Navigation with Semantic Frontiers}.
\newblock In \emph{Proceedings of Robotics: Science and System},
  2023{\natexlab{b}}.
\newblock URL
  \url{https://www.research-collection.ethz.ch/handle/20.500.11850/622379}.

\bibitem[Chen et~al.(2025)Chen, Xiao, Liu, Tong, Zhang, Liu, Zhang, Ajoudani,
  and Chen]{chen2025semantic}
Kaiqi Chen, Junhao Xiao, Jialing Liu, Qiyi Tong, Heng Zhang, Ruyu Liu, Jianhua
  Zhang, Arash Ajoudani, and Shengyong Chen.
\newblock Semantic visual simultaneous localization and mapping: {A} survey.
\newblock \emph{IEEE Transactions on Intelligent Transportation Systems}, 2025.
\newblock URL \url{https://ieeexplore.ieee.org/abstract/document/11005698}.

\bibitem[Chen et~al.(2021)Chen, Chen, Chuang, V{\'a}zquez, and
  Savarese]{chen2021topological}
Kevin Chen, Junshen~K Chen, Jo~Chuang, Marynel V{\'a}zquez, and Silvio
  Savarese.
\newblock Topological planning with transformers for vision-and-language
  navigation.
\newblock In \emph{Proceedings of the IEEE/CVF Conference on Computer Vision
  and Pattern Recognition}, pp.\  11276--11286, 2021.
\newblock URL
  \url{http://openaccess.thecvf.com/content/CVPR2021/html/Chen_Topological_Planning_With_Transformers_for_Vision-and-Language_Navigation_CVPR_2021_paper.html}.

\bibitem[Chen et~al.(2023{\natexlab{c}})Chen, Liu, Kong, Zhu, Ma, Li, Hou,
  Qiao, and Wang]{chen2023clip2scene}
Runnan Chen, Youquan Liu, Lingdong Kong, Xinge Zhu, Yuexin Ma, Yikang Li,
  Yuenan Hou, Yu~Qiao, and Wenping Wang.
\newblock {CLIP2Scene}: Towards label-efficient {3D} scene understanding by
  {CLIP}.
\newblock In \emph{Proceedings of the IEEE/CVF Conference on Computer Vision
  and Pattern Recognition}, pp.\  7020--7030, 2023{\natexlab{c}}.
\newblock URL
  \url{http://openaccess.thecvf.com/content/CVPR2023/html/Chen_CLIP2Scene_Towards_Label-Efficient_3D_Scene_Understanding_by_CLIP_CVPR_2023_paper.html}.

\bibitem[Chen et~al.(2020{\natexlab{d}})Chen, Nardari, Lee, Qu, Liu, Romero,
  and Kumar]{chen2020sloam}
Steven~W Chen, Guilherme~V Nardari, Elijah~S Lee, Chao Qu, Xu~Liu, Roseli
  Ap~Francelin Romero, and Vijay Kumar.
\newblock Sloam: {S}emantic lidar odometry and mapping for forest inventory.
\newblock \emph{IEEE Robotics and Automation Letters}, 5\penalty0 (2):\penalty0
  612--619, 2020{\natexlab{d}}.
\newblock URL \url{https://ieeexplore.ieee.org/abstract/document/8949363/}.

\bibitem[Chen et~al.(2019{\natexlab{a}})Chen, Gupta, and
  Gupta]{chen2019learning}
Tao Chen, Saurabh Gupta, and Abhinav Gupta.
\newblock Learning exploration policies for navigation.
\newblock In \emph{International Conference on Learning Representations},
  2019{\natexlab{a}}.
\newblock URL \url{https://openreview.net/forum?id=SyMWn05F7}.

\bibitem[Chen et~al.(2019{\natexlab{b}})Chen, Milioto, Palazzolo, Giguere,
  Behley, and Stachniss]{chen2019suma++}
Xieyuanli Chen, Andres Milioto, Emanuele Palazzolo, Philippe Giguere, Jens
  Behley, and Cyrill Stachniss.
\newblock Suma++: {E}fficient lidar-based semantic slam.
\newblock In \emph{IEEE/RSJ International Conference on Intelligent Robots and
  Systems (IROS)}, pp.\  4530--4537. IEEE, 2019{\natexlab{b}}.
\newblock URL \url{https://ieeexplore.ieee.org/abstract/document/8967704}.

\bibitem[Chen et~al.(2022)Chen, Zhou, Lin, Zhang, Zhang, and
  Shen]{chen2022fast}
Xinyi Chen, Boyu Zhou, Jiarong Lin, Yichen Zhang, Fu~Zhang, and Shaojie Shen.
\newblock Fast 3{D} sparse topological skeleton graph generation for mobile
  robot global planning.
\newblock In \emph{IEEE/RSJ International Conference on Intelligent Robots and
  Systems}, pp.\  10283--10289. IEEE, 2022.
\newblock URL \url{https://ieeexplore.ieee.org/abstract/document/9981397/}.

\bibitem[Cheng et~al.(2021)Cheng, Zhang, and Chen]{cheng2021improved}
Jun Cheng, Liyan Zhang, and Qihong Chen.
\newblock An improved initialization method for monocular visual-inertial
  {SLAM}.
\newblock \emph{Electronics}, 10\penalty0 (24):\penalty0 3063, 2021.
\newblock URL \url{https://www.mdpi.com/2079-9292/10/24/3063}.

\bibitem[Choset \& Nagatani(2001)Choset and Nagatani]{choset2001topological}
Howie Choset and Keiji Nagatani.
\newblock Topological simultaneous localization and mapping ({SLAM}): toward
  exact localization without explicit localization.
\newblock \emph{IEEE Transactions on robotics and automation}, 17\penalty0
  (2):\penalty0 125--137, 2001.
\newblock URL \url{https://ieeexplore.ieee.org/abstract/document/928558/}.

\bibitem[Civera et~al.(2011)Civera, G{\'a}lvez-L{\'o}pez, Riazuelo, Tard{\'o}s,
  and Montiel]{civera2011towards}
Javier Civera, Dorian G{\'a}lvez-L{\'o}pez, Luis Riazuelo, Juan~D Tard{\'o}s,
  and Jose Maria~Martinez Montiel.
\newblock Towards semantic {SLAM} using a monocular camera.
\newblock In \emph{IEEE/RSJ international conference on intelligent robots and
  systems}, pp.\  1277--1284. IEEE, 2011.
\newblock URL \url{https://ieeexplore.ieee.org/abstract/document/6094648/}.

\bibitem[Dai et~al.(2017)Dai, Nie{\ss}ner, Zollh{\"o}fer, Izadi, and
  Theobalt]{dai2017bundlefusion}
Angela Dai, Matthias Nie{\ss}ner, Michael Zollh{\"o}fer, Shahram Izadi, and
  Christian Theobalt.
\newblock {Bundlefusion: Real-time globally consistent 3d reconstruction using
  on-the-fly surface reintegration}.
\newblock \emph{ACM Transactions on Graphics (ToG)}, 36\penalty0 (4):\penalty0
  1, 2017.
\newblock URL \url{https://dl.acm.org/doi/abs/10.1145/3072959.3054739}.

\bibitem[Deitke et~al.(2022)Deitke, Batra, Bisk, Campari, Chang, Chaplot, Chen,
  D'Arpino, Ehsani, Farhadi, et~al.]{deitke2022retrospectives}
Matt Deitke, Dhruv Batra, Yonatan Bisk, Tommaso Campari, Angel~X Chang,
  Devendra~Singh Chaplot, Changan Chen, Claudia~P{\'e}rez D'Arpino, Kiana
  Ehsani, Ali Farhadi, et~al.
\newblock Retrospectives on the embodied {AI} workshop.
\newblock \emph{arXiv preprint arXiv:2210.06849}, 2022.
\newblock URL \url{https://arxiv.org/abs/2210.06849}.

\bibitem[Dellaert et~al.(2000)Dellaert, Seitz, Thrun, and
  Thorpe]{dellaert2000feature}
Frank Dellaert, Steven Seitz, Sebastian Thrun, and Charles Thorpe.
\newblock Feature correspondence: {A} markov chain monte carlo approach.
\newblock \emph{Advances in Neural Information Processing Systems}, 13, 2000.
\newblock URL
  \url{https://proceedings.neurips.cc/paper/2000/hash/d7657583058394c828ee150fada65345-Abstract.html}.

\bibitem[Dellaert et~al.(2017)Dellaert, Kaess, et~al.]{dellaert2017factor}
Frank Dellaert, Michael Kaess, et~al.
\newblock Factor graphs for robot perception.
\newblock \emph{Foundations and Trends{\textregistered} in Robotics},
  6\penalty0 (1-2):\penalty0 1--139, 2017.
\newblock URL \url{https://www.nowpublishers.com/article/Details/ROB-043}.

\bibitem[Deng et~al.(2009)Deng, Dong, Socher, Li, Li, and
  Fei-Fei]{deng2009imagenet}
Jia Deng, Wei Dong, Richard Socher, Li-Jia Li, Kai Li, and Li~Fei-Fei.
\newblock Image{N}et: A large-scale hierarchical image database.
\newblock In \emph{Proceedings of the IEEE/CVF Conference on Computer Vision
  and Pattern Recognition}, pp.\  248--255, 2009.
\newblock URL \url{https://ieeexplore.ieee.org/abstract/document/5206848/}.

\bibitem[Desai et~al.(2022)Desai, Parikh, Kumari, and
  Raman]{desai2022pointresnet}
Aadesh Desai, Saagar Parikh, Seema Kumari, and Shanmuganathan Raman.
\newblock {PointResNet: residual network for 3D point cloud segmentation and
  classification}.
\newblock \emph{arXiv preprint arXiv:2211.11040}, 2022.
\newblock URL \url{https://arxiv.org/abs/2211.11040}.

\bibitem[Detesan \& Moholea(2024)Detesan and Moholea]{detesan2024unimate}
Ovidiu-Aurelian Detesan and Iuliana~Fabiola Moholea.
\newblock {Unimate and Beyond: Exploring the Genesis of Industrial Robotics}.
\newblock In \emph{International Conference on Robotics in Alpe-Adria Danube
  Region}, pp.\  263--271. Springer, 2024.
\newblock URL
  \url{https://link.springer.com/chapter/10.1007/978-3-031-59257-7_27}.

\bibitem[Dobrevski \& Sko{\v{c}}aj(2021)Dobrevski and
  Sko{\v{c}}aj]{dobrevski2021deep}
Matej Dobrevski and Danijel Sko{\v{c}}aj.
\newblock Deep reinforcement learning for map-less goal-driven robot
  navigation.
\newblock \emph{International Journal of Advanced Robotic Systems}, 18\penalty0
  (1):\penalty0 1729881421992621, 2021.
\newblock URL
  \url{https://journals.sagepub.com/doi/abs/10.1177/1729881421992621}.

\bibitem[Dosovitskiy et~al.(2020)Dosovitskiy, Beyer, Kolesnikov, Weissenborn,
  Zhai, Unterthiner, Dehghani, Minderer, Heigold, Gelly,
  et~al.]{dosovitskiy2020image}
Alexey Dosovitskiy, Lucas Beyer, Alexander Kolesnikov, Dirk Weissenborn,
  Xiaohua Zhai, Thomas Unterthiner, Mostafa Dehghani, Matthias Minderer, Georg
  Heigold, Sylvain Gelly, et~al.
\newblock An image is worth 16x16 words: Transformers for image recognition at
  scale.
\newblock \emph{arXiv preprint arXiv:2010.11929}, 2020.
\newblock URL \url{https://arxiv.org/pdf/2010.11929/1000}.

\bibitem[Driess et~al.(2023)Driess, Xia, Sajjadi, Lynch, Chowdhery, Ichter,
  Wahid, Tompson, Vuong, Yu, Huang, Chebotar, Sermanet, Duckworth, Levine,
  Vanhoucke, Hausman, Toussaint, Greff, Zeng, Mordatch, and
  Florence]{Driess2023PaLMEAE}
Danny Driess, F.~Xia, Mehdi S.~M. Sajjadi, Corey Lynch, Aakanksha Chowdhery,
  Brian Ichter, Ayzaan Wahid, Jonathan Tompson, Quan~Ho Vuong, Tianhe Yu,
  Wenlong Huang, Yevgen Chebotar, Pierre Sermanet, Daniel Duckworth, Sergey
  Levine, Vincent Vanhoucke, Karol Hausman, Marc Toussaint, Klaus Greff, Andy
  Zeng, Igor Mordatch, and Peter~R. Florence.
\newblock {PaLM-E: An Embodied Multimodal Language Model}.
\newblock In \emph{International Conference on Machine Learning}, 2023.
\newblock URL \url{https://api.semanticscholar.org/CorpusID:257364842}.

\bibitem[Duan et~al.(2022)Duan, Yu, Tan, Zhu, and Tan]{duan2022survey}
Jiafei Duan, Samson Yu, Hui~Li Tan, Hongyuan Zhu, and Cheston Tan.
\newblock A survey of embodied {AI}: {F}rom simulators to research tasks.
\newblock \emph{IEEE Transactions on Emerging Topics in Computational
  Intelligence}, 6\penalty0 (2):\penalty0 230--244, 2022.
\newblock URL \url{https://ieeexplore.ieee.org/abstract/document/9687596/}.

\bibitem[Duvallet et~al.(2013)Duvallet, Kollar, and
  Stentz]{duvallet2013imitation}
Felix Duvallet, Thomas Kollar, and Anthony Stentz.
\newblock Imitation learning for natural language direction following through
  unknown environments.
\newblock In \emph{IEEE International Conference on Robotics and Automation},
  pp.\  1047--1053. IEEE, 2013.
\newblock URL \url{https://ieeexplore.ieee.org/abstract/document/6630702/}.

\bibitem[Duvallet et~al.(2016)Duvallet, Walter, Howard, Hemachandra, Oh,
  Teller, Roy, and Stentz]{duvallet2016inferring}
Felix Duvallet, Matthew~R Walter, Thomas Howard, Sachithra Hemachandra, Jean
  Oh, Seth Teller, Nicholas Roy, and Anthony Stentz.
\newblock Inferring maps and behaviors from natural language instructions.
\newblock In \emph{Experimental Robotics: The International Symposium on
  Experimental Robotics}, pp.\  373--388. Springer, 2016.
\newblock URL
  \url{https://link.springer.com/chapter/10.1007/978-3-319-23778-7_25}.

\bibitem[Elfes(1989)]{30720}
A.~Elfes.
\newblock Using occupancy grids for mobile robot perception and navigation.
\newblock \emph{Computer}, 22\penalty0 (6):\penalty0 46--57, 1989.
\newblock \doi{10.1109/2.30720}.
\newblock URL \url{https://ieeexplore.ieee.org/abstract/document/30720/}.

\bibitem[Endres et~al.(2013)Endres, Hess, Sturm, Cremers, and
  Burgard]{endres20133}
Felix Endres, J{\"u}rgen Hess, J{\"u}rgen Sturm, Daniel Cremers, and Wolfram
  Burgard.
\newblock {3-D mapping with an RGB-D camera}.
\newblock \emph{IEEE transactions on robotics}, 30\penalty0 (1):\penalty0
  177--187, 2013.
\newblock URL \url{https://ieeexplore.ieee.org/abstract/document/6594910/}.

\bibitem[Engel et~al.(2014)Engel, Sch{\"o}ps, and Cremers]{engel2014lsd}
Jakob Engel, Thomas Sch{\"o}ps, and Daniel Cremers.
\newblock {LSD-SLAM: Large-scale direct monocular SLAM}.
\newblock In \emph{European conference on computer vision}, pp.\  834--849.
  Springer, 2014.
\newblock URL
  \url{https://link.springer.com/chapter/10.1007/978-3-319-10605-2_54}.

\bibitem[Epstein \& Vass(2014)Epstein and Vass]{epstein2014neural}
Russell~A Epstein and Lindsay~K Vass.
\newblock Neural systems for landmark-based wayfinding in humans.
\newblock \emph{Philosophical Transactions of the Royal Society B: Biological
  Sciences}, 369\penalty0 (1635):\penalty0 20120533, 2014.
\newblock URL
  \url{https://royalsocietypublishing.org/doi/abs/10.1098/rstb.2012.0533}.

\bibitem[Fairfield(2009)]{fairfield2009localization}
Nathaniel Fairfield.
\newblock Localization, mapping, and planning in {3D} environments.
\newblock \emph{Ph. D. dissertation}, 2009.
\newblock URL
  \url{https://citeseerx.ist.psu.edu/document?repid=rep1\&type=pdf\&doi=478691d39bdfd6b9b44c85797dfc80d94232bc7e}.

\bibitem[Feng et~al.(2019)Feng, Meng, Shan, and Atanasov]{feng2019localization}
Qiaojun Feng, Yue Meng, Mo~Shan, and Nikolay Atanasov.
\newblock Localization and mapping using instance-specific mesh models.
\newblock In \emph{IEEE/RSJ International Conference on Intelligent Robots and
  Systems (IROS)}, pp.\  4985--4991. IEEE, 2019.
\newblock URL \url{https://ieeexplore.ieee.org/abstract/document/8967662/}.

\bibitem[Ferris et~al.(2007)Ferris, Fox, and Lawrence]{ferris2007wifi}
Brian Ferris, Dieter Fox, and Neil~D Lawrence.
\newblock {Wifi-slam using gaussian process latent variable models.}
\newblock In \emph{Proceedings of the International Joint Conference on
  Artificial Intelligence}, volume~7, pp.\  2480--2485, 2007.
\newblock URL \url{https://dl.acm.org/doi/abs/10.5555/1625275.1625675}.

\bibitem[Fioraio \& Di~Stefano(2013)Fioraio and Di~Stefano]{fioraio2013joint}
Nicola Fioraio and Luigi Di~Stefano.
\newblock Joint detection, tracking and mapping by semantic bundle adjustment.
\newblock In \emph{Proceedings of the IEEE Conference on Computer Vision and
  Pattern Recognition}, pp.\  1538--1545, 2013.
\newblock URL
  \url{http://openaccess.thecvf.com/content_cvpr_2013/html/Fioraio_Joint_Detection_Tracking_2013_CVPR_paper.html}.

\bibitem[Fischer et~al.(2024)Fischer, Porzi, Bulo, Pollefeys, and
  Kontschieder]{fischer2024multi}
Tobias Fischer, Lorenzo Porzi, Samuel~Rota Bulo, Marc Pollefeys, and Peter
  Kontschieder.
\newblock Multi-level neural scene graphs for dynamic urban environments.
\newblock In \emph{Proceedings of the IEEE/CVF Conference on Computer Vision
  and Pattern Recognition}, pp.\  21125--21135, 2024.
\newblock URL
  \url{http://openaccess.thecvf.com/content/CVPR2024/html/Fischer_Multi-Level_Neural_Scene_Graphs_for_Dynamic_Urban_Environments_CVPR_2024_paper.html}.

\bibitem[Flint et~al.(2011)Flint, Murray, and Reid]{flint2011manhattan}
Alex Flint, David Murray, and Ian Reid.
\newblock Manhattan scene understanding using monocular, stereo, and 3d
  features.
\newblock In \emph{International Conference on Computer Vision}, pp.\
  2228--2235. IEEE, 2011.
\newblock URL \url{https://ieeexplore.ieee.org/abstract/document/6126501/}.

\bibitem[Floros \& Leibe(2012)Floros and Leibe]{floros2012joint}
Georgios Floros and Bastian Leibe.
\newblock Joint 2d-3d temporally consistent semantic segmentation of street
  scenes.
\newblock In \emph{Proceedings of the IEEE/CVF Conference on Computer Vision
  and Pattern Recognition}, pp.\  2823--2830. IEEE, 2012.
\newblock URL \url{https://ieeexplore.ieee.org/abstract/document/6248007/}.

\bibitem[Foo et~al.(2005)Foo, Warren, Duchon, and Tarr]{foo2005humans}
Patrick Foo, William~H Warren, Andrew Duchon, and Michael~J Tarr.
\newblock Do humans integrate routes into a cognitive map? map-versus
  landmark-based navigation of novel shortcuts.
\newblock \emph{Journal of Experimental Psychology: Learning, Memory, and
  Cognition}, 31\penalty0 (2):\penalty0 195, 2005.
\newblock URL \url{https://psycnet.apa.org/record/2005-02160-002}.

\bibitem[Fox et~al.(1997)Fox, Burgard, and Thrun]{Fox1997TheDW}
Dieter Fox, Wolfram Burgard, and Sebastian Thrun.
\newblock The dynamic window approach to collision avoidance.
\newblock \emph{IEEE Robotics Autom. Mag.}, 4:\penalty0 23--33, 1997.
\newblock URL \url{https://api.semanticscholar.org/CorpusID:1718553}.

\bibitem[Fredriksson et~al.(2023)Fredriksson, Saradagi, and
  Nikolakopoulos]{fredriksson2023semantic}
Scott Fredriksson, Akshit Saradagi, and George Nikolakopoulos.
\newblock Semantic and topological mapping using intersection identification.
\newblock \emph{IFAC-PapersOnLine}, 56\penalty0 (2):\penalty0 9251--9256, 2023.
\newblock URL
  \url{https://www.sciencedirect.com/science/article/pii/S2405896323003415}.

\bibitem[Fu et~al.(2022)Fu, Zhang, Chen, Lu, Zhu, Zhou, Geiger, and
  Liao]{fu2022panoptic}
Xiao Fu, Shangzhan Zhang, Tianrun Chen, Yichong Lu, Lanyun Zhu, Xiaowei Zhou,
  Andreas Geiger, and Yiyi Liao.
\newblock Panoptic nerf: 3d-to-2d label transfer for panoptic urban scene
  segmentation.
\newblock In \emph{International Conference on 3D Vision (3DV)}, pp.\  1--11.
  IEEE, 2022.
\newblock URL \url{https://ieeexplore.ieee.org/abstract/document/10044395/}.

\bibitem[Gadre et~al.(2023)Gadre, Wortsman, Ilharco, Schmidt, and
  Song]{gadre2023cows}
Samir~Yitzhak Gadre, Mitchell Wortsman, Gabriel Ilharco, Ludwig Schmidt, and
  Shuran Song.
\newblock {Cows on pasture: Baselines and benchmarks for language-driven
  zero-shot object navigation}.
\newblock In \emph{Proceedings of the IEEE/CVF Conference on Computer Vision
  and Pattern Recognition}, pp.\  23171--23181, 2023.
\newblock URL
  \url{http://openaccess.thecvf.com/content/CVPR2023/html/Gadre_CoWs_on_Pasture_Baselines_and_Benchmarks_for_Language-Driven_Zero-Shot_Object_CVPR_2023_paper.html}.

\bibitem[Galindo et~al.(2005)Galindo, Saffiotti, Coradeschi, Buschka,
  Fernandez-Madrigal, and Gonz{\'a}lez]{galindo2005multi}
Cipriano Galindo, Alessandro Saffiotti, Silvia Coradeschi, P{\"a}r Buschka,
  Juan-Antonio Fernandez-Madrigal, and Javier Gonz{\'a}lez.
\newblock Multi-hierarchical semantic maps for mobile robotics.
\newblock In \emph{IEEE/RSJ international conference on intelligent robots and
  systems}, pp.\  2278--2283. IEEE, 2005.
\newblock URL \url{https://ieeexplore.ieee.org/abstract/document/1545511/}.

\bibitem[Gan et~al.(2019)Gan, Zhang, Wu, Gong, and Tenenbaum]{gan2019look}
Chuang Gan, Yiwei Zhang, Jiajun Wu, Boqing Gong, and Joshua~B Tenenbaum.
\newblock {Look, Listen, and Act: Towards Audio-Visual Embodied Navigation}.
\newblock \emph{arXiv preprint arXiv:1912.11684}, 2019.
\newblock URL \url{https://ieeexplore.ieee.org/abstract/document/9197008/}.

\bibitem[Garg et~al.(2020)Garg, S{\"u}nderhauf, Dayoub, Morrison, Cosgun,
  Carneiro, Wu, Chin, Reid, Gould, et~al.]{garg2020semantics}
Sourav Garg, Niko S{\"u}nderhauf, Feras Dayoub, Douglas Morrison, Akansel
  Cosgun, Gustavo Carneiro, Qi~Wu, Tat-Jun Chin, Ian Reid, Stephen Gould,
  et~al.
\newblock {Semantics for robotic mapping, perception and interaction: A
  survey}.
\newblock \emph{Foundations and Trends{\textregistered} in Robotics},
  8\penalty0 (1--2):\penalty0 1--224, 2020.
\newblock URL \url{https://www.nowpublishers.com/article/Details/ROB-059}.

\bibitem[Garg et~al.(2024)Garg, Rana, Hosseinzadeh, Mares, S{\"u}nderhauf,
  Dayoub, and Reid]{garg2024robohop}
Sourav Garg, Krishan Rana, Mehdi Hosseinzadeh, Lachlan Mares, Niko
  S{\"u}nderhauf, Feras Dayoub, and Ian Reid.
\newblock Robohop: Segment-based topological map representation for open-world
  visual navigation.
\newblock In \emph{IEEE International Conference on Robotics and Automation},
  2024.
\newblock URL \url{https://ieeexplore.ieee.org/abstract/document/10610234}.

\bibitem[Garrett et~al.(2020)Garrett, Chitnis, Holladay, Kim, Silver,
  Kaelbling, and Lozano-Perez]{Garrett2020IntegratedTA}
Caelan~Reed Garrett, Rohan Chitnis, Rachel Holladay, Beomjoon Kim, Tom Silver,
  Leslie~Pack Kaelbling, and Tomas Lozano-Perez.
\newblock {Integrated Task and Motion Planning}.
\newblock \emph{ArXiv}, abs/2010.01083, 2020.
\newblock URL \url{https://api.semanticscholar.org/CorpusID:222124824}.

\bibitem[Gautham et~al.(2023)Gautham, Sharma, Dhanalakshmi, and
  Ramamoorthy]{gautham20233d}
JS~Gautham, Akash Sharma, Samiappan Dhanalakshmi, and Kumar Ramamoorthy.
\newblock {3D} scene reconstruction and mapping with real time human detection
  for search and rescue robotics.
\newblock In \emph{AIP Conference Proceedings}, volume 2427. AIP Publishing,
  2023.
\newblock URL
  \url{https://pubs.aip.org/aip/acp/article-abstract/2427/1/020002/2866461}.

\bibitem[Georgakis et~al.(2021)Georgakis, Bucher, Schmeckpeper, Singh, and
  Daniilidis]{georgakis2021learning}
Georgios Georgakis, Bernadette Bucher, Karl Schmeckpeper, Siddharth Singh, and
  Kostas Daniilidis.
\newblock {L}earning to {M}ap for {A}ctive {S}emantic {G}oal {N}avigation.
\newblock In \emph{International Conference on Learning Representations}, 2021.
\newblock URL \url{https://par.nsf.gov/servlets/purl/10340845}.

\bibitem[Georgakis et~al.(2022{\natexlab{a}})Georgakis, Bucher, Arapin,
  Schmeckpeper, Matni, and Daniilidis]{georgakis2022uncertainty}
Georgios Georgakis, Bernadette Bucher, Anton Arapin, Karl Schmeckpeper, Nikolai
  Matni, and Kostas Daniilidis.
\newblock Uncertainty-driven planner for exploration and navigation.
\newblock In \emph{International Conference on Robotics and Automation (ICRA)},
  pp.\  11295--11302. IEEE, 2022{\natexlab{a}}.
\newblock URL \url{https://ieeexplore.ieee.org/abstract/document/9812423/}.

\bibitem[Georgakis et~al.(2022{\natexlab{b}})Georgakis, Bucher, Schmeckpeper,
  Singh, and Daniilidis]{georgakislearning}
Georgios Georgakis, Bernadette Bucher, Karl Schmeckpeper, Siddharth Singh, and
  Kostas Daniilidis.
\newblock Learning to map for active semantic goal navigation.
\newblock In \emph{International Conference on Learning Representations},
  2022{\natexlab{b}}.
\newblock URL \url{https://par.nsf.gov/servlets/purl/10340845}.

\bibitem[Georgakis et~al.(2022{\natexlab{c}})Georgakis, Schmeckpeper, Wanchoo,
  Dan, Miltsakaki, Roth, and Daniilidis]{georgakis2022cross}
Georgios Georgakis, Karl Schmeckpeper, Karan Wanchoo, Soham Dan, Eleni
  Miltsakaki, Dan Roth, and Kostas Daniilidis.
\newblock Cross-modal map learning for vision and language navigation.
\newblock In \emph{Proceedings of the IEEE/CVF Conference on Computer Vision
  and Pattern Recognition}, pp.\  15460--15470, 2022{\natexlab{c}}.
\newblock URL
  \url{http://openaccess.thecvf.com/content/CVPR2022/html/Georgakis_Cross-Modal_Map_Learning_for_Vision_and_Language_Navigation_CVPR_2022_paper.html}.

\bibitem[Gervet et~al.(2023)Gervet, Chintala, Batra, Malik, and
  Chaplot]{gervet2023navigating}
Theophile Gervet, Soumith Chintala, Dhruv Batra, Jitendra Malik, and
  Devendra~Singh Chaplot.
\newblock Navigating to objects in the real world.
\newblock \emph{Science Robotics}, 8\penalty0 (79):\penalty0 eadf6991, 2023.
\newblock URL
  \url{https://www.science.org/doi/abs/10.1126/scirobotics.adf6991}.

\bibitem[Goetting et~al.(2024)Goetting, Singh, and Loquercio]{goetting2024end}
Dylan Goetting, Himanshu~Gaurav Singh, and Antonio Loquercio.
\newblock End-to-{E}nd {N}avigation with {V}ision {L}anguage {M}odels:
  {T}ransforming {S}patial {R}easoning into {Q}uestion-{A}nswering.
\newblock \emph{arXiv preprint arXiv:2411.05755}, 2024.
\newblock URL \url{https://arxiv.org/abs/2411.05755}.

\bibitem[Gong \& Zhang(2018)Gong and Zhang]{gong2018temporal}
Ze~Gong and Yu~Zhang.
\newblock Temporal spatial inverse semantics for robots communicating with
  humans.
\newblock In \emph{IEEE International Conference on Robotics and Automation
  (ICRA)}, pp.\  4451--4458. IEEE, 2018.
\newblock URL \url{https://ieeexplore.ieee.org/abstract/document/8460754/}.

\bibitem[Gordon et~al.(2018)Gordon, Kembhavi, Rastegari, Redmon, Fox, and
  Farhadi]{gordon2018iqa}
Daniel Gordon, Aniruddha Kembhavi, Mohammad Rastegari, Joseph Redmon, Dieter
  Fox, and Ali Farhadi.
\newblock {IQA: Visual Question Answering in interactive environments}.
\newblock In \emph{Proceedings of the IEEE/CVF Conference on Computer Vision
  and Pattern Recognition}, pp.\  4089--4098, 2018.
\newblock URL
  \url{http://openaccess.thecvf.com/content_cvpr_2018/html/Gordon_IQA_Visual_Question_CVPR_2018_paper.html}.

\bibitem[Greve et~al.(2024)Greve, B{\"u}chner, V{\"o}disch, Burgard, and
  Valada]{greve2024collaborative}
Elias Greve, Martin B{\"u}chner, Niclas V{\"o}disch, Wolfram Burgard, and
  Abhinav Valada.
\newblock Collaborative dynamic 3d scene graphs for automated driving.
\newblock In \emph{IEEE International Conference on Robotics and Automation},
  pp.\  11118--11124. IEEE, 2024.
\newblock URL \url{https://ieeexplore.ieee.org/abstract/document/10610112/}.

\bibitem[Grinvald et~al.(2019)Grinvald, Furrer, Novkovic, Chung, Cadena,
  Siegwart, and Nieto]{grinvald2019volumetric}
Margarita Grinvald, Fadri Furrer, Tonci Novkovic, Jen~Jen Chung, Cesar Cadena,
  Roland Siegwart, and Juan Nieto.
\newblock Volumetric instance-aware semantic mapping and 3d object discovery.
\newblock \emph{IEEE Robotics and Automation Letters}, 4\penalty0 (3):\penalty0
  3037--3044, 2019.
\newblock URL \url{https://ieeexplore.ieee.org/abstract/document/8741085/}.

\bibitem[Gu et~al.(2023)Gu, Xiang, Li, Ling, Liu, Mu, Tang, Tao, Wei, Yao,
  Yuan, Xie, Huang, Chen, and Su]{gu2023maniskill2}
Jiayuan Gu, Fanbo Xiang, Xuanlin Li, Zhan Ling, Xiqiang Liu, Tongzhou Mu, Yihe
  Tang, Stone Tao, Xinyue Wei, Yunchao Yao, Xiaodi Yuan, Pengwei Xie, Zhiao
  Huang, Rui Chen, and Hao Su.
\newblock {ManiSkill2: A Unified Benchmark for Generalizable Manipulation
  Skills}.
\newblock In \emph{International Conference on Learning Representations}, 2023.
\newblock URL \url{https://openreview.net/pdf?id=b_CQDy9vrD1}.

\bibitem[Gu et~al.(2024)Gu, Kuwajerwala, Morin, Jatavallabhula, Sen, Agarwal,
  Rivera, Paul, Ellis, Chellappa, et~al.]{conceptgraphs}
Qiao Gu, Ali Kuwajerwala, Sacha Morin, Krishna~Murthy Jatavallabhula, Bipasha
  Sen, Aditya Agarwal, Corban Rivera, William Paul, Kirsty Ellis, Rama
  Chellappa, et~al.
\newblock {Conceptgraphs: Open-vocabulary 3d scene graphs for perception and
  planning}.
\newblock In \emph{IEEE International Conference on Robotics and Automation
  (ICRA)}, pp.\  5021--5028. IEEE, 2024.
\newblock URL \url{https://ieeexplore.ieee.org/abstract/document/10610243}.

\bibitem[Gu et~al.(2021)Gu, Lin, Kuo, and Cui]{gu2021open}
Xiuye Gu, Tsung-Yi Lin, Weicheng Kuo, and Yin Cui.
\newblock Open-vocabulary object detection via vision and language knowledge
  distillation.
\newblock In \emph{International Conference on Learning Representations}, 2021.
\newblock URL \url{https://iclr.cc/media/iclr-2022/Slides/6372.pdf}.

\bibitem[Guan et~al.(2024)Guan, Liao, Li, Hu, Yuan, Zhang, and
  Xu]{guan2024world}
Yanchen Guan, Haicheng Liao, Zhenning Li, Jia Hu, Runze Yuan, Guohui Zhang, and
  Chengzhong Xu.
\newblock {World models for autonomous driving: An initial survey}.
\newblock \emph{IEEE Transactions on Intelligent Vehicles}, 2024.
\newblock URL \url{https://ieeexplore.ieee.org/abstract/document/10522953}.

\bibitem[Guo et~al.(2024)Guo, Ma, Fan, Liu, and Li]{guo2024semantic}
Jun Guo, Xiaojian Ma, Yue Fan, Huaping Liu, and Qing Li.
\newblock Semantic {G}aussians: {O}pen-{V}ocabulary {S}cene {U}nderstanding
  with {3D} {G}aussian {S}platting.
\newblock \emph{arXiv preprint arXiv:2403.15624}, 2024.
\newblock URL \url{https://arxiv.org/abs/2403.15624}.

\bibitem[Gupta et~al.(2017)Gupta, Davidson, Levine, Sukthankar, and
  Malik]{gupta2017cognitive}
Saurabh Gupta, James Davidson, Sergey Levine, Rahul Sukthankar, and Jitendra
  Malik.
\newblock Cognitive mapping and planning for visual navigation.
\newblock In \emph{Proceedings of the IEEE/CVF Conference on Computer Vision
  and Pattern Recognition}, pp.\  2616--2625, 2017.
\newblock URL
  \url{http://openaccess.thecvf.com/content\_cvpr\_2017/html/Gupta\_Cognitive\_Mapping\_and\_CVPR\_2017\_paper.html}.

\bibitem[Hane et~al.(2013)Hane, Zach, Cohen, Angst, and
  Pollefeys]{hane2013joint}
Christian Hane, Christopher Zach, Andrea Cohen, Roland Angst, and Marc
  Pollefeys.
\newblock Joint 3{D} scene reconstruction and class segmentation.
\newblock In \emph{Proceedings of the IEEE Conference on Computer Vision and
  Pattern Recognition}, pp.\  97--104, 2013.
\newblock URL
  \url{http://openaccess.thecvf.com/content_cvpr_2013/html/Hane_Joint_3D_Scene_2013_CVPR_paper.html}.

\bibitem[He et~al.(2016)He, Zhang, Ren, and Sun]{he2016deep}
Kaiming He, Xiangyu Zhang, Shaoqing Ren, and Jian Sun.
\newblock Deep residual learning for image recognition.
\newblock In \emph{Proceedings of the IEEE/CVF Conference on Computer Vision
  and Pattern Recognition}, pp.\  770--778, 2016.
\newblock URL
  \url{http://openaccess.thecvf.com/content\_cvpr\_2016/html/He\_Deep\_Residual\_Learning\_CVPR\_2016\_paper.html}.

\bibitem[He et~al.(2017{\natexlab{a}})He, Gkioxari, Doll{\'a}r, and
  Girshick]{he2017mask}
Kaiming He, Georgia Gkioxari, Piotr Doll{\'a}r, and Ross Girshick.
\newblock Mask {R-CNN}.
\newblock In \emph{Proceedings of the IEEE/CVF International Conference on
  Computer Vision}, pp.\  2961--2969, 2017{\natexlab{a}}.
\newblock URL
  \url{http://openaccess.thecvf.com/content\_iccv\_2017/html/He\_Mask\_R-CNN\_ICCV\_2017\_paper.html}.

\bibitem[He et~al.(2017{\natexlab{b}})He, Liang, Yang, Li, and
  He]{he2017iterative}
Ying He, Bin Liang, Jun Yang, Shunzhi Li, and Jin He.
\newblock An iterative closest points algorithm for registration of 3{D} laser
  scanner point clouds with geometric features.
\newblock \emph{Sensors}, 17\penalty0 (8):\penalty0 1862, 2017{\natexlab{b}}.
\newblock URL \url{https://www.mdpi.com/1424-8220/17/8/1862}.

\bibitem[Hemachandra et~al.(2011)Hemachandra, Kollar, Roy, and
  Teller]{hemachandra2011following}
Sachithra Hemachandra, Thomas Kollar, Nicholas Roy, and Seth Teller.
\newblock Following and interpreting narrated guided tours.
\newblock In \emph{IEEE International Conference on Robotics and Automation},
  pp.\  2574--2579. IEEE, 2011.
\newblock URL \url{https://ieeexplore.ieee.org/abstract/document/5980209/}.

\bibitem[Hemachandra et~al.(2014)Hemachandra, Walter, Tellex, and
  Teller]{hemachandra2014learning}
Sachithra Hemachandra, Matthew~R Walter, Stefanie Tellex, and Seth Teller.
\newblock Learning spatial-semantic representations from natural language
  descriptions and scene classifications.
\newblock In \emph{IEEE International Conference on Robotics and Automation},
  pp.\  2623--2630. IEEE, 2014.
\newblock URL \url{https://ieeexplore.ieee.org/abstract/document/6907235/}.

\bibitem[Hemachandra et~al.(2015)Hemachandra, Duvallet, Howard, Roy, Stentz,
  and Walter]{hemachandra2015learning}
Sachithra Hemachandra, Felix Duvallet, Thomas~M Howard, Nicholas Roy, Anthony
  Stentz, and Matthew~R Walter.
\newblock Learning models for following natural language directions in unknown
  environments.
\newblock In \emph{IEEE International Conference on Robotics and Automation
  (ICRA)}, pp.\  5608--5615. IEEE, 2015.
\newblock URL \url{https://ieeexplore.ieee.org/abstract/document/7139984/}.

\bibitem[Henriques \& Vedaldi(2018)Henriques and Vedaldi]{henriques2018mapnet}
Joao~F Henriques and Andrea Vedaldi.
\newblock {MapNet}: An allocentric spatial memory for mapping environments.
\newblock In \emph{Proceedings of the IEEE/CVF Conference on Computer Vision
  and Pattern Recognition}, pp.\  8476--8484, 2018.
\newblock URL
  \url{http://openaccess.thecvf.com/content_cvpr_2018/html/Henriques_MapNet_An_Allocentric_CVPR_2018_paper}.

\bibitem[Hermans et~al.(2014)Hermans, Floros, and Leibe]{hermans2014dense}
Alexander Hermans, Georgios Floros, and Bastian Leibe.
\newblock Dense 3d semantic mapping of indoor scenes from rgb-d images.
\newblock In \emph{IEEE International Conference on Robotics and Automation},
  pp.\  2631--2638. IEEE, 2014.
\newblock URL \url{https://ieeexplore.ieee.org/abstract/document/6907236/}.

\bibitem[Hu et~al.(2013)Hu, Munoz, Bagnell, and Hebert]{hu2013efficient}
Hanzhang Hu, Daniel Munoz, J~Andrew Bagnell, and Martial Hebert.
\newblock Efficient 3-d scene analysis from streaming data.
\newblock In \emph{IEEE international conference on robotics and automation},
  pp.\  2297--2304. IEEE, 2013.
\newblock URL \url{https://ieeexplore.ieee.org/abstract/document/6630888/}.

\bibitem[Huang et~al.(2023{\natexlab{a}})Huang, Mees, Zeng, and
  Burgard]{huang2023visual}
Chenguang Huang, Oier Mees, Andy Zeng, and Wolfram Burgard.
\newblock Visual language maps for robot navigation.
\newblock In \emph{IEEE International Conference on Robotics and Automation},
  pp.\  10608--10615. IEEE, 2023{\natexlab{a}}.
\newblock URL \url{https://ieeexplore.ieee.org/abstract/document/10160969/}.

\bibitem[Huang et~al.(2011)Huang, Millman, Quigley, Stavens, Thrun, and
  Aggarwal]{huang2011efficient}
Joseph Huang, David Millman, Morgan Quigley, David Stavens, Sebastian Thrun,
  and Alok Aggarwal.
\newblock Efficient, generalized indoor wifi graphslam.
\newblock In \emph{IEEE international conference on robotics and automation},
  pp.\  1038--1043. IEEE, 2011.
\newblock URL \url{https://ieeexplore.ieee.org/abstract/document/5979643/}.

\bibitem[Huang et~al.(2023{\natexlab{b}})Huang, Wang, Zhang, Li, Wu, and
  Fei-Fei]{huang2023voxposer}
Wenlong Huang, Chen Wang, Ruohan Zhang, Yunzhu Li, Jiajun Wu, and Li~Fei-Fei.
\newblock {VoxPoser}: Composable {3D} value maps for robotic manipulation with
  language models.
\newblock In \emph{Conference on Robot Learning}, 2023{\natexlab{b}}.
\newblock URL \url{https://voxposer.github.io/}.

\bibitem[Hughes et~al.(2022)Hughes, Chang, and Carlone]{hughes2022hydra}
Nathan Hughes, Yun Chang, and Luca Carlone.
\newblock Hydra: {A} real-time spatial perception system for 3{D} scene graph
  construction and optimization.
\newblock \emph{arXiv preprint arXiv:2201.13360}, 2022.
\newblock URL \url{https://arxiv.org/abs/2201.13360}.

\bibitem[Hughes et~al.(2024)Hughes, Chang, Hu, Talak, Abdulhai, Strader, and
  Carlone]{hughes2024foundations}
Nathan Hughes, Yun Chang, Siyi Hu, Rajat Talak, Rumaia Abdulhai, Jared Strader,
  and Luca Carlone.
\newblock Foundations of spatial perception for robotics: {H}ierarchical
  representations and real-time systems.
\newblock \emph{The International Journal of Robotics Research}, pp.\
  02783649241229725, 2024.
\newblock URL
  \url{https://journals.sagepub.com/doi/abs/10.1177/02783649241229725}.

\bibitem[Ilharco et~al.(2019)Ilharco, Jain, Ku, Ie, and
  Baldridge]{ilharco2019generalevaluationinstructionconditioned}
Gabriel Ilharco, Vihan Jain, Alexander Ku, Eugene Ie, and Jason Baldridge.
\newblock {General Evaluation for Instruction Conditioned Navigation using
  Dynamic Time Warping}, 2019.
\newblock URL \url{https://arxiv.org/abs/1907.05446}.

\bibitem[Janzen \& Van~Turennout(2004)Janzen and
  Van~Turennout]{janzen2004selective}
Gabriele Janzen and Miranda Van~Turennout.
\newblock Selective neural representation of objects relevant for navigation.
\newblock \emph{Nature neuroscience}, 7\penalty0 (6):\penalty0 673--677, 2004.
\newblock URL \url{https://psycnet.apa.org/record/2004-14819-010}.

\bibitem[Jatavallabhula et~al.(2023)Jatavallabhula, Kuwajerwala, Gu, Omama,
  Chen, Li, Iyer, Saryazdi, Keetha, Tewari, Tenenbaum, {de Melo}, Krishna,
  Paull, Shkurti, and Torralba]{conceptfusion}
{Krishna Murthy} Jatavallabhula, Alihusein Kuwajerwala, Qiao Gu, Mohd Omama,
  Tao Chen, Shuang Li, Ganesh Iyer, Soroush Saryazdi, Nikhil Keetha, Ayush
  Tewari, {Joshua B.} Tenenbaum, {Celso Miguel} {de Melo}, Madhava Krishna,
  Liam Paull, Florian Shkurti, and Antonio Torralba.
\newblock Concept{F}usion: {O}pen-set {M}ultimodal 3{D} {M}apping.
\newblock \emph{Proceedings of Robotics: Science and System}, 2023.
\newblock URL \url{https://ieeexplore.ieee.org/abstract/document/10610193/}.

\bibitem[Johnson(2018)]{johnson2018topological}
Collin~Eugene Johnson.
\newblock \emph{Topological {M}apping and {N}avigation in {R}eal-World
  {E}nvironments}.
\newblock PhD thesis, University of Michigan, 2018.
\newblock URL \url{https://deepblue.lib.umich.edu/handle/2027.42/144014}.

\bibitem[Kaess et~al.(2012)Kaess, Fallon, Johannsson, and
  Leonard]{kaess2012kintinuous}
M~Kaess, M~Fallon, H~Johannsson, and JJ~Leonard.
\newblock Kintinuous: {S}patially extended kinectfusion.
\newblock In \emph{Proceedings of the RSS Workshop on RGB-D: Advanced Reasoning
  with Depth Cameras, Sydney, Australia}, pp.\  9--10, 2012.
\newblock URL \url{https://dspace.mit.edu/handle/1721.1/71756}.

\bibitem[Kaess(2015)]{kaess2015simultaneous}
Michael Kaess.
\newblock Simultaneous localization and mapping with infinite planes.
\newblock In \emph{IEEE International Conference on Robotics and Automation
  (ICRA)}, pp.\  4605--4611. IEEE, 2015.
\newblock URL \url{https://ieeexplore.ieee.org/abstract/document/7139837/}.

\bibitem[Kalapos et~al.(2020)Kalapos, G{\'o}r, Moni, and
  Harmati]{kalapos2020sim}
Andr{\'a}s Kalapos, Csaba G{\'o}r, R{\'o}bert Moni, and Istv{\'a}n Harmati.
\newblock Sim-to-real reinforcement learning applied to end-to-end vehicle
  control.
\newblock In \emph{International Symposium on Measurement and Control in
  Robotics (ISMCR)}, pp.\  1--6. IEEE, 2020.
\newblock URL \url{https://ieeexplore.ieee.org/abstract/document/9263751/}.

\bibitem[Kalashnikov et~al.(2018)Kalashnikov, Irpan, Pastor, Ibarz, Herzog,
  Jang, Quillen, Holly, Kalakrishnan, Vanhoucke, and
  Levine]{Kalashnikov2018QTOptSD}
Dmitry Kalashnikov, Alex Irpan, Peter Pastor, Julian Ibarz, Alexander Herzog,
  Eric Jang, Deirdre Quillen, Ethan Holly, Mrinal Kalakrishnan, Vincent
  Vanhoucke, and Sergey Levine.
\newblock {QT-Opt: Scalable Deep Reinforcement Learning for Vision-Based
  Robotic Manipulation}.
\newblock \emph{ArXiv}, abs/1806.10293, 2018.
\newblock URL \url{https://api.semanticscholar.org/CorpusID:49470584}.

\bibitem[Kassab et~al.(2025)Kassab, Morin, B{\"u}chner, Mattamala, Gupta,
  Valada, Paull, and Fallon]{kassab2025openlex3d}
Christina Kassab, Sacha Morin, Martin B{\"u}chner, Mat{\'\i}as Mattamala,
  Kumaraditya Gupta, Abhinav Valada, Liam Paull, and Maurice Fallon.
\newblock {OpenLex3D: A New Evaluation Benchmark for Open-Vocabulary 3D Scene
  Representations}.
\newblock \emph{arXiv preprint arXiv:2503.19764}, 2025.
\newblock URL \url{https://arxiv.org/abs/2503.19764}.

\bibitem[Kerr et~al.(2023)Kerr, Kim, Goldberg, Kanazawa, and
  Tancik]{kerr2023lerf}
Justin Kerr, Chung~Min Kim, Ken Goldberg, Angjoo Kanazawa, and Matthew Tancik.
\newblock {LeRF}: Language embedded radiance fields.
\newblock In \emph{Proceedings of the IEEE/CVF International Conference on
  Computer Vision}, pp.\  19729--19739, 2023.
\newblock URL
  \url{https://openaccess.thecvf.com/content/ICCV2023/html/Kerr_LERF_Language_Embedded_Radiance_Fields_ICCV_2023_paper.html}.

\bibitem[Khan et~al.(2021)Khan, Zaidi, Ishtiaq, Bukhari, Samer, and
  Farman]{khan2021comparative}
Misha~Urooj Khan, Syed Azhar~Ali Zaidi, Arslan Ishtiaq, Syeda Ume~Rubab
  Bukhari, Sana Samer, and Ayesha Farman.
\newblock A comparative survey of lidar-slam and lidar based sensor
  technologies.
\newblock In \emph{Mohammad Ali Jinnah University International Conference on
  Computing (MAJICC)}, pp.\  1--8. IEEE, 2021.
\newblock URL \url{https://ieeexplore.ieee.org/abstract/document/9526266/}.

\bibitem[Khandelwal et~al.(2022)Khandelwal, Weihs, Mottaghi, and
  Kembhavi]{khandelwal2022simple}
Apoorv Khandelwal, Luca Weihs, Roozbeh Mottaghi, and Aniruddha Kembhavi.
\newblock Simple but effective: {CLIP} embeddings for embodied {AI}.
\newblock \emph{Proceedings of the IEEE/CVF Conference on Computer Vision and
  Pattern Recognition}, 2022.
\newblock URL
  \url{http://openaccess.thecvf.com/content/CVPR2022/html/Khandelwal\_Simple\_but\_Effective\_CLIP\_Embeddings\_for\_Embodied\_AI\_CVPR\_2022\_paper.html}.

\bibitem[Khanna et~al.(2024)Khanna, Ramrakhya, Chhablani, Yenamandra, Gervet,
  Chang, Kira, Chaplot, Batra, and Mottaghi]{khanna2024goat}
Mukul Khanna, Ram Ramrakhya, Gunjan Chhablani, Sriram Yenamandra, Theophile
  Gervet, Matthew Chang, Zsolt Kira, Devendra~Singh Chaplot, Dhruv Batra, and
  Roozbeh Mottaghi.
\newblock Goat-bench: {A} benchmark for multi-modal lifelong navigation.
\newblock In \emph{Proceedings of the IEEE/CVF Conference on Computer Vision
  and Pattern Recognition}, pp.\  16373--16383, 2024.
\newblock URL
  \url{http://openaccess.thecvf.com/content/CVPR2024/html/Khanna_GOAT-Bench_A_Benchmark_for_Multi-Modal_Lifelong_Navigation_CVPR_2024_paper.html}.

\bibitem[Kim et~al.(2023)Kim, Kwon, Yoo, Choi, Park, and
  Oh]{kim2023topological}
Nuri Kim, Obin Kwon, Hwiyeon Yoo, Yunho Choi, Jeongho Park, and Songhwai Oh.
\newblock Topological semantic graph memory for image-goal navigation.
\newblock In \emph{Conference on Robot Learning}, pp.\  393--402. PMLR, 2023.
\newblock URL \url{https://proceedings.mlr.press/v205/kim23a.html}.

\bibitem[Kirillov et~al.(2023)Kirillov, Mintun, Ravi, Mao, Rolland, Gustafson,
  Xiao, Whitehead, Berg, Lo, et~al.]{kirillov2023segment}
Alexander Kirillov, Eric Mintun, Nikhila Ravi, Hanzi Mao, Chloe Rolland, Laura
  Gustafson, Tete Xiao, Spencer Whitehead, Alexander~C Berg, Wan-Yen Lo, et~al.
\newblock {Segment Anything}.
\newblock \emph{arXiv preprint arXiv:2304.02643}, 2023.
\newblock URL
  \url{http://openaccess.thecvf.com/content/ICCV2023/html/Kirillov_Segment_Anything_ICCV_2023_paper.html}.

\bibitem[Ko et~al.(2013)Ko, Yi, and Suh]{ko2013semantic}
Dong~Wook Ko, Chuho Yi, and Il~Hong Suh.
\newblock Semantic mapping and navigation: {A} {B}ayesian approach.
\newblock In \emph{IEEE/RSJ International Conference on Intelligent Robots and
  Systems}, pp.\  2630--2636. IEEE, 2013.
\newblock URL \url{https://ieeexplore.ieee.org/abstract/document/6696727/}.

\bibitem[Kolve et~al.(2017)Kolve, Mottaghi, Han, VanderBilt, Weihs, Herrasti,
  Deitke, Ehsani, Gordon, Zhu, Kembhavi, Gupta, and
  Farhadi]{Kolve2017AI2THORAI}
Eric Kolve, Roozbeh Mottaghi, Winson Han, Eli VanderBilt, Luca Weihs, Alvaro
  Herrasti, Matt Deitke, Kiana Ehsani, Daniel Gordon, Yuke Zhu, Aniruddha
  Kembhavi, Abhinav~Kumar Gupta, and Ali Farhadi.
\newblock {AI2-THOR: An Interactive 3D Environment for Visual AI}.
\newblock \emph{ArXiv}, abs/1712.05474, 2017.
\newblock URL \url{https://api.semanticscholar.org/CorpusID:28328610}.

\bibitem[Konolige et~al.(2011)Konolige, Marder-Eppstein, and
  Marthi]{konolige2011navigation}
Kurt Konolige, Eitan Marder-Eppstein, and Bhaskara Marthi.
\newblock Navigation in hybrid metric-topological maps.
\newblock In \emph{IEEE International Conference on Robotics and Automation},
  pp.\  3041--3047. IEEE, 2011.
\newblock URL \url{https://ieeexplore.ieee.org/abstract/document/5980074/}.

\bibitem[Kostavelis \& Gasteratos(2015)Kostavelis and
  Gasteratos]{kostavelis2015semantic}
Ioannis Kostavelis and Antonios Gasteratos.
\newblock Semantic mapping for mobile robotics tasks: {A} survey.
\newblock \emph{Robotics and Autonomous Systems}, 66:\penalty0 86--103, 2015.
\newblock URL
  \url{https://www.sciencedirect.com/science/article/pii/S0921889014003030}.

\bibitem[Krantz et~al.(2020)Krantz, Wijmans, Majumdar, Batra, and
  Lee]{krantz2020beyond}
Jacob Krantz, Erik Wijmans, Arjun Majumdar, Dhruv Batra, and Stefan Lee.
\newblock {Beyond the nav-graph: Vision-and-language navigation in continuous
  environments}.
\newblock In \emph{European Conference on Computer Vision (ECCV)}, pp.\
  104--120. Springer, 2020.
\newblock URL \url{https://jacobkrantz.github.io/vlnce/}.

\bibitem[Krantz et~al.(2022)Krantz, Lee, Malik, Batra, and
  Chaplot]{krantz2022instance}
Jacob Krantz, Stefan Lee, Jitendra Malik, Dhruv Batra, and Devendra~Singh
  Chaplot.
\newblock Instance-specific image goal navigation: Training embodied agents to
  find object instances.
\newblock \emph{arXiv preprint arXiv:2211.15876}, 2022.
\newblock URL \url{https://arxiv.org/abs/2211.15876}.

\bibitem[Kundu et~al.(2014)Kundu, Li, Dellaert, Li, and Rehg]{kundu2014joint}
Abhijit Kundu, Yin Li, Frank Dellaert, Fuxin Li, and James~M Rehg.
\newblock {Joint semantic segmentation and 3d reconstruction from monocular
  video}.
\newblock In \emph{European Conference on Computer Vision}, pp.\  703--718.
  Springer, 2014.
\newblock URL
  \url{https://link.springer.com/chapter/10.1007/978-3-319-10599-4_45}.

\bibitem[Kwon et~al.(2021)Kwon, Kim, Choi, Yoo, Park, and Oh]{kwon2021visual}
Obin Kwon, Nuri Kim, Yunho Choi, Hwiyeon Yoo, Jeongho Park, and Songhwai Oh.
\newblock Visual graph memory with unsupervised representation for visual
  navigation.
\newblock In \emph{Proceedings of the IEEE/CVF International Conference on
  Computer Vision}, pp.\  15890--15899, 2021.
\newblock URL
  \url{http://openaccess.thecvf.com/content/ICCV2021/html/Kwon_Visual_Graph_Memory_With_Unsupervised_Representation_for_Visual_Navigation_ICCV_2021_paper.html}.

\bibitem[Kwon et~al.(2023)Kwon, Park, and Oh]{kwon2023renderable}
Obin Kwon, Jeongho Park, and Songhwai Oh.
\newblock Renderable neural radiance map for visual navigation.
\newblock In \emph{Proceedings of the IEEE/CVF Conference on Computer Vision
  and Pattern Recognition}, pp.\  9099--9108, 2023.
\newblock URL
  \url{http://openaccess.thecvf.com/content/CVPR2023/html/Kwon_Renderable_Neural_Radiance_Map_for_Visual_Navigation_CVPR_2023_paper.html}.

\bibitem[Lai et~al.(2014)Lai, Bo, and Fox]{lai2014unsupervised}
Kevin Lai, Liefeng Bo, and Dieter Fox.
\newblock Unsupervised feature learning for 3d scene labeling.
\newblock In \emph{IEEE International Conference on Robotics and Automation
  (ICRA)}, pp.\  3050--3057. IEEE, 2014.
\newblock URL \url{https://ieeexplore.ieee.org/abstract/document/6907298/}.

\bibitem[Landsiedel et~al.(2017)Landsiedel, Rieser, Walter, and
  Wollherr]{landsiedel2017review}
Christian Landsiedel, Verena Rieser, Matthew Walter, and Dirk Wollherr.
\newblock A review of spatial reasoning and interaction for real-world
  robotics.
\newblock \emph{Advanced Robotics}, 31\penalty0 (5):\penalty0 222--242, 2017.
\newblock URL
  \url{https://www.tandfonline.com/doi/abs/10.1080/01691864.2016.1277554}.

\bibitem[Lee et~al.(2025)Lee, Kong, Park, and Kim]{lee2025geomgs}
Jaewon Lee, Mangyu Kong, Minseong Park, and Euntai Kim.
\newblock {GeomGS: LiDAR-Guided Geometry-Aware Gaussian Splatting for Robot
  Localization}.
\newblock \emph{arXiv preprint arXiv:2501.13417}, 2025.
\newblock URL \url{https://arxiv.org/abs/2501.13417}.

\bibitem[Lei et~al.(2025)Lei, Wang, Zhou, and Li]{lei2025gaussnav}
Xiaohan Lei, Min Wang, Wengang Zhou, and Houqiang Li.
\newblock {Gaussnav: Gaussian splatting for visual navigation}.
\newblock \emph{IEEE Transactions on Pattern Analysis and Machine
  Intelligence}, 2025.
\newblock URL \url{https://arxiv.org/abs/2403.11625}.

\bibitem[Li et~al.(2022)Li, Weinberger, Belongie, Koltun, and
  Ranftl]{li2022languagedriven}
Boyi Li, Kilian~Q Weinberger, Serge Belongie, Vladlen Koltun, and Rene Ranftl.
\newblock Language-driven semantic segmentation.
\newblock In \emph{International Conference on Learning Representations}, 2022.
\newblock URL \url{https://openreview.net/forum?id=RriDjddCLN}.

\bibitem[Li et~al.(2021{\natexlab{a}})Li, Zhou, Xiong, and
  Hoi]{li2021prototypical}
Junnan Li, Pan Zhou, Caiming Xiong, and Steven Hoi.
\newblock Prototypical contrastive learning of unsupervised representations.
\newblock In \emph{International Conference on Learning Representations},
  2021{\natexlab{a}}.
\newblock URL \url{https://openreview.net/forum?id=KmykpuSrjcq}.

\bibitem[Li et~al.(2023)Li, Li, Savarese, and Hoi]{li2023blip}
Junnan Li, Dongxu Li, Silvio Savarese, and Steven Hoi.
\newblock Blip-2: Bootstrapping language-image pre-training with frozen image
  encoders and large language models.
\newblock \emph{arXiv preprint arXiv:2301.12597}, 2023.
\newblock URL \url{https://proceedings.mlr.press/v202/li23q}.

\bibitem[Li et~al.(2021{\natexlab{b}})Li, Kong, Zhao, Li, Wen, Zhang, and
  Liu]{Li2021SALOAMSL}
Lin Li, Xin Kong, Xiangrui Zhao, Wanlong Li, Feng Wen, Hongbo Zhang, and Yong
  Liu.
\newblock {SA-LOAM: Semantic-aided LiDAR SLAM with Loop Closure}.
\newblock \emph{IEEE International Conference on Robotics and Automation
  (ICRA)}, pp.\  7627--7634, 2021{\natexlab{b}}.
\newblock URL \url{https://api.semanticscholar.org/CorpusID:235592935}.

\bibitem[Li et~al.(2024{\natexlab{a}})Li, Liu, Zhou, Zhu, Cheng, Deng, and
  Wang]{li2024sgs}
Mingrui Li, Shuhong Liu, Heng Zhou, Guohao Zhu, Na~Cheng, Tianchen Deng, and
  Hongyu Wang.
\newblock {SGS-SLAM: Semantic Gaussian splatting for neural dense SLAM}.
\newblock In \emph{European Conference on Computer Vision}, pp.\  163--179.
  Springer, 2024{\natexlab{a}}.
\newblock URL
  \url{https://link.springer.com/chapter/10.1007/978-3-031-72751-1_10}.

\bibitem[Li et~al.(2024{\natexlab{b}})Li, Jia, Wang, Chen, Jiang, Yan, and
  Li]{li2024lanesegnet}
Tianyu Li, Peijin Jia, Bangjun Wang, Li~Chen, Kun Jiang, Junchi Yan, and
  Hongyang Li.
\newblock {LaneSegNet: Map Learning with Lane Segment Perception for Autonomous
  Driving}.
\newblock In \emph{International Conference on Learning Representations},
  2024{\natexlab{b}}.
\newblock URL \url{https://hub.hku.hk/handle/10722/351367}.

\bibitem[Li et~al.(2021{\natexlab{c}})Li, Song, Bai, Zhang, and
  Jiang]{li2021ion}
Weijie Li, Xinhang Song, Yubing Bai, Sixian Zhang, and Shuqiang Jiang.
\newblock {Ion: Instance-level object navigation}.
\newblock In \emph{Proceedings of the ACM international conference on
  multimedia}, pp.\  4343--4352, 2021{\natexlab{c}}.
\newblock URL \url{https://dl.acm.org/doi/abs/10.1145/3474085.3475575}.

\bibitem[Li et~al.(2024{\natexlab{c}})Li, Wang, Li, Xie, Sima, Lu, Yu, and
  Dai]{li2024bevformer}
Zhiqi Li, Wenhai Wang, Hongyang Li, Enze Xie, Chonghao Sima, Tong Lu, Qiao Yu,
  and Jifeng Dai.
\newblock Bevformer: learning bird's-eye-view representation from lidar-camera
  via spatiotemporal transformers.
\newblock \emph{IEEE Transactions on Pattern Analysis and Machine
  Intelligence}, 2024{\natexlab{c}}.
\newblock URL \url{https://ieeexplore.ieee.org/abstract/document/10791908/}.

\bibitem[Lin et~al.(2024)Lin, Lee, and Lu]{lin2024embodied}
Ming-Yi Lin, Ou-Wen Lee, and Chih-Ying Lu.
\newblock Embodied {AI} with {L}arge {L}anguage {M}odels: {A} {S}urvey and
  {N}ew {HRI} {F}ramework.
\newblock In \emph{International Conference on Advanced Robotics and
  Mechatronics (ICARM)}, pp.\  978--983. IEEE, 2024.
\newblock URL \url{https://ieeexplore.ieee.org/abstract/document/10715872/}.

\bibitem[Lin et~al.(2022)Lin, Wang, Liu, and Qiu]{lin2022survey}
Tianyang Lin, Yuxin Wang, Xiangyang Liu, and Xipeng Qiu.
\newblock A survey of transformers.
\newblock \emph{AI open}, 3:\penalty0 111--132, 2022.
\newblock URL
  \url{https://www.sciencedirect.com/science/article/pii/S2666651022000146}.

\bibitem[Lin et~al.(2020)Lin, Zeng, Song, Isola, and Lin]{lin2020learning}
Yen-Chen Lin, Andy Zeng, Shuran Song, Phillip Isola, and Tsung-Yi Lin.
\newblock Learning to see before learning to act: Visual pre-training for
  manipulation.
\newblock In \emph{IEEE International Conference on Robotics and Automation},
  2020.
\newblock URL \url{https://ieeexplore.ieee.org/abstract/document/9197331/}.

\bibitem[Liu et~al.(2023{\natexlab{a}})Liu, Li, Wu, and Lee]{liu2023visual}
Haotian Liu, Chunyuan Li, Qingyang Wu, and Yong~Jae Lee.
\newblock Visual instruction tuning.
\newblock \emph{Advances in neural information processing systems},
  36:\penalty0 34892--34916, 2023{\natexlab{a}}.
\newblock URL
  \url{https://proceedings.neurips.cc/paper_files/paper/2023/hash/6dcf277ea32ce3288914faf369fe6de0-Abstract-Conference.html}.

\bibitem[Liu et~al.(2012)Liu, Tuzel, Veeraraghavan, Taguchi, Marks, and
  Chellappa]{Liu2012FastOL}
Ming-Yu Liu, Oncel Tuzel, Ashok Veeraraghavan, Yuichi Taguchi, Tim~K. Marks,
  and Rama Chellappa.
\newblock Fast object localization and pose estimation in heavy clutter for
  robotic bin picking.
\newblock \emph{The International Journal of Robotics Research}, 31:\penalty0
  951 -- 973, 2012.
\newblock URL \url{https://api.semanticscholar.org/CorpusID:7197454}.

\bibitem[Liu et~al.(2024)Liu, Zeng, Ren, Li, Zhang, Yang, Jiang, Li, Yang, Su,
  et~al.]{liu2024grounding}
Shilong Liu, Zhaoyang Zeng, Tianhe Ren, Feng Li, Hao Zhang, Jie Yang, Qing
  Jiang, Chunyuan Li, Jianwei Yang, Hang Su, et~al.
\newblock Grounding dino: Marrying dino with grounded pre-training for open-set
  object detection.
\newblock In \emph{European conference on computer vision}, pp.\  38--55.
  Springer, 2024.
\newblock URL
  \url{https://link.springer.com/chapter/10.1007/978-3-031-72970-6_3}.

\bibitem[Liu et~al.(2019)Liu, P{\'e}tillot, Lane, and Wang]{Liu2019GlobalLW}
Yu~Liu, Yvan~R. P{\'e}tillot, David~M. Lane, and Sen Wang.
\newblock {Global Localization with Object-Level Semantics and Topology}.
\newblock \emph{International Conference on Robotics and Automation (ICRA)},
  pp.\  4909--4915, 2019.
\newblock URL \url{https://api.semanticscholar.org/CorpusID:182012871}.

\bibitem[Liu et~al.(2023{\natexlab{b}})Liu, Tang, Amini, Yang, Mao, Rus, and
  Han]{liu2023bevfusion}
Zhijian Liu, Haotian Tang, Alexander Amini, Xinyu Yang, Huizi Mao, Daniela~L
  Rus, and Song Han.
\newblock {Bevfusion: Multi-task multi-sensor fusion with unified bird's-eye
  view representation}.
\newblock In \emph{IEEE international conference on robotics and automation
  (ICRA)}, pp.\  2774--2781. IEEE, 2023{\natexlab{b}}.
\newblock URL \url{https://ieeexplore.ieee.org/abstract/document/10160968/}.

\bibitem[Lluvia et~al.(2021)Lluvia, Lazkano, and Ansuategi]{lluvia2021active}
Iker Lluvia, Elena Lazkano, and Ander Ansuategi.
\newblock Active mapping and robot exploration: {A} survey.
\newblock \emph{Sensors}, 21\penalty0 (7):\penalty0 2445, 2021.
\newblock URL \url{https://www.mdpi.com/1424-8220/21/7/2445}.

\bibitem[Loeliger(2004)]{loeliger2004introduction}
H-A Loeliger.
\newblock An introduction to factor graphs.
\newblock \emph{IEEE Signal Processing Magazine}, 21\penalty0 (1):\penalty0
  28--41, 2004.
\newblock URL \url{https://ieeexplore.ieee.org/abstract/document/1267047/}.

\bibitem[Long et~al.(2024)Long, Cai, Wang, Zhan, and Dong]{longinstructnav}
Yuxing Long, Wenzhe Cai, Hongcheng Wang, Guanqi Zhan, and Hao Dong.
\newblock Instruct{N}av: {Z}ero-shot {S}ystem for {G}eneric {I}nstruction
  {N}avigation in {U}nexplored {E}nvironment.
\newblock In \emph{Conference on Robot Learning}, 2024.
\newblock URL \url{https://openreview.net/forum?id=fCDOfpTCzZ}.

\bibitem[Lorbach et~al.(2014)Lorbach, H{\"o}fer, and Brock]{lorbach2014prior}
Malte Lorbach, Sebastian H{\"o}fer, and Oliver Brock.
\newblock Prior-assisted propagation of spatial information for object search.
\newblock In \emph{IEEE/RSJ International Conference on Intelligent Robots and
  Systems}, pp.\  2904--2909. IEEE, 2014.
\newblock URL \url{https://ieeexplore.ieee.org/abstract/document/6942962/}.

\bibitem[Lowe(2004)]{lowe2004distinctive}
David~G Lowe.
\newblock Distinctive image features from scale-invariant keypoints.
\newblock \emph{International journal of computer vision}, 60:\penalty0
  91--110, 2004.
\newblock URL
  \url{https://3d.bk.tudelft.nl/courses/geo1016/handouts/05-image_matching.pdf}.

\bibitem[Luo et~al.(2022)Luo, Yue, Hong, and Agrawal]{luo2022stubborn}
Haokuan Luo, Albert Yue, Zhang-Wei Hong, and Pulkit Agrawal.
\newblock Stubborn: A strong baseline for indoor object navigation.
\newblock \emph{arXiv preprint arXiv:2203.07359}, 2022.
\newblock URL \url{https://ieeexplore.ieee.org/abstract/document/9981646/}.

\bibitem[Luperto et~al.(2024)Luperto, Ferrara, Boracchi, and
  Amigoni]{luperto2024estimating}
Matteo Luperto, Marco~Maria Ferrara, Giacomo Boracchi, and Francesco Amigoni.
\newblock Estimating {M}ap {C}ompleteness in {R}obot {E}xploration.
\newblock \emph{arXiv preprint arXiv:2406.13482}, 2024.
\newblock URL \url{https://arxiv.org/abs/2406.13482}.

\bibitem[MacMahon et~al.(2006)MacMahon, Stankiewicz, and
  Kuipers]{macmahon2006walk}
Matt MacMahon, Brian Stankiewicz, and Benjamin Kuipers.
\newblock Walk the talk: {C}onnecting language, knowledge, and action in route
  instructions.
\newblock \emph{Def}, 2\penalty0 (6):\penalty0 4, 2006.
\newblock URL \url{https://cdn.aaai.org/AAAI/2006/AAAI06-232.pdf}.

\bibitem[Maggio et~al.(2024)Maggio, Chang, Hughes, Trang, Griffith, Dougherty,
  Cristofalo, Schmid, and Carlone]{maggio2024clio}
Dominic Maggio, Yun Chang, Nathan Hughes, Matthew Trang, Dan Griffith, Carlyn
  Dougherty, Eric Cristofalo, Lukas Schmid, and Luca Carlone.
\newblock Clio: {R}eal-time task-driven open-set 3d scene graphs.
\newblock \emph{IEEE Robotics and Automation Letters}, 2024.
\newblock URL \url{https://ieeexplore.ieee.org/abstract/document/10659066/}.

\bibitem[Magnusson et~al.(2009)Magnusson, Andreasson, Nuchter, and
  Lilienthal]{5152712}
Martin Magnusson, Henrik Andreasson, Andreas Nuchter, and Achim~J. Lilienthal.
\newblock {Appearance-based loop detection from 3D laser data using the normal
  distributions transform}.
\newblock In \emph{IEEE International Conference on Robotics and Automation},
  pp.\  23--28, 2009.
\newblock \doi{10.1109/ROBOT.2009.5152712}.
\newblock URL \url{https://ieeexplore.ieee.org/abstract/document/5152712/}.

\bibitem[Mahendran et~al.(2016)Mahendran, Bilen, Henriques, and
  Vedaldi]{mahendran2016researchdoom}
Aravindh Mahendran, Hakan Bilen, Jo{\~a}o~F Henriques, and Andrea Vedaldi.
\newblock Researchdoom and cocodoom: Learning computer vision with games.
\newblock \emph{arXiv preprint arXiv:1610.02431}, 2016.
\newblock URL \url{https://arxiv.org/abs/1610.02431}.

\bibitem[Majumdar et~al.(2024)Majumdar, Ajay, Zhang, Putta, Yenamandra, Henaff,
  Silwal, Mcvay, Maksymets, Arnaud, Yadav, Li, Newman, Sharma, Berges, Zhang,
  Agrawal, Bisk, Batra, Kalakrishnan, Meier, Paxton, Sax, and
  Rajeswaran]{Majumdar2024OpenEQAEQ}
Arjun Majumdar, Anurag Ajay, Xiaohan Zhang, Pranav Putta, Sriram Yenamandra,
  Mikael Henaff, Sneha Silwal, Paul Mcvay, Oleksandr Maksymets, Sergio Arnaud,
  Karmesh Yadav, Qiyang Li, Ben Newman, Mohit Sharma, Vincent-Pierre Berges,
  Shiqi Zhang, Pulkit Agrawal, Yonatan Bisk, Dhruv Batra, Mrinal Kalakrishnan,
  Franziska Meier, Chris Paxton, Alexander Sax, and Aravind Rajeswaran.
\newblock {OpenEQA: Embodied Question Answering in the Era of Foundation
  Models}.
\newblock \emph{IEEE/CVF Conference on Computer Vision and Pattern Recognition
  (CVPR)}, pp.\  16488--16498, 2024.
\newblock URL \url{https://api.semanticscholar.org/CorpusID:268066655}.

\bibitem[Maturana et~al.(2018)Maturana, Chou, Uenoyama, and
  Scherer]{maturana2018real}
Daniel Maturana, Po-Wei Chou, Masashi Uenoyama, and Sebastian Scherer.
\newblock Real-time semantic mapping for autonomous off-road navigation.
\newblock In \emph{Field and Service Robotics: Results of the International
  Conference}, pp.\  335--350. Springer, 2018.
\newblock URL
  \url{https://link.springer.com/chapter/10.1007/978-3-319-67361-5_22}.

\bibitem[Matuszek et~al.(2010)Matuszek, Fox, and
  Koscher]{matuszek2010following}
Cynthia Matuszek, Dieter Fox, and Karl Koscher.
\newblock Following directions using statistical machine translation.
\newblock In \emph{ACM/IEEE International Conference on Human-Robot Interaction
  (HRI)}, pp.\  251--258. IEEE, 2010.
\newblock URL \url{https://ieeexplore.ieee.org/abstract/document/5453189/}.

\bibitem[Matuszek et~al.(2012)Matuszek, Herbst, Zettlemoyer, and
  Fox]{Matuszek2012LearningTP}
Cynthia Matuszek, Evan~V. Herbst, Luke Zettlemoyer, and Dieter Fox.
\newblock {Learning to Parse Natural Language Commands to a Robot Control
  System}.
\newblock In \emph{International Symposium on Experimental Robotics}, 2012.
\newblock URL \url{https://api.semanticscholar.org/CorpusID:1658890}.

\bibitem[Matuszek et~al.(2013)Matuszek, Herbst, Zettlemoyer, and
  Fox]{matuszek2013learning}
Cynthia Matuszek, Evan Herbst, Luke Zettlemoyer, and Dieter Fox.
\newblock Learning to parse natural language commands to a robot control
  system.
\newblock In \emph{Experimental robotics: the international symposium on
  experimental robotics}, pp.\  403--415. Springer, 2013.
\newblock URL
  \url{https://link.springer.com/chapter/10.1007/978-3-319-00065-7_28}.

\bibitem[May et~al.(2009)May, Dr{\"o}schel, Fuchs, Holz, and
  N{\"u}chter]{may2009robust}
Stefan May, David Dr{\"o}schel, Stefan Fuchs, Dirk Holz, and Andreas
  N{\"u}chter.
\newblock Robust 3{D}-mapping with time-of-flight cameras.
\newblock In \emph{IEEE/RSJ International Conference on Intelligent Robots and
  Systems}, pp.\  1673--1678. IEEE, 2009.
\newblock URL \url{https://ieeexplore.ieee.org/abstract/document/5354684/}.

\bibitem[Mazur et~al.(2023)Mazur, Sucar, and Davison]{mazur2023feature}
Kirill Mazur, Edgar Sucar, and Andrew~J Davison.
\newblock Feature-realistic neural fusion for real-time, open set scene
  understanding.
\newblock In \emph{IEEE International Conference on Robotics and Automation},
  pp.\  8201--8207. IEEE, 2023.
\newblock URL \url{https://ieeexplore.ieee.org/abstract/document/10160800/}.

\bibitem[Mazzaglia et~al.(2022)Mazzaglia, Catal, Verbelen, and
  Dhoedt]{mazzaglia2022curiosity}
Pietro Mazzaglia, Ozan Catal, Tim Verbelen, and Bart Dhoedt.
\newblock Curiosity-driven exploration via latent bayesian surprise.
\newblock In \emph{Proceedings of the AAAI conference on artificial
  intelligence}, volume~36, pp.\  7752--7760, 2022.
\newblock URL \url{https://ojs.aaai.org/index.php/AAAI/article/view/20743}.

\bibitem[McCormac et~al.(2017)McCormac, Handa, Davison, and
  Leutenegger]{mccormac2017semanticfusion}
John McCormac, Ankur Handa, Andrew Davison, and Stefan Leutenegger.
\newblock Semanticfusion: {D}ense 3d semantic mapping with convolutional neural
  networks.
\newblock In \emph{IEEE International Conference on Robotics and Automation},
  pp.\  4628--4635. IEEE, 2017.
\newblock URL \url{https://ieeexplore.ieee.org/abstract/document/7989538/}.

\bibitem[McCormac et~al.(2018)McCormac, Clark, Bloesch, Davison, and
  Leutenegger]{mccormac2018fusion++}
John McCormac, Ronald Clark, Michael Bloesch, Andrew Davison, and Stefan
  Leutenegger.
\newblock Fusion++: {V}olumetric object-level slam.
\newblock In \emph{International Conference on 3D Vision (3DV)}, pp.\  32--41.
  IEEE, 2018.
\newblock URL \url{https://ieeexplore.ieee.org/abstract/document/8490953/}.

\bibitem[Meger et~al.(2008)Meger, Forss{\'e}n, Lai, Helmer, McCann, Southey,
  Baumann, Little, and Lowe]{meger2008curious}
David Meger, Per-Erik Forss{\'e}n, Kevin Lai, Scott Helmer, Sancho McCann,
  Tristram Southey, Matthew Baumann, James~J Little, and David~G Lowe.
\newblock Curious george: {A}n attentive semantic robot.
\newblock \emph{Robotics and Autonomous Systems}, 56\penalty0 (6):\penalty0
  503--511, 2008.
\newblock URL
  \url{https://www.sciencedirect.com/science/article/pii/S0921889008000316}.

\bibitem[Mehan et~al.(2024)Mehan, Gupta, Jayanti, Govil, Garg, and
  Krishna]{mehan2024questmaps}
Yash Mehan, Kumaraditya Gupta, Rohit Jayanti, Anirudh Govil, Sourav Garg, and
  Madhava Krishna.
\newblock Que{STM}aps: {Q}ueryable {S}emantic {T}opological {M}aps for {3D}
  {S}cene {U}nderstanding.
\newblock \emph{arXiv preprint arXiv:2404.06442}, 2024.
\newblock URL \url{https://ieeexplore.ieee.org/abstract/document/10801814/}.

\bibitem[Mei et~al.(2016)Mei, Bansal, and Walter]{mei2016listen}
Hongyuan Mei, Mohit Bansal, and Matthew Walter.
\newblock Listen, attend, and walk: {N}eural mapping of navigational
  instructions to action sequences.
\newblock In \emph{Proceedings of the AAAI Conference on Artificial
  Intelligence}, volume~30, 2016.
\newblock URL \url{https://ojs.aaai.org/index.php/AAAI/article/view/10364}.

\bibitem[Mescheder et~al.(2019)Mescheder, Oechsle, Niemeyer, Nowozin, and
  Geiger]{mescheder2019occupancy}
Lars Mescheder, Michael Oechsle, Michael Niemeyer, Sebastian Nowozin, and
  Andreas Geiger.
\newblock {Occupancy networks: Learning 3d reconstruction in function space}.
\newblock In \emph{Proceedings of the IEEE/CVF conference on computer vision
  and pattern recognition}, pp.\  4460--4470, 2019.
\newblock URL
  \url{http://openaccess.thecvf.com/content_CVPR_2019/html/Mescheder_Occupancy_Networks_Learning_3D_Reconstruction_in_Function_Space_CVPR_2019_paper.html}.

\bibitem[Mildenhall et~al.(2021)Mildenhall, Srinivasan, Tancik, Barron,
  Ramamoorthi, and Ng]{mildenhall2021nerf}
Ben Mildenhall, Pratul~P Srinivasan, Matthew Tancik, Jonathan~T Barron, Ravi
  Ramamoorthi, and Ren Ng.
\newblock {Nerf: Representing scenes as neural radiance fields for view
  synthesis}.
\newblock \emph{Communications of the ACM}, 65\penalty0 (1):\penalty0 99--106,
  2021.
\newblock URL \url{https://dl.acm.org/doi/abs/10.1145/3503250}.

\bibitem[Miller et~al.(2021)Miller, Cowley, Konkimalla, Shivakumar, Nguyen,
  Smith, Taylor, and Kumar]{miller2021any}
Ian~D Miller, Anthony Cowley, Ravi Konkimalla, Shreyas~S Shivakumar, Ty~Nguyen,
  Trey Smith, Camillo~Jose Taylor, and Vijay Kumar.
\newblock Any way you look at it: {S}emantic crossview localization and mapping
  with lidar.
\newblock \emph{IEEE Robotics and Automation Letters}, 6\penalty0 (2):\penalty0
  2397--2404, 2021.
\newblock URL \url{https://ieeexplore.ieee.org/abstract/document/9361130/}.

\bibitem[Miller et~al.(2022)Miller, Cladera, Smith, Taylor, and
  Kumar]{miller2022stronger}
Ian~D Miller, Fernando Cladera, Trey Smith, Camillo~Jose Taylor, and Vijay
  Kumar.
\newblock Stronger together: {A}ir-ground robotic collaboration using
  semantics.
\newblock \emph{IEEE Robotics and Automation Letters}, 7\penalty0 (4):\penalty0
  9643--9650, 2022.
\newblock URL \url{https://ieeexplore.ieee.org/abstract/document/9830880/}.

\bibitem[Min \& Dunn(2021)Min and Dunn]{min2021voldor+}
Zhixiang Min and Enrique Dunn.
\newblock Voldor+ slam: {F}or the times when feature-based or direct methods
  are not good enough.
\newblock In \emph{IEEE International Conference on Robotics and Automation},
  pp.\  13813--13819. IEEE, 2021.
\newblock URL \url{https://ieeexplore.ieee.org/abstract/document/9561230/}.

\bibitem[Mozos et~al.(2007)Mozos, Triebel, Jensfelt, Rottmann, and
  Burgard]{mozos2007supervised}
Oscar~Martinez Mozos, Rudolph Triebel, Patric Jensfelt, Axel Rottmann, and
  Wolfram Burgard.
\newblock Supervised semantic labeling of places using information extracted
  from sensor data.
\newblock \emph{Robotics and Autonomous Systems}, 55\penalty0 (5):\penalty0
  391--402, 2007.
\newblock URL
  \url{https://www.sciencedirect.com/science/article/pii/S092188900600203X}.

\bibitem[Mur-Artal \& Tard{\'o}s(2017)Mur-Artal and Tard{\'o}s]{mur2017visual}
Ra{\'u}l Mur-Artal and Juan~D Tard{\'o}s.
\newblock Visual-inertial monocular {SLAM} with map reuse.
\newblock \emph{IEEE Robotics and Automation Letters}, 2\penalty0 (2):\penalty0
  796--803, 2017.
\newblock URL \url{https://ieeexplore.ieee.org/abstract/document/7817784/}.

\bibitem[Mur-Artal et~al.(2015)Mur-Artal, Montiel, and Tardos]{mur2015orb}
Raul Mur-Artal, Jose Maria~Martinez Montiel, and Juan~D Tardos.
\newblock {ORB-SLAM: A versatile and accurate monocular SLAM system}.
\newblock \emph{IEEE transactions on robotics}, 31\penalty0 (5):\penalty0
  1147--1163, 2015.
\newblock URL \url{https://ieeexplore.ieee.org/abstract/document/7219438/}.

\bibitem[Narasimhan et~al.(2020)Narasimhan, Wijmans, Chen, Darrell, Batra,
  Parikh, and Singh]{narasimhan2020seeing}
Medhini Narasimhan, Erik Wijmans, Xinlei Chen, Trevor Darrell, Dhruv Batra,
  Devi Parikh, and Amanpreet Singh.
\newblock Seeing the un-scene: {L}earning amodal semantic maps for room
  navigation.
\newblock In \emph{ECCV}, pp.\  513--529. Springer, 2020.
\newblock URL
  \url{https://link.springer.com/chapter/10.1007/978-3-030-58523-5_30}.

\bibitem[Newcombe et~al.(2011{\natexlab{a}})Newcombe, Izadi, Hilliges,
  Molyneaux, Kim, Davison, Kohi, Shotton, Hodges, and
  Fitzgibbon]{newcombe2011kinectfusion}
Richard~A Newcombe, Shahram Izadi, Otmar Hilliges, David Molyneaux, David Kim,
  Andrew~J Davison, Pushmeet Kohi, Jamie Shotton, Steve Hodges, and Andrew
  Fitzgibbon.
\newblock Kinectfusion: {R}eal-time dense surface mapping and tracking.
\newblock In \emph{IEEE international symposium on mixed and augmented
  reality}, pp.\  127--136. Ieee, 2011{\natexlab{a}}.
\newblock URL \url{https://ieeexplore.ieee.org/abstract/document/6162880/}.

\bibitem[Newcombe et~al.(2011{\natexlab{b}})Newcombe, Lovegrove, and
  Davison]{newcombe2011dtam}
Richard~A Newcombe, Steven~J Lovegrove, and Andrew~J Davison.
\newblock {DTAM: Dense tracking and mapping in real-time}.
\newblock In \emph{international conference on computer vision}, pp.\
  2320--2327. IEEE, 2011{\natexlab{b}}.
\newblock URL \url{https://ieeexplore.ieee.org/abstract/document/6126513/}.

\bibitem[Nicholson et~al.(2018)Nicholson, Milford, and
  S{\"u}nderhauf]{nicholson2018quadricslam}
Lachlan Nicholson, Michael Milford, and Niko S{\"u}nderhauf.
\newblock Quadricslam: Dual quadrics from object detections as landmarks in
  object-oriented slam.
\newblock \emph{IEEE Robotics and Automation Letters}, 4\penalty0 (1):\penalty0
  1--8, 2018.
\newblock URL \url{https://ieeexplore.ieee.org/abstract/document/8440105/}.

\bibitem[N{\"u}chter \& Hertzberg(2008)N{\"u}chter and
  Hertzberg]{nuchter2008towards}
Andreas N{\"u}chter and Joachim Hertzberg.
\newblock Towards semantic maps for mobile robots.
\newblock \emph{Robotics and Autonomous Systems}, 56\penalty0 (11):\penalty0
  915--926, 2008.
\newblock URL
  \url{https://www.sciencedirect.com/science/article/pii/S0921889008001127}.

\bibitem[N{\"u}chter et~al.(2007)N{\"u}chter, Lingemann, Hertzberg, and
  Surmann]{nuchter20076d}
Andreas N{\"u}chter, Kai Lingemann, Joachim Hertzberg, and Hartmut Surmann.
\newblock 6{D} {SLAM—3D} mapping outdoor environments.
\newblock \emph{Journal of Field Robotics}, 24\penalty0 (8-9):\penalty0
  699--722, 2007.
\newblock URL \url{https://onlinelibrary.wiley.com/doi/abs/10.1002/rob.20209}.

\bibitem[OpenAI(2023)]{openai2023gpt4}
OpenAI.
\newblock Gpt-4 technical report, 2023.
\newblock URL \url{https://arxiv.org/abs/2303.08774}.

\bibitem[Oquab et~al.(2023)Oquab, Darcet, Moutakanni, Vo, Szafraniec, Khalidov,
  Fernandez, Haziza, Massa, El-Nouby, et~al.]{oquab2023dinov2}
Maxime Oquab, Timoth{\'e}e Darcet, Th{\'e}o Moutakanni, Huy Vo, Marc
  Szafraniec, Vasil Khalidov, Pierre Fernandez, Daniel Haziza, Francisco Massa,
  Alaaeldin El-Nouby, et~al.
\newblock Dinov2: Learning robust visual features without supervision.
\newblock \emph{arXiv preprint arXiv:2304.07193}, 2023.
\newblock URL \url{https://arxiv.org/abs/2304.07193}.

\bibitem[Ortiz et~al.(2022)Ortiz, Clegg, Dong, Sucar, Novotny, Zollhoefer, and
  Mukadam]{ortiz2022isdf}
Joseph Ortiz, Alexander Clegg, Jing Dong, Edgar Sucar, David Novotny, Michael
  Zollhoefer, and Mustafa Mukadam.
\newblock {ISDF: Real-time neural signed distance fields for robot perception}.
\newblock \emph{arXiv preprint arXiv:2204.02296}, 2022.
\newblock URL \url{https://arxiv.org/abs/2204.02296}.

\bibitem[Padmakumar et~al.(2022)Padmakumar, Thomason, Shrivastava, Lange,
  Narayan-Chen, Gella, Piramuthu, and Gokhan Tur~and]{TEACh22}
Aishwarya Padmakumar, Jesse Thomason, Ayush Shrivastava, Patrick Lange, Anjali
  Narayan-Chen, Spandana Gella, Robinson Piramuthu, and Dilek Hakkani-Tur
  Gokhan Tur~and.
\newblock {TEACh: Task-driven Embodied Agents that Chat}.
\newblock In \emph{Conference on Artificial Intelligence (AAAI)}, 2022.
\newblock URL \url{https://arxiv.org/abs/2110.00534}.

\bibitem[Pan et~al.(2019)Pan, Seita, Gao, and Canny]{Pan2019RiskAR}
Xinlei Pan, Daniel Seita, Yang Gao, and John~F. Canny.
\newblock {Risk Averse Robust Adversarial Reinforcement Learning}.
\newblock \emph{International Conference on Robotics and Automation (ICRA)},
  pp.\  8522--8528, 2019.
\newblock URL \url{https://api.semanticscholar.org/CorpusID:90262410}.

\bibitem[Pathak et~al.(2017)Pathak, Agrawal, Efros, and
  Darrell]{pathak2017curiosity}
Deepak Pathak, Pulkit Agrawal, Alexei~A Efros, and Trevor Darrell.
\newblock Curiosity-driven exploration by self-supervised prediction.
\newblock In \emph{ICML}, pp.\  2778--2787, 2017.
\newblock URL \url{https://proceedings.mlr.press/v70/pathak17a.html}.

\bibitem[Patki et~al.(2019)Patki, Daniele, Walter, and
  Howard]{patki2019inferring}
Siddharth Patki, Andrea~F Daniele, Matthew~R Walter, and Thomas~M Howard.
\newblock Inferring compact representations for efficient natural language
  understanding of robot instructions.
\newblock In \emph{IEEE International Conference on Robotics and Automation},
  pp.\  6926--6933. IEEE, 2019.
\newblock URL \url{https://ieeexplore.ieee.org/abstract/document/8793667/}.

\bibitem[Patki et~al.(2020)Patki, Fahnestock, Howard, and
  Walter]{patki2020language}
Siddharth Patki, Ethan Fahnestock, Thomas~M Howard, and Matthew~R Walter.
\newblock Language-guided semantic mapping and mobile manipulation in partially
  observable environments.
\newblock In \emph{Conference on Robot Learning}, pp.\  1201--1210. PMLR, 2020.
\newblock URL \url{http://proceedings.mlr.press/v100/patki20a.html}.

\bibitem[Paul et~al.(2018)Paul, Arkin, Aksaray, Roy, and
  Howard]{paul2018efficient}
Rohan Paul, Jacob Arkin, Derya Aksaray, Nicholas Roy, and Thomas~M Howard.
\newblock Efficient grounding of abstract spatial concepts for natural language
  interaction with robot platforms.
\newblock \emph{The International Journal of Robotics Research}, 37\penalty0
  (10):\penalty0 1269--1299, 2018.
\newblock URL \url{https://dspace.mit.edu/handle/1721.1/116438}.

\bibitem[Peng et~al.(2023)Peng, Genova, Jiang, Tagliasacchi, Pollefeys, and
  Funkhouser]{Peng2023OpenScene}
Songyou Peng, Kyle Genova, Chiyu~"Max" Jiang, Andrea Tagliasacchi, Marc
  Pollefeys, and Thomas Funkhouser.
\newblock {OpenScene}: {3D} scene understanding with open vocabularies.
\newblock In \emph{Proceedings of the IEEE/CVF Conference on Computer Vision
  and Pattern Recognition}, 2023.
\newblock URL
  \url{http://openaccess.thecvf.com/content/CVPR2023/html/Peng_OpenScene_3D_Scene_Understanding_With_Open_Vocabularies_CVPR_2023_paper.html}.

\bibitem[Pfeifer \& Iida(2004)Pfeifer and Iida]{pfeifer2004embodied}
Rolf Pfeifer and Fumiya Iida.
\newblock Embodied artificial intelligence: {T}rends and challenges.
\newblock \emph{Lecture notes in computer science}, pp.\  1--26, 2004.
\newblock URL \url{https://link.springer.com/content/pdf/10.1007/b99075.pdf}.

\bibitem[Pillai \& Leonard(2015)Pillai and Leonard]{pillai2015monocular}
Sudeep Pillai and John Leonard.
\newblock Monocular slam supported object recognition.
\newblock \emph{arXiv preprint arXiv:1506.01732}, 2015.
\newblock URL \url{https://arxiv.org/abs/1506.01732}.

\bibitem[Placed \& Castellanos(2022)Placed and Castellanos]{placed2022enough}
Julio~A Placed and Jos{\'e}~A Castellanos.
\newblock Enough is enough: {T}owards autonomous uncertainty-driven stopping
  criteria.
\newblock \emph{IFAC-PapersOnLine}, 55\penalty0 (14):\penalty0 126--132, 2022.
\newblock URL
  \url{https://www.sciencedirect.com/science/article/pii/S2405896322010059}.

\bibitem[Prokhorov et~al.(2019)Prokhorov, Zhukov, Barinova, Konushin, and
  Vorontsova]{Prokhorov2019MeasuringRO}
David Prokhorov, Dmitry Zhukov, Olga Barinova, Anton Konushin, and Anna
  Vorontsova.
\newblock {Measuring robustness of Visual SLAM}.
\newblock \emph{International Conference on Machine Vision Applications (MVA)},
  pp.\  1--6, 2019.
\newblock URL \url{https://api.semanticscholar.org/CorpusID:195884278}.

\bibitem[Pronobis \& Jensfelt(2012)Pronobis and Jensfelt]{pronobis2012large}
Andrzej Pronobis and Patric Jensfelt.
\newblock Large-scale semantic mapping and reasoning with heterogeneous
  modalities.
\newblock In \emph{IEEE international conference on robotics and automation},
  pp.\  3515--3522. IEEE, 2012.
\newblock URL \url{https://ieeexplore.ieee.org/abstract/document/6224637/}.

\bibitem[Pronobis et~al.(2010)Pronobis, Martinez~Mozos, Caputo, and
  Jensfelt]{pronobis2010multi}
Andrzej Pronobis, Oscar Martinez~Mozos, Barbara Caputo, and Patric Jensfelt.
\newblock Multi-modal semantic place classification.
\newblock \emph{The International Journal of Robotics Research}, 29\penalty0
  (2-3):\penalty0 298--320, 2010.
\newblock URL
  \url{https://journals.sagepub.com/doi/abs/10.1177/0278364909356483}.

\bibitem[Pu et~al.(2023)Pu, Luo, Wang, Huang, and Liu]{pu2023visual}
Huayan Pu, Jun Luo, Gang Wang, Tao Huang, and Hongliang Liu.
\newblock Visual {SLAM} integration with semantic segmentation and deep
  learning: {A} review.
\newblock \emph{IEEE Sensors Journal}, 23\penalty0 (19):\penalty0 22119--22138,
  2023.
\newblock URL \url{https://ieeexplore.ieee.org/abstract/document/10227894/}.

\bibitem[Qi et~al.(2017{\natexlab{a}})Qi, Su, Mo, and Guibas]{qi2017pointnet}
Charles~R Qi, Hao Su, Kaichun Mo, and Leonidas~J Guibas.
\newblock {Pointnet: Deep learning on point sets for 3d classification and
  segmentation}.
\newblock In \emph{Proceedings of the IEEE conference on computer vision and
  pattern recognition}, pp.\  652--660, 2017{\natexlab{a}}.
\newblock URL
  \url{http://openaccess.thecvf.com/content_cvpr_2017/html/Qi_PointNet_Deep_Learning_CVPR_2017_paper.html}.

\bibitem[Qi et~al.(2017{\natexlab{b}})Qi, Yi, Su, and Guibas]{qi2017pointnet++}
Charles~Ruizhongtai Qi, Li~Yi, Hao Su, and Leonidas~J Guibas.
\newblock {Pointnet++: Deep hierarchical feature learning on point sets in a
  metric space}.
\newblock \emph{Advances in neural information processing systems}, 30,
  2017{\natexlab{b}}.
\newblock URL
  \url{https://proceedings.neurips.cc/paper/2017/hash/d8bf84be3800d12f74d8b05e9b89836f-Abstract.html}.

\bibitem[Qian et~al.(2021)Qian, Patath, Fu, and Xiao]{qian2021semantic}
Zhentian Qian, Kartik Patath, Jie Fu, and Jing Xiao.
\newblock Semantic slam with autonomous object-level data association.
\newblock In \emph{IEEE international conference on robotics and automation
  (ICRA)}, pp.\  11203--11209. IEEE, 2021.
\newblock URL \url{https://ieeexplore.ieee.org/abstract/document/9561532/}.

\bibitem[Qian et~al.(2022)Qian, Fu, and Xiao]{Qian2022TowardsAL}
Zhentian Qian, Jie Fu, and Jing Xiao.
\newblock {Towards Accurate Loop Closure Detection in Semantic SLAM With 3D
  Semantic Covisibility Graphs}.
\newblock \emph{IEEE Robotics and Automation Letters}, 7:\penalty0 2455--2462,
  2022.
\newblock URL \url{https://api.semanticscholar.org/CorpusID:246357678}.

\bibitem[Qin et~al.(2021)Qin, Zhang, Liu, and Lv]{qin2021semantic}
Cao Qin, Yunzhou Zhang, Yingda Liu, and Guanghao Lv.
\newblock Semantic loop closure detection based on graph matching in
  multi-objects scenes.
\newblock \emph{Journal of Visual Communication and Image Representation},
  76:\penalty0 103072, 2021.
\newblock URL
  \url{https://www.sciencedirect.com/science/article/pii/S1047320321000389}.

\bibitem[Qin et~al.(2024)Qin, Li, Zhou, Wang, and Pfister]{qin2024langsplat}
Minghan Qin, Wanhua Li, Jiawei Zhou, Haoqian Wang, and Hanspeter Pfister.
\newblock {LangSplat}: {3D} language {Gaussian} splatting.
\newblock In \emph{Proceedings of the IEEE/CVF Conference on Computer Vision
  and Pattern Recognition}, pp.\  20051--20060, 2024.
\newblock URL
  \url{http://openaccess.thecvf.com/content/CVPR2024/html/Qin_LangSplat_3D_Language_Gaussian_Splatting_CVPR_2024_paper.html}.

\bibitem[Qiu et~al.(2024)Qiu, Hu, Yang, Song, Fu, Ye, Mu, Yang, Atanasov,
  Scherer, et~al.]{qiu2024learning}
Ri-Zhao Qiu, Yafei Hu, Ge~Yang, Yuchen Song, Yang Fu, Jianglong Ye, Jiteng Mu,
  Ruihan Yang, Nikolay Atanasov, Sebastian Scherer, et~al.
\newblock Learning generalizable feature fields for mobile manipulation.
\newblock \emph{arXiv preprint arXiv:2403.07563}, 2024.
\newblock URL \url{https://arxiv.org/abs/2403.07563}.

\bibitem[Racinskis et~al.(2023)Racinskis, Arents, and
  Greitans]{racinskis2023constructing}
Peteris Racinskis, Janis Arents, and Modris Greitans.
\newblock Constructing maps for autonomous robotics: {A}n introductory
  conceptual overview.
\newblock \emph{Electronics}, 12\penalty0 (13):\penalty0 2925, 2023.
\newblock URL \url{https://www.mdpi.com/2079-9292/12/13/2925}.

\bibitem[Radford et~al.(2021)Radford, Kim, Hallacy, Ramesh, Goh, Agarwal,
  Sastry, Askell, Mishkin, Clark, et~al.]{radford2021learning}
Alec Radford, Jong~Wook Kim, Chris Hallacy, Aditya Ramesh, Gabriel Goh,
  Sandhini Agarwal, Girish Sastry, Amanda Askell, Pamela Mishkin, Jack Clark,
  et~al.
\newblock Learning transferable visual models from natural language
  supervision.
\newblock In \emph{ICML}, pp.\  8748--8763. PMLR, 2021.
\newblock URL \url{http://proceedings.mlr.press/v139/radford21a}.

\bibitem[Raibert \& Craig(1981)Raibert and Craig]{10.1115/1.3139652}
M.~H. Raibert and J.~J. Craig.
\newblock Hybrid position/force control of manipulators.
\newblock \emph{Journal of Dynamic Systems, Measurement, and Control},
  103\penalty0 (2):\penalty0 126--133, 06 1981.
\newblock ISSN 0022-0434.
\newblock \doi{10.1115/1.3139652}.
\newblock URL \url{https://doi.org/10.1115/1.3139652}.

\bibitem[Ramakrishnan et~al.(2020)Ramakrishnan, Al-Halah, and
  Grauman]{ramakrishnan2020occupancy}
Santhosh~K Ramakrishnan, Ziad Al-Halah, and Kristen Grauman.
\newblock Occupancy anticipation for efficient exploration and navigation.
\newblock In \emph{ECCV}, pp.\  400--418. Springer, 2020.
\newblock URL
  \url{https://link.springer.com/chapter/10.1007/978-3-030-58558-7_24}.

\bibitem[Ramakrishnan et~al.(2021)Ramakrishnan, Gokaslan, Wijmans, Maksymets,
  Clegg, Turner, Undersander, Galuba, Westbury, Chang, Savva, Zhao, and
  Batra]{ramakrishnan2021hm3d}
Santhosh~Kumar Ramakrishnan, Aaron Gokaslan, Erik Wijmans, Oleksandr Maksymets,
  Alexander Clegg, John~M Turner, Eric Undersander, Wojciech Galuba, Andrew
  Westbury, Angel~X Chang, Manolis Savva, Yili Zhao, and Dhruv Batra.
\newblock Habitat-matterport {3D} dataset ({HM}3d): 1000 large-scale {3D}
  environments for embodied {AI}.
\newblock In \emph{NeurIPS Datasets and Benchmarks Track (Round 2)}, 2021.
\newblock URL
  \url{https://datasets-benchmarks-proceedings.neurips.cc/paper/2021/file/34173cb38f07f89ddbebc2ac9128303f-Paper-round2.pdf}.

\bibitem[Raychaudhuri et~al.(2021)Raychaudhuri, Wani, Patel, Jain, and
  Chang]{raychaudhuri2021language}
Sonia Raychaudhuri, Saim Wani, Shivansh Patel, Unnat Jain, and Angel Chang.
\newblock Language-aligned waypoint (law) supervision for vision-and-language
  navigation in continuous environments.
\newblock In \emph{EMNLP}, pp.\  4018--4028, 2021.
\newblock URL \url{https://aclanthology.org/2021.emnlp-main.328/}.

\bibitem[Raychaudhuri et~al.(2023)Raychaudhuri, Campari, Jain, Savva, and
  Chang]{raychaudhuri2023mopa}
Sonia Raychaudhuri, Tommaso Campari, Unnat Jain, Manolis Savva, and Angel~X.
  Chang.
\newblock {MOPA: Modular Object Navigation with PointGoal Agents}.
\newblock \emph{arXiv preprint arXiv:2304.03696}, 2023.
\newblock URL \url{https://3dlg-hcvc.github.io/mopa/}.

\bibitem[Raychaudhuri et~al.(2025)Raychaudhuri, Ta, Ashton, Chang, Wang, and
  Bucher]{raychaudhuri2025zeroshotobjectcentricinstructionfollowing}
Sonia Raychaudhuri, Duy Ta, Katrina Ashton, Angel~X. Chang, Jiuguang Wang, and
  Bernadette Bucher.
\newblock Zero-shot {O}bject-{C}entric {I}nstruction {F}ollowing: {I}ntegrating
  {F}oundation {M}odels with {T}raditional {N}avigation, 2025.
\newblock URL \url{https://arxiv.org/abs/2411.07848}.

\bibitem[Ren et~al.(2024)Ren, Clark, Dixit, Itkina, Majumdar, and
  Sadigh]{ren2024explore}
Allen~Z Ren, Jaden Clark, Anushri Dixit, Masha Itkina, Anirudha Majumdar, and
  Dorsa Sadigh.
\newblock {Explore until Confident: Efficient Exploration for Embodied Question
  Answering}.
\newblock \emph{arXiv preprint arXiv:2403.15941}, 2024.
\newblock URL \url{https://arxiv.org/abs/2403.15941}.

\bibitem[Ren et~al.(2015)Ren, He, Girshick, and Sun]{ren2015faster}
Shaoqing Ren, Kaiming He, Ross Girshick, and Jian Sun.
\newblock Faster {R-CNN}: Towards real-time object detection with region
  proposal networks.
\newblock \emph{Advances in Neural Information Processing Systems}, 28, 2015.
\newblock URL
  \url{https://proceedings.neurips.cc/paper/2015/hash/14bfa6bb14875e45bba028a21ed38046-Abstract.html}.

\bibitem[Rosinol et~al.(2020{\natexlab{a}})Rosinol, Gupta, Abate, Shi, and
  Carlone]{Rosinol20rss-dynamicSceneGraphs}
A.~Rosinol, A.~Gupta, M.~Abate, J.~Shi, and L.~Carlone.
\newblock {3D Dynamic Scene Graphs: Actionable Spatial Perception with Places,
  Objects, and Humans}.
\newblock In \emph{RSS}, 2020{\natexlab{a}}.
\newblock URL \url{https://arxiv.org/abs/2002.06289}.

\bibitem[Rosinol et~al.(2020{\natexlab{b}})Rosinol, Abate, Chang, and
  Carlone]{rosinol2020kimera}
Antoni Rosinol, Marcus Abate, Yun Chang, and Luca Carlone.
\newblock Kimera: an open-source library for real-time metric-semantic
  localization and mapping.
\newblock In \emph{IEEE International Conference on Robotics and Automation
  (ICRA)}, pp.\  1689--1696. IEEE, 2020{\natexlab{b}}.
\newblock URL \url{https://ieeexplore.ieee.org/abstract/document/9196885/}.

\bibitem[Rosinol et~al.(2021)Rosinol, Violette, Abate, Hughes, Chang, Shi,
  Gupta, and Carlone]{rosinol2021kimera}
Antoni Rosinol, Andrew Violette, Marcus Abate, Nathan Hughes, Yun Chang,
  Jingnan Shi, Arjun Gupta, and Luca Carlone.
\newblock Kimera: From {SLAM} to spatial perception with {3D} dynamic scene
  graphs.
\newblock \emph{The International Journal of Robotics Research}, 40\penalty0
  (12-14):\penalty0 1510--1546, 2021.
\newblock URL
  \url{https://journals.sagepub.com/doi/abs/10.1177/02783649211056674}.

\bibitem[Rosinol et~al.(2023)Rosinol, Leonard, and Carlone]{rosinol2023nerf}
Antoni Rosinol, John~J Leonard, and Luca Carlone.
\newblock Nerf-slam: {R}eal-time dense monocular slam with neural radiance
  fields.
\newblock In \emph{IEEE/RSJ International Conference on Intelligent Robots and
  Systems (IROS)}, pp.\  3437--3444. IEEE, 2023.
\newblock URL \url{https://ieeexplore.ieee.org/abstract/document/10341922/}.

\bibitem[Rublee et~al.(2011)Rublee, Rabaud, Konolige, and
  Bradski]{rublee2011orb}
Ethan Rublee, Vincent Rabaud, Kurt Konolige, and Gary Bradski.
\newblock {ORB}: An efficient alternative to {SIFT} or {SURF}.
\newblock In \emph{Proceedings of the IEEE/CVF International Conference on
  Computer Vision}, pp.\  2564--2571. IEEE, 2011.
\newblock URL \url{https://ieeexplore.ieee.org/abstract/document/6126544/}.

\bibitem[R{\"u}nz \& Agapito(2017)R{\"u}nz and Agapito]{runz2017co}
Martin R{\"u}nz and Lourdes Agapito.
\newblock {Co-fusion: Real-time segmentation, tracking and fusion of multiple
  objects}.
\newblock In \emph{IEEE International Conference on Robotics and Automation
  (ICRA)}, pp.\  4471--4478. IEEE, 2017.
\newblock URL \url{https://ieeexplore.ieee.org/abstract/document/7989518/}.

\bibitem[Salas-Moreno et~al.(2013)Salas-Moreno, Newcombe, Strasdat, Kelly, and
  Davison]{salas2013slam++}
Renato~F Salas-Moreno, Richard~A Newcombe, Hauke Strasdat, Paul~HJ Kelly, and
  Andrew~J Davison.
\newblock Slam++: {S}imultaneous localisation and mapping at the level of
  objects.
\newblock In \emph{Proceedings of the IEEE/CVF Conference on Computer Vision
  and Pattern Recognition}, pp.\  1352--1359, 2013.
\newblock URL
  \url{http://openaccess.thecvf.com/content_cvpr_2013/html/Salas-Moreno_SLAM_Simultaneous_Localisation_2013_CVPR_paper.html}.

\bibitem[Saputra et~al.(2018)Saputra, Markham, and Trigoni]{saputra2018visual}
Muhamad Risqi~U Saputra, Andrew Markham, and Niki Trigoni.
\newblock {Visual SLAM and structure from motion in dynamic environments: A
  survey}.
\newblock \emph{ACM Computing Surveys (CSUR)}, 51\penalty0 (2):\penalty0 1--36,
  2018.
\newblock URL \url{https://dl.acm.org/doi/abs/10.1145/3177853}.

\bibitem[Sarlin et~al.(2020)Sarlin, DeTone, Malisiewicz, and
  Rabinovich]{sarlin2020superglue}
Paul-Edouard Sarlin, Daniel DeTone, Tomasz Malisiewicz, and Andrew Rabinovich.
\newblock Superglue: {L}earning feature matching with graph neural networks.
\newblock In \emph{Proceedings of the IEEE/CVF Conference on Computer Vision
  and Pattern Recognition}, pp.\  4938--4947, 2020.
\newblock URL
  \url{http://openaccess.thecvf.com/content_CVPR_2020/html/Sarlin_SuperGlue_Learning_Feature_Matching_With_Graph_Neural_Networks_CVPR_2020_paper.html}.

\bibitem[Savinov et~al.(2018)Savinov, Dosovitskiy, and Koltun]{savinov2018semi}
Nikolay Savinov, Alexey Dosovitskiy, and Vladlen Koltun.
\newblock Semi-parametric topological memory for navigation.
\newblock In \emph{International Conference on Learning Representations}, 2018.
\newblock URL \url{https://arxiv.org/abs/1803.00653}.

\bibitem[Savva et~al.(2019)Savva, Kadian, Maksymets, Zhao, Wijmans, Jain,
  Straub, Liu, Koltun, Malik, et~al.]{savva2019habitat}
Manolis Savva, Abhishek Kadian, Oleksandr Maksymets, Yili Zhao, Erik Wijmans,
  Bhavana Jain, Julian Straub, Jia Liu, Vladlen Koltun, Jitendra Malik, et~al.
\newblock Habitat: A platform for embodied {AI} research.
\newblock In \emph{Proceedings of the IEEE/CVF International Conference on
  Computer Vision}, pp.\  9339--9347, 2019.
\newblock URL
  \url{http://openaccess.thecvf.com/content\_ICCV\_2019/html/Savva\_Habitat\_A\_Platform\_for\_Embodied\_AI\_Research\_ICCV\_2019\_paper.html}.

\bibitem[Schops et~al.(2019)Schops, Sattler, and Pollefeys]{schops2019bad}
Thomas Schops, Torsten Sattler, and Marc Pollefeys.
\newblock {Bad slam: Bundle adjusted direct rgb-d slam}.
\newblock In \emph{Proceedings of the IEEE/CVF Conference on Computer Vision
  and Pattern Recognition}, pp.\  134--144, 2019.
\newblock URL
  \url{http://openaccess.thecvf.com/content_CVPR_2019/html/Schops_BAD_SLAM_Bundle_Adjusted_Direct_RGB-D_SLAM_CVPR_2019_paper.html}.

\bibitem[Sengupta \& Sturgess(2015)Sengupta and Sturgess]{sengupta2015semantic}
Sunando Sengupta and Paul Sturgess.
\newblock Semantic octree: {U}nifying recognition, reconstruction and
  representation via an octree constrained higher order mrf.
\newblock In \emph{IEEE international conference on robotics and automation
  (ICRA)}, pp.\  1874--1879. IEEE, 2015.
\newblock URL \url{https://ieeexplore.ieee.org/abstract/document/7139442/}.

\bibitem[Sethian(1996)]{sethian1996fast}
James~A Sethian.
\newblock Fast-marching level-set methods for three-dimensional
  photolithography development.
\newblock In \emph{Optical Microlithography IX}, volume 2726, pp.\  262--272.
  International Society for Optics and Photonics, 1996.
\newblock URL
  \url{https://www.spiedigitallibrary.org/conference-proceedings-of-spie/2726/0000/Fast-marching-level-set-methods-for-three-dimensional-photolithography-development/10.1117/12.240962.short}.

\bibitem[Shafiullah et~al.(2023)Shafiullah, Paxton, Pinto, Chintala, and
  Szlam]{shafiullah2023clip}
Nur Muhammad~Mahi Shafiullah, Chris Paxton, Lerrel Pinto, Soumith Chintala, and
  Arthur Szlam.
\newblock {CLIP}-{F}ields: Weakly supervised semantic fields for robotic
  memory.
\newblock In \emph{IEEE International Conference on Robotics and Automation},
  2023.
\newblock URL \url{https://arxiv.org/abs/2210.05663}.

\bibitem[Shah et~al.(2023)Shah, Osinski, Ichter, and Levine]{shah2022lmnav}
Dhruv Shah, Blazej Osinski, Brian Ichter, and Sergey Levine.
\newblock {LM-Nav}: Robotic navigation with large pre-trained models of
  language, vision, and action.
\newblock In \emph{Conference on Robot Learning}, pp.\  492--504. PMLR, 2023.
\newblock URL \url{https://openreview.net/forum?id=UW5A3SweAH}.

\bibitem[Shan et~al.(2020{\natexlab{a}})Shan, Feng, and
  Atanasov]{shan2020orcvio}
Mo~Shan, Qiaojun Feng, and Nikolay Atanasov.
\newblock {OrcVIO: Object residual constrained visual-inertial odometry}.
\newblock In \emph{IEEE/RSJ International Conference on Intelligent Robots and
  Systems (IROS)}, pp.\  5104--5111. IEEE, 2020{\natexlab{a}}.
\newblock URL \url{https://ieeexplore.ieee.org/abstract/document/9341660/}.

\bibitem[Shan et~al.(2020{\natexlab{b}})Shan, Englot, Meyers, Wang, Ratti, and
  Rus]{shan2020lio}
Tixiao Shan, Brendan Englot, Drew Meyers, Wei Wang, Carlo Ratti, and Daniela
  Rus.
\newblock Lio-sam: {T}ightly-coupled lidar inertial odometry via smoothing and
  mapping.
\newblock In \emph{IEEE/RSJ international conference on intelligent robots and
  systems (IROS)}, pp.\  5135--5142. IEEE, 2020{\natexlab{b}}.
\newblock URL \url{https://ieeexplore.ieee.org/abstract/document/9341176/}.

\bibitem[Shen et~al.(2023)Shen, Yang, Yu, Wong, Kaelbling, and
  Isola]{shen2023distilled}
William Shen, Ge~Yang, Alan Yu, Jansen Wong, Leslie~Pack Kaelbling, and Phillip
  Isola.
\newblock Distilled feature fields enable few-shot language-guided
  manipulation.
\newblock \emph{arXiv preprint arXiv:2308.07931}, 2023.
\newblock URL \url{https://arxiv.org/abs/2308.07931}.

\bibitem[Shen et~al.(2019)Shen, Xu, Zhu, Guibas, Fei-Fei, and
  Savarese]{shen2019situational}
William~B Shen, Danfei Xu, Yuke Zhu, Leonidas~J Guibas, Li~Fei-Fei, and Silvio
  Savarese.
\newblock Situational fusion of visual representation for visual navigation.
\newblock In \emph{Proceedings of the IEEE/CVF International Conference on
  Computer Vision}, pp.\  2881--2890, 2019.
\newblock URL
  \url{http://openaccess.thecvf.com/content\_ICCV\_2019/html/Shen\_Situational\_Fusion\_of\_Visual\_Representation\_for\_Visual\_Navigation\_ICCV\_2019\_paper.html}.

\bibitem[Shridhar et~al.(2020)Shridhar, Thomason, Gordon, Bisk, Han, Mottaghi,
  Zettlemoyer, and Fox]{ALFRED20}
Mohit Shridhar, Jesse Thomason, Daniel Gordon, Yonatan Bisk, Winson Han,
  Roozbeh Mottaghi, Luke Zettlemoyer, and Dieter Fox.
\newblock {ALFRED: A Benchmark for Interpreting Grounded Instructions for
  Everyday Tasks}.
\newblock In \emph{The IEEE Conference on Computer Vision and Pattern
  Recognition (CVPR)}, 2020.
\newblock URL \url{https://arxiv.org/abs/1912.01734}.

\bibitem[Singh et~al.(2023)Singh, Wijegunawardana, Samarakoon, Muthugala, and
  Elara]{singh2023vision}
Ishneet~Sukhvinder Singh, ID~Wijegunawardana, SM~Bhagya~P Samarakoon,
  MA~Viraj~J Muthugala, and Mohan~Rajesh Elara.
\newblock Vision-based dirt distribution mapping using deep learning.
\newblock \emph{Scientific Reports}, 13\penalty0 (1):\penalty0 12741, 2023.
\newblock URL \url{https://www.nature.com/articles/s41598-023-38538-3}.

\bibitem[Song et~al.(2025)Song, Liang, and Huaidong]{song2025semantic}
Xu~Song, Xuan Liang, and Zhou Huaidong.
\newblock {Semantic mapping techniques for indoor mobile robots: Review and
  prospect}.
\newblock \emph{Measurement and Control}, 58\penalty0 (3):\penalty0 377--393,
  2025.
\newblock URL
  \url{https://journals.sagepub.com/doi/abs/10.1177/00202940241259903}.

\bibitem[Sousa et~al.(2023)Sousa, Sobreira, and Moreira]{sousa2023systematic}
Ricardo~B Sousa, H{\'e}ber~M Sobreira, and Ant{\'o}nio~Paulo Moreira.
\newblock A systematic literature review on long-term localization and mapping
  for mobile robots.
\newblock \emph{Journal of Field Robotics}, 40\penalty0 (5):\penalty0
  1245--1322, 2023.
\newblock URL \url{https://onlinelibrary.wiley.com/doi/abs/10.1002/rob.22170}.

\bibitem[Stachniss et~al.(2005)Stachniss, Grisetti, and
  Burgard]{Stachniss2005InformationGE}
C.~Stachniss, Giorgio Grisetti, and Wolfram Burgard.
\newblock {Information Gain-based Exploration Using Rao-Blackwellized Particle
  Filters}.
\newblock In \emph{Robotics: Science and Systems}, 2005.
\newblock URL \url{https://api.semanticscholar.org/CorpusID:8255559}.

\bibitem[St{\"u}ckler \& Behnke(2014)St{\"u}ckler and
  Behnke]{stuckler2014multi}
J{\"o}rg St{\"u}ckler and Sven Behnke.
\newblock Multi-resolution surfel maps for efficient dense 3{D} modeling and
  tracking.
\newblock \emph{Journal of Visual Communication and Image Representation},
  25\penalty0 (1):\penalty0 137--147, 2014.
\newblock URL
  \url{https://www.sciencedirect.com/science/article/pii/S104732031300028X}.

\bibitem[St{\"u}ckler et~al.(2012)St{\"u}ckler, Biresev, and
  Behnke]{stuckler2012semantic}
J{\"o}rg St{\"u}ckler, Nenad Biresev, and Sven Behnke.
\newblock Semantic mapping using object-class segmentation of {RGB-D} images.
\newblock In \emph{IEEE/RSJ International Conference on Intelligent Robots and
  Systems}, pp.\  3005--3010. IEEE, 2012.
\newblock URL \url{https://ieeexplore.ieee.org/abstract/document/6385983/}.

\bibitem[Sualeh \& Kim(2019)Sualeh and Kim]{sualeh2019simultaneous}
Muhammad Sualeh and Gon-Woo Kim.
\newblock Simultaneous localization and mapping in the epoch of semantics: a
  survey.
\newblock \emph{International Journal of Control, Automation and Systems},
  17\penalty0 (3):\penalty0 729--742, 2019.
\newblock URL
  \url{https://link.springer.com/article/10.1007/s12555-018-0130-x}.

\bibitem[Sucar et~al.(2021)Sucar, Liu, Ortiz, and Davison]{sucar2021imap}
Edgar Sucar, Shikun Liu, Joseph Ortiz, and Andrew~J Davison.
\newblock {imap: Implicit mapping and positioning in real-time}.
\newblock In \emph{Proceedings of the IEEE/CVF international conference on
  computer vision}, pp.\  6229--6238, 2021.
\newblock URL
  \url{http://openaccess.thecvf.com/content/ICCV2021/html/Sucar_iMAP_Implicit_Mapping_and_Positioning_in_Real-Time_ICCV_2021_paper.html}.

\bibitem[Szot et~al.(2021)Szot, Clegg, Undersander, Wijmans, Zhao, Turner,
  Maestre, Mukadam, Chaplot, Maksymets, et~al.]{szot2021habitat}
Andrew Szot, Alex Clegg, Eric Undersander, Erik Wijmans, Yili Zhao, John
  Turner, Noah Maestre, Mustafa Mukadam, Devendra Chaplot, Oleksandr Maksymets,
  et~al.
\newblock Habitat 2.0: Training home assistants to rearrange their habitat.
\newblock \emph{Advances in Neural Information Processing Systems}, 2021.
\newblock URL
  \url{https://proceedings.neurips.cc/paper/2021/hash/021bbc7ee20b71134d53e20206bd6feb-Abstract.html}.

\bibitem[Taheri \& Xia(2021)Taheri and Xia]{taheri2021slam}
Hamid Taheri and Zhao~Chun Xia.
\newblock {SLAM}; definition and evolution.
\newblock \emph{Engineering Applications of Artificial Intelligence},
  97:\penalty0 104032, 2021.
\newblock URL
  \url{https://www.sciencedirect.com/science/article/pii/S0952197620303092}.

\bibitem[Tai et~al.(2017)Tai, Paolo, and Liu]{tai2017virtual}
Lei Tai, Giuseppe Paolo, and Ming Liu.
\newblock Virtual-to-real deep reinforcement learning: {C}ontinuous control of
  mobile robots for mapless navigation.
\newblock In \emph{IEEE/RSJ international conference on intelligent robots and
  systems (IROS)}, pp.\  31--36. IEEE, 2017.
\newblock URL \url{https://ieeexplore.ieee.org/abstract/document/8202134/}.

\bibitem[Taioli et~al.(2023)Taioli, Cunico, Girella, Bologna, Farinelli, and
  Cristani]{taioli2023language}
Francesco Taioli, Federico Cunico, Federico Girella, Riccardo Bologna,
  Alessandro Farinelli, and Marco Cristani.
\newblock Language-enhanced {RNR-map}: Querying renderable neural radiance
  field maps with natural language.
\newblock In \emph{Proceedings of the IEEE/CVF International Conference on
  Computer Vision}, pp.\  4669--4674, 2023.
\newblock URL
  \url{https://openaccess.thecvf.com/content/ICCV2023W/VLAR/html/Taioli_Language-Enhanced_RNR-Map_Querying_Renderable_Neural_Radiance_Field_Maps_with_Natural_ICCVW_2023_paper.html}.

\bibitem[Taioli et~al.(2024)Taioli, Zorzi, Franchi, Castellini, Farinelli,
  Cristani, and Wang]{taioli2024collaborative}
Francesco Taioli, Edoardo Zorzi, Gianni Franchi, Alberto Castellini, Alessandro
  Farinelli, Marco Cristani, and Yiming Wang.
\newblock Collaborative {I}nstance {N}avigation: {L}everaging {A}gent
  {S}elf-{D}ialogue to {M}inimize {U}ser {I}nput.
\newblock \emph{arXiv preprint arXiv:2412.01250}, 2024.
\newblock URL \url{https://ui.adsabs.harvard.edu/abs/2024arXiv241201250T/}.

\bibitem[Taketomi et~al.(2017)Taketomi, Uchiyama, and
  Ikeda]{taketomi2017visual}
Takafumi Taketomi, Hideaki Uchiyama, and Sei Ikeda.
\newblock {Visual SLAM algorithms: A survey from 2010 to 2016}.
\newblock \emph{IPSJ transactions on computer vision and applications},
  9\penalty0 (1):\penalty0 16, 2017.
\newblock URL
  \url{https://link.springer.com/article/10.1186/s41074-017-0027-2}.

\bibitem[Talha~Bukhari et~al.(2025)Talha~Bukhari, Lawson, and
  Qureshi]{talha2025differentiable}
S~Talha~Bukhari, Daniel Lawson, and Ahmed~H Qureshi.
\newblock {Differentiable Composite Neural Signed Distance Fields for Robot
  Navigation in Dynamic Indoor Environments}.
\newblock \emph{arXiv e-prints}, pp.\  arXiv--2502, 2025.
\newblock URL \url{https://arxiv.org/abs/2502.02664}.

\bibitem[Tang et~al.(2025)Tang, Wang, Deng, Zheng, Deng, and
  Yue]{tang2025openin}
Yujie Tang, Meiling Wang, Yinan Deng, Zibo Zheng, Jingchuan Deng, and Yufeng
  Yue.
\newblock Open{IN}: {O}pen-{V}ocabulary {I}nstance-{O}riented {N}avigation in
  {D}ynamic {D}omestic {E}nvironments.
\newblock \emph{arXiv preprint arXiv:2501.04279}, 2025.
\newblock URL
  \url{https://ieeexplore.ieee.org/iel8/7083369/7339444/11091457.pdf}.

\bibitem[Tateno et~al.(2017)Tateno, Tombari, Laina, and Navab]{tateno2017cnn}
Keisuke Tateno, Federico Tombari, Iro Laina, and Nassir Navab.
\newblock Cnn-slam: {R}eal-time dense monocular slam with learned depth
  prediction.
\newblock In \emph{Proceedings of the IEEE conference on computer vision and
  pattern recognition}, pp.\  6243--6252, 2017.
\newblock URL
  \url{http://openaccess.thecvf.com/content_cvpr_2017/html/Tateno_CNN-SLAM_Real-Time_Dense_CVPR_2017_paper.html}.

\bibitem[Teed \& Deng(2021)Teed and Deng]{teed2021droid}
Zachary Teed and Jia Deng.
\newblock Droid-slam: {D}eep visual slam for monocular, stereo, and rgb-d
  cameras.
\newblock \emph{Advances in neural information processing systems},
  34:\penalty0 16558--16569, 2021.
\newblock URL
  \url{https://proceedings.neurips.cc/paper/2021/hash/89fcd07f20b6785b92134bd6c1d0fa42-Abstract.html}.

\bibitem[Tellex et~al.(2011{\natexlab{a}})Tellex, Kollar, Dickerson, Walter,
  Banerjee, Teller, and Roy]{tellex2011understanding}
Stefanie Tellex, Thomas Kollar, Steven Dickerson, Matthew Walter, Ashis
  Banerjee, Seth Teller, and Nicholas Roy.
\newblock Understanding natural language commands for robotic navigation and
  mobile manipulation.
\newblock In \emph{Proceedings of the AAAI conference on artificial
  intelligence}, volume~25, pp.\  1507--1514, 2011{\natexlab{a}}.
\newblock URL \url{https://ojs.aaai.org/index.php/AAAI/article/view/7979}.

\bibitem[Tellex et~al.(2011{\natexlab{b}})Tellex, Kollar, Dickerson, Walter,
  Banerjee, Teller, and Roy]{Tellex2011UnderstandingNL}
Stefanie Tellex, Thomas Kollar, Steven Dickerson, Matthew~R. Walter,
  Ashis~Gopal Banerjee, Seth~J. Teller, and Nicholas Roy.
\newblock {Understanding Natural Language Commands for Robotic Navigation and
  Mobile Manipulation}.
\newblock In \emph{AAAI Conference on Artificial Intelligence},
  2011{\natexlab{b}}.
\newblock URL \url{https://api.semanticscholar.org/CorpusID:220828823}.

\bibitem[Thomas et~al.(2019)Thomas, Qi, Deschaud, Marcotegui, Goulette, and
  Guibas]{thomas2019kpconv}
Hugues Thomas, Charles~R Qi, Jean-Emmanuel Deschaud, Beatriz Marcotegui,
  Fran{\c{c}}ois Goulette, and Leonidas~J Guibas.
\newblock {Kpconv: Flexible and deformable convolution for point clouds}.
\newblock In \emph{Proceedings of the IEEE/CVF international conference on
  computer vision}, pp.\  6411--6420, 2019.
\newblock URL
  \url{http://openaccess.thecvf.com/content_ICCV_2019/html/Thomas_KPConv_Flexible_and_Deformable_Convolution_for_Point_Clouds_ICCV_2019_paper.html}.

\bibitem[Thrun(2002)]{thrun2002probabilistic}
Sebastian Thrun.
\newblock Probabilistic robotics.
\newblock \emph{Communications of the ACM}, 45\penalty0 (3):\penalty0 52--57,
  2002.
\newblock URL \url{https://dl.acm.org/doi/fullHtml/10.1145/504729.504754}.

\bibitem[Thrun(2003)]{Thrun2003RoboticMA}
Sebastian Thrun.
\newblock Robotic mapping: a survey.
\newblock In \emph{Robotic mapping: a survey}, 2003.
\newblock URL \url{https://api.semanticscholar.org/CorpusID:14633188}.

\bibitem[Thrun \& Montemerlo(2006)Thrun and Montemerlo]{thrun2006graph}
Sebastian Thrun and Michael Montemerlo.
\newblock The graph {SLAM} algorithm with applications to large-scale mapping
  of urban structures.
\newblock \emph{The International Journal of Robotics Research}, 25\penalty0
  (5-6):\penalty0 403--429, 2006.
\newblock URL
  \url{https://journals.sagepub.com/doi/abs/10.1177/0278364906065387}.

\bibitem[Thrun et~al.(1998{\natexlab{a}})Thrun, Burgard, and
  Fox]{thrun1998probabilistic}
Sebastian Thrun, Wolfram Burgard, and Dieter Fox.
\newblock A probabilistic approach to concurrent mapping and localization for
  mobile robots.
\newblock \emph{Autonomous Robots}, 5:\penalty0 253--271, 1998{\natexlab{a}}.
\newblock URL \url{https://link.springer.com/article/10.1023/A:1008806205438}.

\bibitem[Thrun et~al.(1998{\natexlab{b}})Thrun, Gutmann, Fox, Burgard, Kuipers,
  et~al.]{thrun1998integrating}
Sebastian Thrun, Jens-Steffen Gutmann, Dieter Fox, Wolfram Burgard, Benjamin
  Kuipers, et~al.
\newblock Integrating topological and metric maps for mobile robot navigation:
  A statistical approach.
\newblock \emph{AAAI/IAAI}, 9:\penalty0 989--995, 1998{\natexlab{b}}.
\newblock URL \url{https://cdn.aaai.org/AAAI/1998/AAAI98-140.pdf}.

\bibitem[Thrun et~al.(2000)Thrun, Burgard, and Fox]{thrun2000real}
Sebastian Thrun, Wolfram Burgard, and Dieter Fox.
\newblock A real-time algorithm for mobile robot mapping with applications to
  multi-robot and 3{D} mapping.
\newblock In \emph{Proceedings ICRA. Millennium Conference. IEEE International
  Conference on Robotics and Automation. Symposia Proceedings (Cat. No.
  00CH37065)}, volume~1, pp.\  321--328. IEEE, 2000.
\newblock URL \url{https://ieeexplore.ieee.org/abstract/document/844077/}.

\bibitem[Thrun et~al.(2001)Thrun, Fox, Burgard, and
  Dellaert]{Thrun2001RobustMC}
Sebastian Thrun, Dieter Fox, Wolfram Burgard, and Frank Dellaert.
\newblock {Robust Monte Carlo localization for mobile robots}.
\newblock \emph{Artif. Intell.}, 128:\penalty0 99--141, 2001.
\newblock URL \url{https://api.semanticscholar.org/CorpusID:128871}.

\bibitem[Tomatis et~al.(2001)Tomatis, Nourbakhsh, and
  Siegwart]{tomatis2001combining}
Nicola Tomatis, Illah Nourbakhsh, and Roland Siegwart.
\newblock Combining topological and metric: A natural integration for
  simultaneous localization and map building.
\newblock In \emph{Proceedings of the Fourth European Workshop on Advanced
  Mobile Robots (Eurobot)}. ETH-Z{\"u}rich, 2001.
\newblock URL
  \url{https://www.research-collection.ethz.ch/bitstream/handle/20.500.11850/82590/eth-8357-01.pdf}.

\bibitem[Tourani et~al.(2022)Tourani, Bavle, Sanchez-Lopez, and
  Voos]{tourani2022visual}
Ali Tourani, Hriday Bavle, Jose~Luis Sanchez-Lopez, and Holger Voos.
\newblock Visual slam: {W}hat are the current trends and what to expect?
\newblock \emph{Sensors}, 22\penalty0 (23):\penalty0 9297, 2022.
\newblock URL \url{https://www.mdpi.com/1424-8220/22/23/9297}.

\bibitem[Trabucco et~al.(2022)Trabucco, Sigurdsson, Piramuthu, Sukhatme, and
  Salakhutdinov]{trabucco2022simple}
Brandon Trabucco, Gunnar~A Sigurdsson, Robinson Piramuthu, Gaurav~S Sukhatme,
  and Ruslan Salakhutdinov.
\newblock A simple approach for visual room rearrangement: {3D} mapping and
  semantic search.
\newblock In \emph{The International Conference on Learning Representations},
  2022.
\newblock URL \url{https://openreview.net/forum?id=1C6nCCaRe6p}.

\bibitem[Triebel et~al.(2012)Triebel, Paul, Rus, and
  Newman]{triebel2012parsing}
Rudolph Triebel, Rohan Paul, Daniela Rus, and Paul Newman.
\newblock Parsing outdoor scenes from streamed 3d laser data using online
  clustering and incremental belief updates.
\newblock In \emph{AAAI}, volume~26, pp.\  2088--2095, 2012.
\newblock URL \url{https://ojs.aaai.org/index.php/AAAI/article/view/8378}.

\bibitem[Tsintotas et~al.(2022)Tsintotas, Bampis, and
  Gasteratos]{tsintotas2022revisiting}
Konstantinos~A Tsintotas, Loukas Bampis, and Antonios Gasteratos.
\newblock The revisiting problem in simultaneous localization and mapping: {A}
  survey on visual loop closure detection.
\newblock \emph{IEEE Transactions on Intelligent Transportation Systems},
  23\penalty0 (11):\penalty0 19929--19953, 2022.
\newblock URL \url{https://ieeexplore.ieee.org/abstract/document/9780121/}.

\bibitem[Tucker et~al.(2019)Tucker, Aksaray, Paul, Stein, and
  Roy]{tucker2019learning}
Mycal Tucker, Derya Aksaray, Rohan Paul, Gregory~J Stein, and Nicholas Roy.
\newblock Learning unknown groundings for natural language interaction with
  mobile robots.
\newblock In \emph{Robotics Research: The International Symposium ISRR}, pp.\
  317--333. Springer, 2019.
\newblock URL
  \url{https://link.springer.com/chapter/10.1007/978-3-030-28619-4_27}.

\bibitem[Valentin et~al.(2013)Valentin, Sengupta, Warrell, Shahrokni, and
  Torr]{valentin2013mesh}
Julien~PC Valentin, Sunando Sengupta, Jonathan Warrell, Ali Shahrokni, and
  Philip~HS Torr.
\newblock Mesh based semantic modelling for indoor and outdoor scenes.
\newblock In \emph{Proceedings of the IEEE/CVF Conference on Computer Vision
  and Pattern Recognition}, pp.\  2067--2074, 2013.
\newblock URL
  \url{http://openaccess.thecvf.com/content_cvpr_2013/html/Valentin_Mesh_Based_Semantic_2013_CVPR_paper.html}.

\bibitem[Vasu et~al.(2024)Vasu, Pouransari, Faghri, Vemulapalli, and
  Tuzel]{vasu2024mobileclip}
Pavan Kumar~Anasosalu Vasu, Hadi Pouransari, Fartash Faghri, Raviteja
  Vemulapalli, and Oncel Tuzel.
\newblock {Mobileclip: Fast image-text models through multi-modal reinforced
  training}.
\newblock In \emph{Proceedings of the IEEE/CVF Conference on Computer Vision
  and Pattern Recognition}, pp.\  15963--15974, 2024.
\newblock URL
  \url{http://openaccess.thecvf.com/content/CVPR2024/html/Vasu_MobileCLIP_Fast_Image-Text_Models_through_Multi-Modal_Reinforced_Training_CVPR_2024_paper.html}.

\bibitem[Vasudevan \& Siegwart(2008)Vasudevan and
  Siegwart]{vasudevan2008bayesian}
Shrihari Vasudevan and Roland Siegwart.
\newblock Bayesian space conceptualization and place classification for
  semantic maps in mobile robotics.
\newblock \emph{Robotics and Autonomous Systems}, 56\penalty0 (6):\penalty0
  522--537, 2008.
\newblock URL
  \url{https://www.sciencedirect.com/science/article/pii/S092188900800033X}.

\bibitem[Vineet et~al.(2015)Vineet, Miksik, Lidegaard, Nie{\ss}ner, Golodetz,
  Prisacariu, K{\"a}hler, Murray, Izadi, P{\'e}rez,
  et~al.]{vineet2015incremental}
Vibhav Vineet, Ondrej Miksik, Morten Lidegaard, Matthias Nie{\ss}ner, Stuart
  Golodetz, Victor~A Prisacariu, Olaf K{\"a}hler, David~W Murray, Shahram
  Izadi, Patrick P{\'e}rez, et~al.
\newblock Incremental dense semantic stereo fusion for large-scale semantic
  scene reconstruction.
\newblock In \emph{IEEE international conference on robotics and automation
  (ICRA)}, pp.\  75--82. IEEE, 2015.
\newblock URL \url{https://ieeexplore.ieee.org/abstract/document/7138983/}.

\bibitem[Vora et~al.(2021)Vora, Radwan, Greff, Meyer, Genova, Sajjadi, Pot,
  Tagliasacchi, and Duckworth]{vora2021nesf}
Suhani Vora, Noha Radwan, Klaus Greff, Henning Meyer, Kyle Genova, Mehdi~SM
  Sajjadi, Etienne Pot, Andrea Tagliasacchi, and Daniel Duckworth.
\newblock {NeSF}: Neural semantic fields for generalizable semantic
  segmentation of {3D} scenes.
\newblock \emph{arXiv preprint arXiv:2111.13260}, 2021.
\newblock URL \url{https://arxiv.org/abs/2111.13260}.

\bibitem[Walter et~al.(2013)Walter, Hemachandra, Homberg, Tellex, and
  Teller]{walter2013learning}
Matthew~R Walter, Sachithra Hemachandra, Bianca Homberg, Stefanie Tellex, and
  Seth~J Teller.
\newblock Learning {S}emantic {M}aps from {N}atural {L}anguage {D}escriptions.
\newblock In \emph{Robotics: science and systems}, volume~2, 2013.
\newblock URL \url{https://dspace.mit.edu/handle/1721.1/87051}.

\bibitem[Walter et~al.(2014)Walter, Hemachandra, Homberg, Tellex, and
  Teller]{walter2014framework}
Matthew~R Walter, Sachithra Hemachandra, Bianca Homberg, Stefanie Tellex, and
  Seth Teller.
\newblock A framework for learning semantic maps from grounded natural language
  descriptions.
\newblock \emph{The International Journal of Robotics Research}, 33\penalty0
  (9):\penalty0 1167--1190, 2014.
\newblock URL
  \url{https://journals.sagepub.com/doi/abs/10.1177/0278364914537359}.

\bibitem[Walter et~al.(2022)Walter, Patki, Daniele, Fahnestock, Duvallet,
  Hemachandra, Oh, Stentz, Roy, and Howard]{10878387}
Matthew~R. Walter, Siddharth Patki, Andrea~F. Daniele, Ethan Fahnestock, Felix
  Duvallet, Sachithra Hemachandra, Jean Oh, Anthony Stentz, Nicholas Roy, and
  Thomas~M. Howard.
\newblock Language {U}nderstanding for {F}ield and {S}ervice {R}obots in a
  {P}riori {U}nknown {E}nvironments.
\newblock \emph{Field Robotics}, 2:\penalty0 1191--1231, 2022.
\newblock \doi{10.55417/fr.2022040}.
\newblock URL \url{https://ieeexplore.ieee.org/abstract/document/10878387/}.

\bibitem[Wang et~al.(2022)Wang, Chai, He, Chen, and Liao]{wang2022clip}
Can Wang, Menglei Chai, Mingming He, Dongdong Chen, and Jing Liao.
\newblock {Clip-nerf: Text-and-image driven manipulation of neural radiance
  fields}.
\newblock In \emph{Proceedings of the IEEE/CVF conference on computer vision
  and pattern recognition}, pp.\  3835--3844, 2022.
\newblock URL
  \url{http://openaccess.thecvf.com/content/CVPR2022/html/Wang_CLIP-NeRF_Text-and-Image_Driven_Manipulation_of_Neural_Radiance_Fields_CVPR_2022_paper.html}.

\bibitem[Wang et~al.(2024)Wang, Tian, Chen, Xu, and Ding]{wang2024survey}
Yanan Wang, Yaobin Tian, Jiawei Chen, Kun Xu, and Xilun Ding.
\newblock A survey of visual {SLAM} in dynamic environment: {T}he evolution
  from geometric to semantic approaches.
\newblock \emph{IEEE Transactions on Instrumentation and Measurement}, 2024.
\newblock URL \url{https://ieeexplore.ieee.org/abstract/document/10577209/}.

\bibitem[Wang et~al.(2025)Wang, Li, Wu, Moens, and
  Tuytelaars]{wang2025instruction}
Zehao Wang, Mingxiao Li, Minye Wu, Marie-Francine Moens, and Tinne Tuytelaars.
\newblock Instruction-guided path planning with 3{D} semantic maps for
  vision-language navigation.
\newblock \emph{Neurocomputing}, pp.\  129457, 2025.
\newblock URL
  \url{https://www.sciencedirect.com/science/article/pii/S0925231225001298}.

\bibitem[Wani et~al.(2020)Wani, Patel, Jain, Chang, and Savva]{wani2020multion}
Saim Wani, Shivansh Patel, Unnat Jain, Angel Chang, and Manolis Savva.
\newblock Multi{ON}: Benchmarking semantic map memory using multi-object
  navigation.
\newblock \emph{Advances in Neural Information Processing Systems},
  33:\penalty0 9700--9712, 2020.
\newblock URL
  \url{https://proceedings.neurips.cc/paper/2020/hash/6e01383fd96a17ae51cc3e15447e7533-Abstract.html}.

\bibitem[Weihs et~al.(2021)Weihs, Deitke, Kembhavi, and
  Mottaghi]{weihs2021visual}
Luca Weihs, Matt Deitke, Aniruddha Kembhavi, and Roozbeh Mottaghi.
\newblock Visual room rearrangement.
\newblock In \emph{Proceedings of the IEEE/CVF Conference on Computer Vision
  and Pattern Recognition}, 2021.
\newblock URL
  \url{http://openaccess.thecvf.com/content/CVPR2021/html/Weihs\_Visual\_Room\_Rearrangement\_CVPR\_2021\_paper.html}.

\bibitem[Wellhausen et~al.(2020)Wellhausen, Ranftl, and
  Hutter]{Wellhausen2020SafeRN}
Lorenz Wellhausen, Ren{\'e} Ranftl, and Marco Hutter.
\newblock {Safe Robot Navigation Via Multi-Modal Anomaly Detection}.
\newblock \emph{IEEE Robotics and Automation Letters}, 5:\penalty0 1326--1333,
  2020.
\newblock URL \url{https://api.semanticscholar.org/CorpusID:210859447}.

\bibitem[Wen et~al.(2025)Wen, Zhang, Sun, and Wang]{wen2025ovl}
Shuhuan Wen, Ziyuan Zhang, Yuxiang Sun, and Zhiwen Wang.
\newblock {OVL-MAP: An Online Visual Language Map Approach for
  Vision-and-Language Navigation in Continuous Environments}.
\newblock \emph{IEEE Robotics and Automation Letters}, 2025.
\newblock URL \url{https://ieeexplore.ieee.org/abstract/document/10879413/}.

\bibitem[Werby et~al.(2024)Werby, Huang, Büchner, Valada, and
  Burgard]{Werby-RSS-24}
Abdelrhman Werby, Chenguang Huang, Martin Büchner, Abhinav Valada, and Wolfram
  Burgard.
\newblock {Hierarchical Open-Vocabulary 3D Scene Graphs for Language-Grounded
  Robot Navigation}.
\newblock In \emph{RSS}, 2024.
\newblock URL \url{https://openreview.net/forum?id=TL0Hb9OwfR}.

\bibitem[Whelan et~al.(2015)Whelan, Leutenegger, Salas-Moreno, Glocker, and
  Davison]{whelan2015elasticfusion}
Thomas Whelan, Stefan Leutenegger, Renato~F Salas-Moreno, Ben Glocker, and
  Andrew~J Davison.
\newblock {ElasticFusion: Dense SLAM without a pose graph.}
\newblock In \emph{Robotics: science and systems}, volume~11, pp.\ ~3. Rome,
  2015.
\newblock URL \url{https://roboticsproceedings.org/rss11/p01.pdf}.

\bibitem[Whelan et~al.(2016)Whelan, Salas-Moreno, Glocker, Davison, and
  Leutenegger]{whelan2016elasticfusion}
Thomas Whelan, Renato~F Salas-Moreno, Ben Glocker, Andrew~J Davison, and Stefan
  Leutenegger.
\newblock {ElasticFusion: Real-time dense SLAM and light source estimation}.
\newblock \emph{The International Journal of Robotics Research}, 35\penalty0
  (14):\penalty0 1697--1716, 2016.
\newblock URL \url{https://roboticsproceedings.org/rss11/p01.pdf}.

\bibitem[Wijmans et~al.(2019{\natexlab{a}})Wijmans, Datta, Maksymets, Das,
  Gkioxari, Lee, Essa, Parikh, and Batra]{eqa_matterport}
Erik Wijmans, Samyak Datta, Oleksandr Maksymets, Abhishek Das, Georgia
  Gkioxari, Stefan Lee, Irfan Essa, Devi Parikh, and Dhruv Batra.
\newblock {E}mbodied {Q}uestion {A}nswering in {P}hotorealistic {E}nvironments
  with {P}oint {C}loud {P}erception.
\newblock In \emph{Proceedings of the IEEE Conference on Computer Vision and
  Pattern Recognition (CVPR)}, 2019{\natexlab{a}}.
\newblock URL
  \url{http://openaccess.thecvf.com/content\_CVPR\_2019/html/Wijmans\_Embodied\_Question\_Answering\_in\_Photorealistic\_Environments\_With\_Point\_Cloud\_Perception\_CVPR\_2019\_paper.html}.

\bibitem[Wijmans et~al.(2019{\natexlab{b}})Wijmans, Kadian, Morcos, Lee, Essa,
  Parikh, Savva, and Batra]{wijmans2019dd}
Erik Wijmans, Abhishek Kadian, Ari Morcos, Stefan Lee, Irfan Essa, Devi Parikh,
  Manolis Savva, and Dhruv Batra.
\newblock {DD-PPO}: Learning near-perfect pointgoal navigators from 2.5 billion
  frames.
\newblock In \emph{International Conference on Learning Representations},
  2019{\natexlab{b}}.
\newblock URL \url{https://arxiv.org/abs/1911.00357}.

\bibitem[Wu et~al.(2021)Wu, Wald, Tateno, Navab, and
  Tombari]{wu2021scenegraphfusion}
Shun-Cheng Wu, Johanna Wald, Keisuke Tateno, Nassir Navab, and Federico
  Tombari.
\newblock Scenegraphfusion: {I}sncremental 3d scene graph prediction from rgb-d
  sequences.
\newblock In \emph{Proceedings of the IEEE/CVF Conference on Computer Vision
  and Pattern Recognition}, pp.\  7515--7525, 2021.
\newblock URL
  \url{http://openaccess.thecvf.com/content/CVPR2021/html/Wu_SceneGraphFusion_Incremental_3D_Scene_Graph_Prediction_From_RGB-D_Sequences_CVPR_2021_paper.html}.

\bibitem[Wu et~al.(2024)Wu, Zhang, Gu, Zheng, and Bai]{wu2024embodied}
Yuchen Wu, Pengcheng Zhang, Meiying Gu, Jin Zheng, and Xiao Bai.
\newblock Embodied navigation with multi-modal information: {A} survey from
  tasks to methodology.
\newblock \emph{Information Fusion}, pp.\  102532, 2024.
\newblock URL
  \url{https://www.sciencedirect.com/science/article/pii/S1566253524003105}.

\bibitem[Xia et~al.(2018)Xia, R.~Zamir, He, Sax, Malik, and
  Savarese]{GIBSONENV}
Fei Xia, Amir R.~Zamir, Zhi-Yang He, Alexander Sax, Jitendra Malik, and Silvio
  Savarese.
\newblock {Gibson Env}: real-world perception for embodied agents.
\newblock In \emph{Proceedings of the IEEE/CVF Conference on Computer Vision
  and Pattern Recognition}, 2018.
\newblock URL
  \url{http://openaccess.thecvf.com/content\_cvpr\_2018/html/Xia\_Gibson\_Env\_Real-World\_CVPR\_2018\_paper.html}.

\bibitem[Xia et~al.(2020)Xia, Cui, Shen, Xu, Gao, and Li]{xia2020survey}
Linlin Xia, Jiashuo Cui, Ran Shen, Xun Xu, Yiping Gao, and Xinying Li.
\newblock {A survey of image semantics-based visual simultaneous localization
  and mapping: Application-oriented solutions to autonomous navigation of
  mobile robots}.
\newblock \emph{International Journal of Advanced Robotic Systems}, 17\penalty0
  (3):\penalty0 1729881420919185, 2020.
\newblock URL
  \url{https://journals.sagepub.com/doi/abs/10.1177/1729881420919185}.

\bibitem[Xiang et~al.(2020)Xiang, Qin, Mo, Xia, Zhu, Liu, Liu, Jiang, Yuan,
  Wang, et~al.]{xiang2020sapien}
Fanbo Xiang, Yuzhe Qin, Kaichun Mo, Yikuan Xia, Hao Zhu, Fangchen Liu, Minghua
  Liu, Hanxiao Jiang, Yifu Yuan, He~Wang, et~al.
\newblock {Sapien: A simulated part-based interactive environment}.
\newblock In \emph{Proceedings of the IEEE/CVF Conference on Computer Vision
  and Pattern Recognition}, pp.\  11097--11107, 2020.
\newblock URL
  \url{http://openaccess.thecvf.com/content_CVPR_2020/html/Xiang_SAPIEN_A_SimulAted_Part-Based_Interactive_ENvironment_CVPR_2020_paper.html}.

\bibitem[Xiang \& Fox(2017)Xiang and Fox]{xiang2017rnn}
Yu~Xiang and Dieter Fox.
\newblock {DA-RNN: Semantic mapping with data associated recurrent neural
  networks}.
\newblock \emph{arXiv preprint arXiv:1703.03098}, 2017.
\newblock URL \url{https://arxiv.org/abs/1703.03098}.

\bibitem[Xie et~al.(2024)Xie, Cao, Xie, Khan, and Pang]{xie2024sed}
Bin Xie, Jiale Cao, Jin Xie, Fahad~Shahbaz Khan, and Yanwei Pang.
\newblock {SED}: {A} simple encoder-decoder for open-vocabulary semantic
  segmentation.
\newblock In \emph{Proceedings of the IEEE/CVF Conference on Computer Vision
  and Pattern Recognition}, pp.\  3426--3436, 2024.
\newblock URL
  \url{http://openaccess.thecvf.com/content/CVPR2024/html/Xie_SED_A_Simple_Encoder-Decoder_for_Open-Vocabulary_Semantic_Segmentation_CVPR_2024_paper.html}.

\bibitem[Xiong et~al.(2011)Xiong, Munoz, Bagnell, and Hebert]{xiong20113}
Xuehan Xiong, Daniel Munoz, J~Andrew Bagnell, and Martial Hebert.
\newblock {3-D scene analysis via sequenced predictions over points and
  regions}.
\newblock In \emph{IEEE International Conference on Robotics and Automation},
  pp.\  2609--2616. IEEE, 2011.
\newblock URL \url{https://ieeexplore.ieee.org/abstract/document/5980125/}.

\bibitem[Xu et~al.(2024)Xu, Chen, Zhao, Wang, Zhou, and Lu]{xu2024esam}
Xiuwei Xu, Huangxing Chen, Linqing Zhao, Ziwei Wang, Jie Zhou, and Jiwen Lu.
\newblock Embodied{SAM}: {O}nline {S}egment {A}ny {3D} {T}hing in {R}eal
  {T}ime.
\newblock \emph{arXiv preprint arXiv:2408.11811}, 2024.
\newblock URL \url{https://arxiv.org/abs/2408.11811}.

\bibitem[Yadav et~al.(2022)Yadav, Ramakrishnan, Turner, Gokaslan, Maksymets,
  Jain, Ramrakhya, Chang, Clegg, Savva, Undersander, Chaplot, and
  Batra]{habitatchallenge2022}
Karmesh Yadav, Santhosh~Kumar Ramakrishnan, John Turner, Aaron Gokaslan,
  Oleksandr Maksymets, Rishabh Jain, Ram Ramrakhya, Angel~X Chang, Alexander
  Clegg, Manolis Savva, Eric Undersander, Devendra~Singh Chaplot, and Dhruv
  Batra.
\newblock Habitat challenge 2022.
\newblock \url{https://aihabitat.org/challenge/2022/}, 2022.

\bibitem[Yadav et~al.(2023)Yadav, Ramrakhya, Ramakrishnan, Gervet, Turner,
  Gokaslan, Maestre, Chang, Batra, Savva, et~al.]{yadav2023habitat}
Karmesh Yadav, Ram Ramrakhya, Santhosh~Kumar Ramakrishnan, Theo Gervet, John
  Turner, Aaron Gokaslan, Noah Maestre, Angel~Xuan Chang, Dhruv Batra, Manolis
  Savva, et~al.
\newblock Habitat-{M}atterport 3{D} {S}emantics dataset.
\newblock In \emph{Proceedings of the IEEE/CVF Conference on Computer Vision
  and Pattern Recognition}, pp.\  4927--4936, 2023.
\newblock URL
  \url{http://openaccess.thecvf.com/content/CVPR2023/html/Yadav_Habitat-Matterport_3D_Semantics_Dataset_CVPR_2023_paper.html}.

\bibitem[Yamauchi(1997)]{yamauchi1997frontier}
Brian Yamauchi.
\newblock A frontier-based approach for autonomous exploration.
\newblock In \emph{IEEE International Symposium on Computational Intelligence
  in Robotics and Automation (CIRA). 'Towards New Computational Principles for
  Robotics and Automation'}, pp.\  146--151, 1997.
\newblock URL \url{https://ieeexplore.ieee.org/abstract/document/613851/}.

\bibitem[Yang et~al.(2025)Yang, Pham, and
  Yang]{yang2025robustnesslidarbasedposeestimation}
Bo~Yang, Tri Minh~Triet Pham, and Jinqiu Yang.
\newblock {Robustness of LiDAR-Based Pose Estimation: Evaluating and Improving
  Odometry and Localization Under Common Point Cloud Corruptions}, 2025.
\newblock URL \url{https://arxiv.org/abs/2409.10824}.

\bibitem[Yang \& Scherer(2017)Yang and Scherer]{yang2017direct}
Shichao Yang and Sebastian Scherer.
\newblock Direct monocular odometry using points and lines.
\newblock In \emph{IEEE International Conference on Robotics and Automation
  (ICRA)}, pp.\  3871--3877. IEEE, 2017.
\newblock URL \url{https://ieeexplore.ieee.org/abstract/document/7989446/}.

\bibitem[Yang \& Scherer(2019)Yang and Scherer]{yang2019cubeslam}
Shichao Yang and Sebastian Scherer.
\newblock Cubeslam: {M}onocular 3-d object slam.
\newblock \emph{IEEE Transactions on Robotics}, 35\penalty0 (4):\penalty0
  925--938, 2019.
\newblock URL \url{https://ieeexplore.ieee.org/abstract/document/8708251/}.

\bibitem[Yang et~al.(2024)Yang, Yang, Zhou, Chen, Zhang, Du, and
  Gan]{yang20243dmem3dscenememory}
Yuncong Yang, Han Yang, Jiachen Zhou, Peihao Chen, Hongxin Zhang, Yilun Du, and
  Chuang Gan.
\newblock {3D-Mem}: {3D} {S}cene {M}emory for {E}mbodied {E}xploration and
  {R}easoning, 2024.
\newblock URL \url{https://arxiv.org/abs/2411.17735}.

\bibitem[Yin et~al.(2020)Yin, Cheng, Wu, Song, Yu, and Niu]{yin2020fusionlane}
Ruochen Yin, Yong Cheng, Huapeng Wu, Yuntao Song, Biao Yu, and Runxin Niu.
\newblock Fusionlane: {M}ulti-sensor fusion for lane marking semantic
  segmentation using deep neural networks.
\newblock \emph{IEEE Transactions on Intelligent Transportation Systems},
  23\penalty0 (2):\penalty0 1543--1553, 2020.
\newblock URL \url{https://ieeexplore.ieee.org/abstract/document/9237136/}.

\bibitem[Yokoyama et~al.(2023)Yokoyama, Ha, Batra, Wang, and
  Bucher]{yokoyama2023vlfm}
Naoki Yokoyama, Sehoon Ha, Dhruv Batra, Jiuguang Wang, and Bernadette Bucher.
\newblock {VLFM}: Vision-language frontier maps for zero-shot semantic
  navigation.
\newblock Workshop on Language and Robot Learning, CoRL 2023, 2023.
\newblock URL \url{https://ieeexplore.ieee.org/abstract/document/10610712/}.
\newblock Atlanta, GA, USA.

\bibitem[Yokoyama et~al.(2024)Yokoyama, Ramrakhya, Das, Batra, and
  Ha]{yokoyama2024hm3d}
Naoki Yokoyama, Ram Ramrakhya, Abhishek Das, Dhruv Batra, and Sehoon Ha.
\newblock {HM3D-OVON: A Dataset and Benchmark for Open-Vocabulary Object Goal
  Navigation}.
\newblock \emph{arXiv preprint arXiv:2409.14296}, 2024.
\newblock URL \url{https://ieeexplore.ieee.org/abstract/document/10802709/}.

\bibitem[Yoshikawa(1985)]{Yoshikawa1985ManipulabilityOR}
Tsuneo Yoshikawa.
\newblock {Manipulability of Robotic Mechanisms}.
\newblock \emph{The International Journal of Robotics Research}, 4:\penalty0 3
  -- 9, 1985.
\newblock URL \url{https://api.semanticscholar.org/CorpusID:121737271}.

\bibitem[Younes et~al.(2017)Younes, Asmar, Shammas, and
  Zelek]{younes2017keyframe}
Georges Younes, Daniel Asmar, Elie Shammas, and John Zelek.
\newblock Keyframe-based monocular {SLAM}: design, survey, and future
  directions.
\newblock \emph{Robotics and Autonomous Systems}, 98:\penalty0 67--88, 2017.
\newblock URL
  \url{https://www.sciencedirect.com/science/article/pii/S0921889017300647}.

\bibitem[Zender et~al.(2008)Zender, Mozos, Jensfelt, Kruijff, and
  Burgard]{zender2008conceptual}
Hendrik Zender, O~Mart{\'\i}nez Mozos, Patric Jensfelt, G-JM Kruijff, and
  Wolfram Burgard.
\newblock Conceptual spatial representations for indoor mobile robots.
\newblock \emph{Robotics and Autonomous Systems}, 56\penalty0 (6):\penalty0
  493--502, 2008.
\newblock URL
  \url{https://www.sciencedirect.com/science/article/pii/S0921889008000304}.

\bibitem[Zeng et~al.(2017)Zeng, Song, Yu, Donlon, Hogan, Bauz{\'a}, Ma, Taylor,
  Liu, Romo, Fazeli, Alet, Dafle, Holladay, Morona, Nair, Green, Taylor, Liu,
  Funkhouser, and Rodriguez]{Zeng2017RoboticPO}
Andy Zeng, Shuran Song, Kuan-Ting Yu, Elliott Donlon, Francois~Robert Hogan,
  Maria Bauz{\'a}, Daolin Ma, Orion Taylor, Melody Liu, Eudald Romo, Nima
  Fazeli, Ferran Alet, Nikhil~Chavan Dafle, Rachel Holladay, Isabella Morona,
  Prem~Qu Nair, Druck Green, Ian~H. Taylor, Weber Liu, Thomas~A. Funkhouser,
  and Alberto Rodriguez.
\newblock Robotic pick-and-place of novel objects in clutter with
  multi-affordance grasping and cross-domain image matching.
\newblock \emph{The International Journal of Robotics Research}, 41:\penalty0
  690 -- 705, 2017.
\newblock URL \url{https://api.semanticscholar.org/CorpusID:4548237}.

\bibitem[Zeng et~al.(2020)Zeng, Florence, Tompson, Welker, Chien, Attarian,
  Armstrong, Krasin, Duong, Sindhwani, and Lee]{Zeng2020TransporterNR}
Andy Zeng, Peter~R. Florence, Jonathan Tompson, Stefan Welker, Jonathan~M.
  Chien, Maria Attarian, Travis Armstrong, Ivan Krasin, Dan Duong, Vikas
  Sindhwani, and Johnny Lee.
\newblock {Transporter Networks: Rearranging the Visual World for Robotic
  Manipulation}.
\newblock In \emph{Conference on Robot Learning}, 2020.
\newblock URL \url{https://api.semanticscholar.org/CorpusID:225076003}.

\bibitem[Zhang et~al.(2020)Zhang, Zhu, Zheng, and Xu]{zhang2020fusion}
Jiazhao Zhang, Chenyang Zhu, Lintao Zheng, and Kai Xu.
\newblock Fusion-aware point convolution for online semantic 3d scene
  segmentation.
\newblock In \emph{Proceedings of the IEEE/CVF Conference on Computer Vision
  and Pattern Recognition}, pp.\  4534--4543, 2020.
\newblock URL
  \url{http://openaccess.thecvf.com/content_CVPR_2020/html/Zhang_Fusion-Aware_Point_Convolution_for_Online_Semantic_3D_Scene_Segmentation_CVPR_2020_paper.html}.

\bibitem[Zhang et~al.(2022{\natexlab{a}})Zhang, Dai, Meng, Fan, Chen, Xu, and
  Wang]{zhang20223d}
Jiazhao Zhang, Liu Dai, Fanpeng Meng, Qingnan Fan, Xuelin Chen, Kai Xu, and
  He~Wang.
\newblock {3D}-aware object goal navigation via simultaneous exploration and
  identification.
\newblock \emph{arXiv preprint arXiv:2212.00338}, 2022{\natexlab{a}}.
\newblock URL
  \url{http://openaccess.thecvf.com/content/CVPR2023/html/Zhang_3D-Aware_Object_Goal_Navigation_via_Simultaneous_Exploration_and_Identification_CVPR_2023_paper.html}.

\bibitem[Zhang et~al.(2023{\natexlab{a}})Zhang, Dai, Meng, Fan, Chen, Xu, and
  Wang]{zhang20233d}
Jiazhao Zhang, Liu Dai, Fanpeng Meng, Qingnan Fan, Xuelin Chen, Kai Xu, and
  He~Wang.
\newblock {3D}-aware object goal navigation via simultaneous exploration and
  identification.
\newblock In \emph{Proceedings of the IEEE/CVF Conference on Computer Vision
  and Pattern Recognition}, pp.\  6672--6682, 2023{\natexlab{a}}.
\newblock URL
  \url{http://openaccess.thecvf.com/content/CVPR2023/html/Zhang_3D-Aware_Object_Goal_Navigation_via_Simultaneous_Exploration_and_Identification_CVPR_2023_paper.html}.

\bibitem[Zhang et~al.(2025)Zhang, Hao, Xu, Zhang, Zhang, Wang, Zhang, Wang,
  Zhang, and Xu]{zhang2025mapnav}
Lingfeng Zhang, Xiaoshuai Hao, Qinwen Xu, Qiang Zhang, Xinyao Zhang, Pengwei
  Wang, Jing Zhang, Zhongyuan Wang, Shanghang Zhang, and Renjing Xu.
\newblock Map{N}av: {A} {N}ovel {M}emory {R}epresentation via {A}nnotated
  {S}emantic {M}aps for {VLM}-based {V}ision-and-{L}anguage {N}avigation.
\newblock \emph{arXiv preprint arXiv:2502.13451}, 2025.
\newblock URL \url{https://arxiv.org/abs/2502.13451}.

\bibitem[Zhang et~al.(2022{\natexlab{b}})Zhang, Hu, Xiao, and
  Zhang]{zhang2022survey}
Tianyao Zhang, Xiaoguang Hu, Jin Xiao, and Guofeng Zhang.
\newblock A survey of visual navigation: {F}rom geometry to embodied {AI}.
\newblock \emph{Engineering Applications of Artificial Intelligence},
  114:\penalty0 105036, 2022{\natexlab{b}}.
\newblock URL \url{https://ieeexplore.ieee.org/abstract/document/10577209/}.

\bibitem[Zhang et~al.(2021)Zhang, Chu, Liu, Zhang, and Zhuang]{zhang2021novel}
Xuetao Zhang, Yubin Chu, Yisha Liu, Xuebo Zhang, and Yan Zhuang.
\newblock A novel informative autonomous exploration strategy with uniform
  sampling for quadrotors.
\newblock \emph{IEEE transactions on industrial electronics}, 69\penalty0
  (12):\penalty0 13131--13140, 2021.
\newblock URL \url{https://ieeexplore.ieee.org/abstract/document/9665333}.

\bibitem[Zhang et~al.(2023{\natexlab{b}})Zhang, Huang, Ma, Li, Luo, Xie, Qin,
  Luo, Li, Liu, et~al.]{zhang2023recognize}
Youcai Zhang, Xinyu Huang, Jinyu Ma, Zhaoyang Li, Zhaochuan Luo, Yanchun Xie,
  Yuzhuo Qin, Tong Luo, Yaqian Li, Shilong Liu, et~al.
\newblock Recognize anything: A strong image tagging model.
\newblock \emph{arXiv preprint arXiv:2306.03514}, 2023{\natexlab{b}}.
\newblock URL
  \url{https://openaccess.thecvf.com/content/CVPR2024W/MMFM/html/Zhang_Recognize_Anything_A_Strong_Image_Tagging_Model_CVPRW_2024_paper.html}.

\bibitem[Zhao et~al.(2021)Zhao, Jiang, Jia, Torr, and Koltun]{zhao2021point}
Hengshuang Zhao, Li~Jiang, Jiaya Jia, Philip~HS Torr, and Vladlen Koltun.
\newblock Point transformer.
\newblock In \emph{Proceedings of the IEEE/CVF international conference on
  computer vision}, pp.\  16259--16268, 2021.
\newblock URL
  \url{https://openaccess.thecvf.com/content/ICCV2021/html/Zhao_Point_Transformer_ICCV_2021_paper.html}.

\bibitem[Zhao et~al.(2024)Zhao, Shi, and Zhuo]{zhao2024bev}
Junhui Zhao, Jingyue Shi, and Li~Zhuo.
\newblock {BEV perception for autonomous driving: State of the art and future
  perspectives}.
\newblock \emph{Expert Systems with Applications}, 258:\penalty0 125103, 2024.
\newblock URL
  \url{https://www.sciencedirect.com/science/article/pii/S0957417424019705}.

\bibitem[Zheng et~al.(2024{\natexlab{a}})Zheng, Yao, Su, Zhang, Wang, Zhao,
  Zhang, and Chau]{zheng2024survey}
Ying Zheng, Lei Yao, Yuejiao Su, Yi~Zhang, Yi~Wang, Sicheng Zhao, Yiyi Zhang,
  and Lap-Pui Chau.
\newblock A survey of embodied learning for object-centric robotic
  manipulation.
\newblock \emph{arXiv preprint arXiv:2408.11537}, 2024{\natexlab{a}}.
\newblock URL
  \url{https://link.springer.com/article/10.1007/s11633-025-1542-8}.

\bibitem[Zheng et~al.(2024{\natexlab{b}})Zheng, Chen, Zheng, Gu, Yang, Jin, Li,
  Zhong, Wang, Liu, et~al.]{zheng2024gaussiangrasper}
Yuhang Zheng, Xiangyu Chen, Yupeng Zheng, Songen Gu, Runyi Yang, Bu~Jin,
  Pengfei Li, Chengliang Zhong, Zengmao Wang, Lina Liu, et~al.
\newblock {GaussianGrasper}: {3D} language gaussian splatting for
  open-vocabulary robotic grasping.
\newblock \emph{IEEE Robotics and Automation Letters}, 2024{\natexlab{b}}.
\newblock URL \url{https://ieeexplore.ieee.org/abstract/document/10607869/}.

\bibitem[Zhi et~al.(2021)Zhi, Laidlow, Leutenegger, and Davison]{zhi2021place}
Shuaifeng Zhi, Tristan Laidlow, Stefan Leutenegger, and Andrew~J Davison.
\newblock In-place scene labelling and understanding with implicit scene
  representation.
\newblock In \emph{Proceedings of the IEEE/CVF International Conference on
  Computer Vision}, pp.\  15838--15847, 2021.
\newblock URL
  \url{https://openaccess.thecvf.com/content/ICCV2021/html/Zhi_In-Place_Scene_Labelling_and_Understanding_With_Implicit_Scene_Representation_ICCV_2021_paper.html}.

\bibitem[Zhou et~al.(2022)Zhou, Loy, and Dai]{zhou2022extract}
Chong Zhou, Chen~Change Loy, and Bo~Dai.
\newblock Extract free dense labels from {CLIP}.
\newblock In \emph{European Conference on Computer Vision}, pp.\  696--712.
  Springer, 2022.
\newblock URL
  \url{https://link.springer.com/chapter/10.1007/978-3-031-19815-1_40}.

\bibitem[Zhou et~al.(2023)Zhou, Zheng, Pryor, Shen, Jin, Getoor, and
  Wang]{zhou2023esc}
Kaiwen Zhou, Kaizhi Zheng, Connor Pryor, Yilin Shen, Hongxia Jin, Lise Getoor,
  and Xin~Eric Wang.
\newblock {ESC}: Exploration with soft commonsense constraints for zero-shot
  object navigation.
\newblock In \emph{ICML}, pp.\  42829--42842. PMLR, 2023.
\newblock URL \url{https://proceedings.mlr.press/v202/zhou23r.html}.

\bibitem[Zhu et~al.(2021)Zhu, Zhu, Lee, Liang, and Chang]{zhu2021deep}
Fengda Zhu, Yi~Zhu, Vincent Lee, Xiaodan Liang, and Xiaojun Chang.
\newblock Deep learning for embodied vision navigation: {A} survey.
\newblock \emph{arXiv preprint arXiv:2108.04097}, 2021.
\newblock URL \url{https://arxiv.org/abs/2108.04097}.

\bibitem[Zhu et~al.(2022)Zhu, Peng, Larsson, Xu, Bao, Cui, Oswald, and
  Pollefeys]{zhu2022nice}
Zihan Zhu, Songyou Peng, Viktor Larsson, Weiwei Xu, Hujun Bao, Zhaopeng Cui,
  Martin~R Oswald, and Marc Pollefeys.
\newblock {NICE-SLAM}: Neural implicit scalable encoding for {SLAM}.
\newblock In \emph{Proceedings of the IEEE/CVF conference on computer vision
  and pattern recognition}, pp.\  12786--12796, 2022.
\newblock URL
  \url{http://openaccess.thecvf.com/content/CVPR2022/html/Zhu_NICE-SLAM_Neural_Implicit_Scalable_Encoding_for_SLAM_CVPR_2022_paper.html}.

\end{thebibliography}
\bibliographystyle{tmlr}

\end{document}